\documentclass{article}
\usepackage{arxiv}

\usepackage{xcolor}
\usepackage{hyperref}
\usepackage{amsfonts}
\usepackage{bbm}
\usepackage{graphicx}
\usepackage{caption}
\usepackage{subcaption}
\usepackage{amssymb}
\usepackage{amsthm}
\usepackage{mathtools}
\usepackage{amsfonts}
\usepackage{dsfont}
\usepackage{natbib}
\allowdisplaybreaks
\usepackage{algorithm}
\usepackage{algorithmic}
\hypersetup{
	colorlinks   = true, 
	urlcolor     = blue, 
	linkcolor    = blue, 
	citecolor   = purple 
}

\newcommand{\mb}{\mathbf}
\newcommand{\mc}{\mathcal}
\newcommand{\tb}{\textbf}
\newcommand{\ti}{\textit}

\newtheorem{assumption}{Assumption}
\newtheorem{theorem}{Theorem}
\newtheorem{lemma}{Lemma}
\newtheorem{prop}{Proposition}
\newtheorem{remark}{Remark}

\title{Multi-Agent congestion cost minimization with linear function approximation}

\author{
 Prashant Trivedi \\
  Industrial Engineering and Operations Research \\
  Indian Institute of Technology Bombay India\\
  \texttt{trivedi.prashant15@iitb.ac.in} \\
   \And
 Nandyala Hemachandra\\
  Industrial Engineering and Operations Research \\
  Indian Institute of Technology Bombay India\\
  \texttt{nh@iitb.ac.in}
}

\begin{document}

\maketitle

\begin{abstract}
  This work considers multiple agents traversing a network from a source node to the goal node. The cost to an agent for traveling a link has a private as well as a congestion component. The agent's objective is to find a path to the goal node with minimum overall cost in a decentralized way. We model this as a fully decentralized multi-agent reinforcement learning problem and propose a novel multi-agent congestion cost minimization (MACCM) algorithm. Our MACCM algorithm uses linear function approximations of transition probabilities and the global cost function. In the absence of a central controller and to preserve privacy, agents communicate the cost function parameters to their neighbors via a time-varying communication network. Moreover, each agent maintains its estimate of the global state-action value, which is updated via a multi-agent extended value iteration (MAEVI) sub-routine. We show that our MACCM algorithm achieves a sub-linear regret. The proof requires the convergence of cost function parameters, the MAEVI algorithm, and analysis of the regret bounds induced by the MAEVI triggering condition for each agent. We implement our algorithm on a two node network with multiple links to validate it. We first identify the optimal policy, the optimal number of agents going to the goal node in each period. We observe that the average regret is close to zero for 2 and 3 agents. The optimal policy captures the trade-off between the minimum cost of staying at a node and the congestion cost of going to the goal node. Our work is a generalization of learning the stochastic shortest path problem.
\end{abstract}


\section{INTRODUCTION}
\label{sec: intro}
The shortest path problems are ubiquitous in many domains, such as driving directions on Google maps, automated warehouse systems, and fleet management. However, in most theoretical research, it is assumed that a single agent is traversing the network \cite{bellman1958routing,min2022learning,vial2022regret}. So, the actual traversal cost does not factor in the crucial components such as congestion due to other agents and the agent's private travel efficiency. 
In this work, we consider a multi-agent setup where a set of agents traverse through a given network from a fixed initial/source node to a pre-specified goal node in a fully decentralized way. 
The cost to an agent for traveling a network link depends on two components: 1) congestion and 2) its private operational/fuel efficiency factor. Here the congestion is the number of agents using same link. The common objective of the agents is to find a path to the goal node in a completely decentralized way, and minimizing overall congestion cost while maintaining the agents' privacy. 
The decentralized setup has two major benefits over a centralized setup: 1) it can handle humongous state and action spaces, and 2) the agents can preserve the privacy of their actions and actual rewards.
This is a generalization of the well-known learning stochastic shortest path (SSP) problem.

We model it as a fully decentralized multi-agent reinforcement learning (MARL) problem and propose a multi-agent congestion cost minimization (MACCM) algorithm. To incorporate privacy, we first parameterize the global cost function and share its parameters across the neighbors via a consensus matrix. Moreover, an agent's transition probability of going to the next node is also unknown to an agent. To this end, we propose the linear mixture MDP model, where the model parameters are expressed as the linear mixture of a given basis function. Our MACCM algorithm considers the privacy and works in an episodic manner. Each episode begins at a fixed initial node and ends if all the agents reach the goal node. In the MACCM algorithm, each agent maintains an estimate of the global state-action value function and takes actions accordingly.  
This estimate is updated according to a multi-agent extended value iteration (MAEVI) sub-routine when a `doubling criteria' is triggered. The intuition of using this 
updated estimate is that it will suggest a `better' policy. We show that the updated optimistic estimator indeed provides a better policy, and our algorithm achieves a sub-linear regret. 
Specifically, our main contributions are:

a) In Section \ref{sec: algorithm} we 
introduce a multi-agent version of SSP and propose a fully decentralized MACCM algorithm and show that it achieves a sub-linear regret (in Section \ref{sec: main_results}).
The regret depends on $\sqrt{nK}$ and $c_{min}$, where $n$ is the number of agents, $K$ is the number of episodes, and $c_{\min}$ is the minimum cost of staying at any node except the goal node.

b) To prove the regret bound (in Section \ref{sec: main_results}), we first show the convergence of the consensus based cost function parameters via the stochastic approximation method. Moreover, we show the convergence of the MAEVI algorithm. Finally, we separately bound the regret terms induced by the agents for whom MAEVI is triggered or not. The results of \cite{min2022learning} and \cite{vial2022regret} that consider learning SSP are special cases of our work (Remark \ref{rem: single_agent_reduction}).

c) To validate the usefulness of our algorithm, we provide some computational evidence on a hard instance in Section \ref{sec: computations}. In particular, we consider a network with two nodes and multiple links on each node. The average regret is very close to zero for 2 and 3 agents' cases. The regret computation requires an optimal policy, the optimal number of agents going to the goal node in each period. This optimal policy captures the agents trade-off between the minimum cost of staying at the initial node and the congestion based cost of going to the goal node. 


\section{PROBLEM SETTING}
\label{sec: problem_setting}
Let $(Z, E)$ be a given network, where $Z = \{s_{init}, 1, 2, \dots, q, g\}$ is the set of nodes and  $E = \{(i,j)~|~ i, j \in Z\}$ are the set of edges in the network, where $s_{init}$ and $g$ are fixed initial and the goal nodes respectively. Let $N = \{1,2,\dots, n\}$ be the set of agents. The common objective of agents is to
traverse through the network from initial node $s_{init}$ to a goal node $g$ while minimizing the sum of all agents' path travel costs. The cost incurred by an agent includes a private efficiency component and another component that is congestion based, as given in Eq. \eqref{eqn: agent_i_cost} below. To achieve this objective and to preserve privacy, we
model this problem as a fully decentralized multi-agent reinforcement learning (MARL) and provide a Multi-Agent Congestion Cost Minimization (MACCM) algorithm that achieves a sub-linear regret.

Formally, an instance of MACCM problem is described as $(N, \mb{\mc{S}} \cup \{\textbf{\textit{s}}_{init}\}, \textbf{\textit{g}}, \{\mb{\mathcal{A}}^i\}_{i\in N}, \{c^i\}_{i\in N}, \mathbb{P}, \{\mathcal{G}_t\}_{t\geq 0})$.
Here $\mathcal{S}\cup \{\textbf{\textit{s}}_{init}\}$ is the global state space with a fixed source state $\textbf{\textit{s}}_{init}$. Each $\textbf{\textit{s}} \in \mb{\mathcal{S}}$ is a vector of size $n$ representing the node at which each agent is present in that order, that is $\textbf{\textit{s}}  = (s^1,s^2, \dots, s^n)$, where $s^i \in Z$ for each agent $i\in N$. Often we write $\tb{\ti{s}} = (s^i, \tb{\ti{s}}^{-i})$ to denote the state $\tb{\ti{s}}$, where agent $i$ is at node $s^i$ and $\tb{\ti{s}}^{-i} = (s^1, \dots, s^{i-1}, s^{i+1}, \dots, s^n)$ is the node of all but agent $i$. 

Let $\mc{A}^{i}(\tb{\ti{s}})$ be the set of links available to agent $i\in N$ when the global state is $\tb{\ti{s}}$. So, the global action at the state $\tb{\ti{s}}$ is $\mb{\mc{A}}(\tb{\ti{s}}) = \prod_{i\in N} \mb{\mc{A}}^i(\tb{\ti{s}})$. A typical element in $\mb{\mc{A}}(\tb{\ti{s}})$ is a vector of size $n$, one for each agent as $(\tb{\ti{a}}^1(\tb{\ti{s}}), \tb{\ti{a}}^2(\tb{\ti{s}}), \dots, \tb{\ti{a}}^n(\tb{\ti{s}}))$, where $\tb{\ti{a}}^i(\tb{\ti{s}}) \in \mb{\mc{A}}^i(\tb{\ti{s}}) $. Here $\tb{\ti{a}}^i(\tb{\ti{s}})$ represents the action taken by agent $i$ when the global state is $\tb{\ti{s}}$. However, the action taken by each agent is private information and not known to other agents. 
For a global state $\tb{\ti{s}}$, and global action $\tb{\ti{a}}$, each agent $i\in N$ realizes a local cost $c^i(\tb{\ti{s}}, \tb{\ti{a}})$. It is important to note that the cost of agent $i$ depends on the action taken by other agents also. In this work, we assume that the cost $c^i(\tb{\ti{s}}, \tb{\ti{a}})$ depends on two components. The first one is the private component that is local to the agent, and the second is the congestion component. 
The private component might represent the agent's efficiency. The congestion is defined as the number of agents using the same link. In particular,  
\begin{equation}
\label{eqn: agent_i_cost}
c^i(\tb{\ti{s}}, \tb{\ti{a}}) = K^i(\tb{\ti{s}}, \tb{\ti{a}}) \cdot \sum_{j\in N} \mathds{1}_{ \{\tb{\ti{a}}^i(\tb{\ti{s}}) = \tb{\ti{a}}^j(\tb{\ti{s}})\} },
\end{equation}
where $K^i(\tb{\ti{s}}, \tb{\ti{a}})$ is a bounded private component of the $i$-th agent  cost, not known to other agents, and the summation of indicators is the congestion seen by agent $i$ in the state $\tb{\ti{s}}$. This implies, $c^i(\cdot, \cdot)$ is bounded. We assume all the costs are realized/incurred just before a period ends.

Moreover, we assume that $c^i(\cdot, \cdot) \geq c_{min} >0$. This assumption is common in many recent works for single agent Stochastic Shortest Path (SSP) problem \citep{rosenberg2020near,tarbouriech2020no}. $c_{\min} >0$  ensures that agents do not have the incentive to wait indefinitely at any node for the clearance of congestion.
Suppose $\tb{\ti{s}} = (g, \tb{\ti{s}}^{-i})$, that is agent $i$ has reached the goal state $g$ and the other agents are somewhere else in the network, then agent $i$ stays at goal node, and $c^i((g,  \tb{\ti{s}}^{-i}), \tb{\ti{a}}) = 0$. Moreover, once an action $\tb{\ti{a}}$ is taken in state $\tb{\ti{s}}$ the state will change to $\tb{\ti{s}}^{\prime}$ with probability $\mathbb{P}(\tb{\ti{s}}^{\prime} | \tb{\ti{s}}, \tb{\ti{a}})$ and this continue till $\tb{\ti{s}} = (g,\dots, g) = \tb{\ti{g}}$, i.e., all the agents reach goal state. 

Let $T^{\pi} (\tb{\ti{s}})$ be the time to reach the goal state $\tb{\ti{g}}$ starting from the state $\tb{\ti{s}}$ and following policy $\pi$. A stationary and deterministic policy $\pi: \mb{\mc{S}} \rightarrow \mb{\mc{A}}$ is called a proper policy if $T^{\pi} (\tb{\ti{s}}) < \infty$ almost surely. Let $\Pi_p$ be the set of all stationary, deterministic and proper policies. We make the following assumption about proper policy, common in SSP literature \cite{tarbouriech2020no,cohen2021minimax}.
\begin{assumption}[Proper Policy Existence]
	\label{ass: existence_proper_policy}
	There exists at least a proper policy, meaning $\Pi_p \neq \emptyset$.
\end{assumption}
Next, we define the cost-to-go or the value function for a given policy $\pi$ 
\begin{equation}
\label{eqn: global-obj}
V^{\pi}(\tb{\ti{s}}) \coloneqq \underset{T \rightarrow \infty}{\lim} \mathbb{E}_{\pi} \Big[\sum_{t=1}^{T} \bar{c}(\tb{\ti{s}}_t, \pi(\tb{\ti{s}}_t)) \Big| \tb{\ti{s}}_1 = \tb{\ti{s}} \Big],
\end{equation}
where $\bar{c}(\tb{\ti{s}}_t, \pi(\tb{\ti{s}}_t)) \coloneqq  \frac{1}{n} \sum_{i=1}^n c^i(\tb{\ti{s}}_t, \pi(\tb{\ti{s}}_t))$. So, our global objective translates to finding a policy $\pi^{\star}$ such that $\pi^{\star} = arg \min_{\pi \in \Pi_p} V^{\pi}(\tb{\ti{s}}_{init})$. Let $V^{\star} = V^{\pi^{\star}}$ be the value of the optimal policy $\pi^{\star}$. We also define the state-action value function $Q^{\pi}(\tb{\ti{s}}, \tb{\ti{a}})$ of a policy $\pi$ as 
\begin{equation}
\underset{T \rightarrow \infty}{\lim} \mathbb{E}_{\pi} \Big[\bar{c}(\tb{\ti{s}}_1, \tb{\ti{a}}_1) + \sum_{t=2}^{T}  \bar{c}(\tb{\ti{s}}_t, \pi(\tb{\ti{s}}_t)) \Big| \tb{\ti{s}} = \tb{\ti{s}}_1, \tb{\ti{a}} = \tb{\ti{a}}_1 \Big].
\end{equation}
Since $\bar{c}(\cdot, \cdot)$ is bounded, for any proper policy $\pi\in \Pi_p$, both $V^{\pi}(\cdot)$ and $Q^{\pi}(\cdot, \cdot)$ are also bounded. This work assumes that the transition probability function $\mathbb{P}$ is written as the linear mixture of given basis functions \citep{min2022learning,vial2022regret}. In particular, we make the following assumption about the transition probability function.
\begin{assumption}[Transition probability approximation]
	\label{ass: tpm_approx}
	Suppose the feature mapping $\phi: \mb{\mc{S}} \times \mb{\mc{A}} \times \mb{\mc{S}} \rightarrow \mathbb{R}^{nd}$ is known and pre-given. There exists a $\boldsymbol{\theta}^{\star} \in \mathbb{R}^{nd}$ with  $|| \boldsymbol{\theta}^{\star}||_2 \leq \sqrt{nd}$ such that $\mathbb{P}(\tb{\ti{s}}^{\prime} | \tb{\ti{s}}, \tb{\ti{a}}) = \left\langle \phi(\tb{\ti{s}}^{\prime}| \tb{\ti{s}},\tb{\ti{a}}), \boldsymbol{\theta}^{\star} \right\rangle$ for any triplet $(\tb{\ti{s}}^{\prime}, \tb{\ti{a}},\tb{\ti{s}}) \in \mb{\mc{S}} \times \mb{\mc{A}} \times \mb{\mc{S}}$. Also, for a bounded function $V: \mathcal{S} \mapsto [0, B]$, it holds that $|| \phi_V(\tb{\ti{s}},\tb{\ti{a}})||_2 \leq B \sqrt{nd}$, where $\phi_V(\tb{\ti{s}},\tb{\ti{a}}) = \sum_{\tb{\ti{s}}^{\prime} \in \mathcal{S}} \phi(\tb{\ti{s}}^{\prime}|\tb{\ti{s}},\tb{\ti{a}}) V(\tb{\ti{s}}^{\prime})$. 
\end{assumption}
For simplicity of notation, given any function $V: \mathcal{S} \rightarrow [0, B]$ we define $\mathbb{P} V(\tb{\ti{s}},\tb{\ti{a}}) = \sum_{\tb{\ti{s}}^{\prime} \in \mathcal{S}} \mathbb{P}(\tb{\ti{s}}^{\prime}|\tb{\ti{s}},\tb{\ti{a}}) V(\tb{\ti{s}}^{\prime})$, $\forall~(\tb{\ti{s}},\tb{\ti{a}}) \in \mathcal{S} \times \mathcal{A}$. Then, under the Assumption \ref{ass: tpm_approx} we have,
\begin{equation*}
\mathbb{P} V(\tb{\ti{s}},\tb{\ti{a}}) = \sum_{\tb{\ti{s}}^{\prime} \in \mathcal{S}} \left\langle  \phi(\tb{\ti{s}}^{\prime}|\tb{\ti{s}},\tb{\ti{a}}) , \boldsymbol{\theta}^{\star} \right\rangle  V(\tb{\ti{s}}^{\prime}) =  \langle \phi_V(\tb{\ti{s}},\tb{\ti{a}}), \boldsymbol{\theta}^{\star} \rangle.
\end{equation*}
Moreover, for any function $V: \mathcal{S} \rightarrow [0, B]$, we define the Bellman operator $\mathcal{L}$ as $\mathcal{L}V (\tb{\ti{s}}) \coloneqq \min_{\tb{\ti{a}}\in \mathcal{A}} \{\bar{c}(\tb{\ti{s}},\tb{\ti{a}}) + \mathbb{P}V(\tb{\ti{s}},\tb{\ti{a}}) \}.$ Throughout, we assume that $B_{\star}$ is the upper bound on the optimal value function $V^{\star}$, i.e., $B_{\star} = \max _{\tb{\ti{s}} \in \mathcal{S}} V^{\star}(\tb{\ti{s}})$. Without loss of generality, we assume that $B_{\star} \geq 1$, and denote the optimal state-action value by $Q^{\star} = Q^{\pi^{\star}}$, which satisfy the following Bellman equation for all $(\tb{\ti{s}},\tb{\ti{a}}) \in \mathcal{S} \times \mathcal{A}$
\begin{equation*}
Q^{\star}(\tb{\ti{s}},\tb{\ti{a}}) = \bar{c}(\tb{\ti{s}},\tb{\ti{a}}) + \mathbb{P}V^{\star}(\tb{\ti{s}},\tb{\ti{a}});V^{\star}(\tb{\ti{s}}) = \min_{\tb{\ti{a}}\in \mathcal{A}} Q^{\star}(\tb{\ti{s}}, \tb{\ti{a}}). 
\end{equation*}
However, in our decentralized model, the global cost $\bar{c}(\cdot, \cdot)$ is unknown to any agent.
So, at every decision epoch, each agent shares some parameters of the model using a time-varying communication network $\mathcal{G}_t$ to its neighbors. To this end, we
propose to estimate the globally averaged cost function $\bar{c}$. 
Let $\bar{c}(\cdot, \cdot; \tb{\ti{w}}) : \mathcal{S} \times \mathcal{A} \rightarrow \mathbb{R}$ be the class of parameterized functions where $\tb{\ti{w}} \in \mathbb{R}^k$ for some $k << |\mathcal{S}||\mathcal{A}|$. To obtain the estimate $\bar{c}(\cdot, \cdot; \tb{\ti{w}})$ we seek to minimize the following least square estimate
\begin{equation}
\label{eqn: op}
\tag{OP 1}
\min_{\tb{\ti{w}}}~~ \mathbb{E}_{\tb{\ti{s}},\tb{\ti{a}}} [\bar{c}(\tb{\ti{s}},\tb{\ti{a}}) - \bar{c}(\tb{\ti{s}},\tb{\ti{a}};\tb{\ti{w}})]^2.
\end{equation}
A key result that ensures the working of a  decentralized algorithm is the following; see also \cite{zhang2018fully,trivedi2022multi} 
\begin{prop} 
	The optimization problem in Eq. \eqref{eqn: op} is equivalently characterized as (both have the same stationary points)
	\begin{equation}
	\label{eqn: op_equivalent}
	\tag{OP 2}
	\min_{\tb{\ti{w}}} \sum_{i=1}^n \mathbb{E}_{\tb{\ti{s}},\tb{\ti{a}}} [c^i(\tb{\ti{s}},\tb{\ti{a}}) - \bar{c}(\tb{\ti{s}},\tb{\ti{a}};\tb{\ti{w}})]^2.
	\end{equation}
\label{prop: equiv-opti-problem}
\end{prop}
The proof details are available in the Appendix \ref{app: equiv-opti-problem}.
Note that the objective function in Eq. \eqref{eqn: op_equivalent} has the same form with separable objectives over agents as in the distributed optimization literature \cite{nedic2009distributed,boyd2006randomized}. This motivates the following updates for parameters of the global cost function estimate by agent $i$, $\tb{\ti{w}}^i$ to minimize the objective in Eq. \eqref{eqn: op_equivalent}
\begin{equation}
\begin{aligned}
\widetilde{\tb{\ti{w}}}^i_{t} \leftarrow \tb{\ti{w}}^i_t &+ \gamma_{t} \cdot [c^i_t(\cdot, \cdot) - \bar{c}(\cdot, \cdot; \tb{\ti{w}}^i_t)] \cdot \nabla_{\tb{\ti{w}}} \bar{c}(\cdot,\cdot; \tb{\ti{w}}^i_t)
\\
\tb{\ti{w}}^i_{t+1} &= \sum_{j\in N} l_t(i,j) \widetilde{\tb{\ti{w}}}^j_{t},
\label{eqn: w_update}
\end{aligned}
\end{equation}
where $l_t(i,j)$ is the $(i,j)$-th entry of the consensus matrix $L_t$ obtained using communication network $\mathcal{G}_t$ at time $t$. $\gamma_t$ is the step-size satisfying $\sum_{t} \gamma_t = \infty$ and $\sum_t \gamma_t^2 < \infty$, and $\bar{c}(\cdot, \cdot; \tb{\ti{w}}^i_t)$ is the estimate of global cost function by agent $i$ at time $t$. 
In the above equation, at time $t$, each agent updates an intermediate cost function parameters $\widetilde{\tb{\ti{w}}}^i_t$ using the stochastic gradient descent method to get the minima of the optimization problem given in \eqref{eqn: op_equivalent}. Each agent shares these intermediate parameters to the neighbors via the communication matrix and update the true parameters $\tb{\ti{w}}^i$.

The central idea of using the communication/consensus matrix in decentralized RL algorithms is to share some information among the consensus matrix neighbors while preserving privacy. Actions, which are private information, are not shared.
However, a global objective cannot be attained without a central controller and without sharing any information or parameters. 
Thus, the cost function parameters $\textbf{\textit{w}}$'s (not actual costs)  
are shared with the neighbors as per the communication matrix. They converge to their true parameters \emph{a.s.} (Theorem \ref{thm: weight_convergence}). Such a sharing of the cost function parameters via communication matrix to the neighbors is an intermediate construct that maintains privacy  of each actions and costs, but, achieves the global objective, Equation \eqref{eqn: global-obj}.
We make following assumption \citep{zhang2018fully,bianchi2013performance} on communication matrix $\{L_t\}_{t\geq 0}$. 

\begin{assumption}[Consensus matrix]
	\label{ass: comm_matrix}
	The consensus matrices $\{L_t\}_{t \geq 0} \subseteq \mathbb{R}^{n\times n}$ satisfy (i) $L_t$ is row stochastic, i.e., $L_t \mathds{1} = \mathds{1}$ and $\mathbb{E}(L_t)$ is column stochastic, i.e., $\mathds{1}^{\top}\mathbb{E}(L_t)=\mathds{1}^{\top}$. Further, there exists a constant $\kappa \in (0,1)$ such that for any $l_t(i,j)>0$, we have $l_t(i,j)\geq \kappa$; (ii) Consensus matrix $L_t$ respects $\mathcal{G}_t$, i.e., $l_t(i,j) = 0$, if $(i,j)\notin \mathcal{E}_t$; (iii) The spectral norm of $\mathbb{E}[L_t^{\top} (I - \mathds{1}\mathds{1}^{\top}/n)L_t]$ is less than one.
\end{assumption}
We make the following assumption \citep{zhang2018fully,trivedi2022multi} on the features associated with the cost function while showing the convergence of the cost function parameters $\tb{\ti{w}}$ (Theorem \ref{thm: weight_convergence}).

\begin{assumption}[Full rank]
	\label{ass: full_rank}
	For each agent $i\in N$, the cost function $\bar{c}(\tb{\ti{s}},\tb{\ti{a}})$ is parameterized as $\bar{c}(\tb{\ti{s}},\tb{\ti{a}}; \tb{\ti{w}}) = \langle \psi(\tb{\ti{s}},\tb{\ti{a}}), \tb{\ti{w}} \rangle$. Here $\psi(\tb{\ti{s}},\tb{\ti{a}})$ = $[\psi_1(\tb{\ti{s}},\tb{\ti{a}}), \dots, \psi_k(\tb{\ti{s}},\tb{\ti{a}})]\in \mathbb{R}^k$ are the features associated with pair $(\tb{\ti{s}},\tb{\ti{a}})$. Further, we assume that these features are uniformly bounded. 
	Moreover, let the feature matrix $\Psi \in \mathbb{R}^{|\mathcal{S}||\mathcal{A}|\times k}$  have $[\psi_m(\tb{\ti{s}},\tb{\ti{a}}), \tb{\ti{s}} \in \mathcal{S}, \tb{\ti{a}} \in \mathcal{A}]^{\top}$ as its $m$-th column for any $m\in [k]$, then $\Psi$ has full column rank.
\end{assumption}

Since the global cost is unknown, each agent uses the parameterized cost and maintains its estimate of $V(\cdot)$ and $Q(\cdot, \cdot)$. Let $V^i(\cdot)$ and $Q^i(\cdot, \cdot)$ be the estimate of these functions by agent $i$. So, the modified Bellman optimality equation for all $(\tb{\ti{s}},\tb{\ti{a}})$ and for all agents $i\in N$ is 
\begin{equation}
\label{eqn: state-action-convergence}
\begin{split}
Q^{i\star}(\tb{\ti{s}},\tb{\ti{a}}; \tb{\ti{w}}^i) = \bar{c}(\tb{\ti{s}},\tb{\ti{a}};\tb{\ti{w}}^{i}) + \mathbb{P}V^{i\star}(\tb{\ti{s}},\tb{\ti{a}}; \tb{\ti{w}}^i)
\\
V^{i\star}(\tb{\ti{s}}; \tb{\ti{w}}^i) = \min_{\tb{\ti{a}}\in \mathcal{A}} Q^{i\star}(\tb{\ti{s}}, \tb{\ti{a}}; \tb{\ti{w}}^i)
\end{split}
\end{equation}
We later show in Theorem \ref{thm: weight_convergence} that $\tb{\ti{w}}^i_t \rightarrow \tb{\ti{w}}^{\star}$. Hence $\bar{c}(\tb{\ti{s}},\tb{\ti{a}}; \tb{\ti{w}}^i_t) \rightarrow \bar{c}(\tb{\ti{s}},\tb{\ti{a}}; \tb{\ti{w}}^{\star})$, $Q^{i\star}(\tb{\ti{s}},\tb{\ti{a}}; \tb{\ti{w}}^i_t) \rightarrow Q^{i\star}(\tb{\ti{s}},\tb{\ti{a}})$ and $V^{i\star}(\tb{\ti{s}};\tb{\ti{w}}^i_t) \rightarrow V^{i\star}(\tb{\ti{s}}) $
as $\bar{c}(\tb{\ti{s}},\tb{\ti{a}}; \tb{\ti{w}}^i_t)$, $Q^{i\star}(\tb{\ti{s}},\tb{\ti{a}}; \tb{\ti{w}}^i_t)$ and $V^{i\star}(\tb{\ti{s}};\tb{\ti{w}}^i_t)$ are continuous functions of $\tb{\ti{w}}^i$, where $Q^{i\star}(\tb{\ti{s}},\tb{\ti{a}})$ and $V^{i\star}(\tb{\ti{s}})$ are defined as
\begin{equation}
Q^{i\star}(\tb{\ti{s}},\tb{\ti{a}}) = \bar{c}(\tb{\ti{s}},\tb{\ti{a}};\tb{\ti{w}}^{\star}) + \mathbb{P}V^{i\star}(\tb{\ti{s}},\tb{\ti{a}});~ V^{i\star}(\tb{\ti{s}}) = \min_{\tb{\ti{a}}\in \mathcal{A}} Q^{i\star}(\tb{\ti{s}}, \tb{\ti{a}}).
\label{eqn: q-defined}
\end{equation}
With the above assumptions, we aim to design an algorithm for the episodic setting where an episode begins from a common initial state $\tb{\ti{s}}_{init}$ and ends at $\tb{\ti{g}}$ such that the following regret over $K$ episodes is minimized
\begin{equation}
\label{eqn: R_K}
R_K  = \sum_{j=1}^{K} \sum_{l=1}^{I_j}  \frac{1}{n}  \sum_{i \in N} \left( \bar{c}(\tb{\ti{s}}_{j,l}, \tb{\ti{a}}_{j,l}; \tb{\ti{w}}^i_{j,l}) - K \cdot V^{i\star}(\tb{\ti{s}}_{init}) \right),
\end{equation}
here $I_j$ is the length of the episode $j=1, \dots, K$, and $\bar{c}(\tb{\ti{s}}_{j,l}, \tb{\ti{a}}_{j,l}; \tb{\ti{w}}^i_{j,l})$ is the estimate of the global cost function by agent $i$ in the $l$-th step of the $j$-th episode. 
Note that in the above regret expression, instead of the global optimal value, we use the average of $V^{i{\star}}$, averaged over all the agents. This is because \textbf{(1)} $V^{\star}$ is not available to any agent; however as mentioned above, each agent $i$ maintains its estimate $V^{i\star}$.  \textbf{(2)} Theorem \ref{thm: weight_convergence} implies $V^{i\star} = V^{\star}$ for all $i\in N$; however, we write $V^{i\star}$ in the regret definition to avoid any confusions, in both the Equations \eqref{eqn: q-defined} and \eqref{eqn: R_K}. Our proofs will remain the same, with this minor change. \textbf{(3)} We empirically observe that $V^{i\star} = V^{\star}$ for all $i\in N$. So, the regret definition in Equation \eqref{eqn: R_K} is the same as the true regret in terms of true optimal values.  In the next section, we present the MACCM algorithm that is fully decentralized and achieves a sub-linear regret.

\section{MACCM ALGORITHM}
\label{sec: algorithm}
We next describe the MACCM algorithm design. It is inspired by the single agent UCLK algorithm for discounted linear mixture MDPs of \cite{zhou2021nearly}. It also uses some structure of the LEVIS algorithm of \cite{min2022learning}.

\begin{algorithm}[h!]
\begin{algorithmic}[1]
\STATE{\textbf{Input:}} regularization parameter $\lambda$, confidence radius $\{\beta_t\}$, an estimate $B \geq B_{\star}$.
\STATE{\textbf{Initialize:}} set $t \leftarrow 1$. For each agent $i\in N$, set $j_i = 0, t^i_0 = 0, \Sigma^i_0 = \lambda I, b^i_0 = 0$, $Q^i_0(\tb{\ti{s}}, \cdot)$, $V^i_0(\tb{\ti{s}}) = 1,~ \forall~\tb{\ti{s}}\neq \tb{\ti{g}}$, and $0$ otherwise; $\tb{\ti{w}}^i_0 = \mathbf{0}$; $\gamma_t = \frac{1}{t+1}$.
\FOR{$k=1, \dots, K$}
\STATE Set $\tb{\ti{s}}_{t} = \tb{\ti{s}}_{init} = (s_{init}, s_{init}, \dots, s_{init})$.
\WHILE{$\tb{\ti{s}}_t \neq \tb{\ti{g}}$}
\FOR {$i\in N$}
\IF{$s^i_t \neq g$}
\STATE $\tb{\ti{a}}^i_t  = arg \min_{\tb{\ti{a}}\in \mathcal{A}^i_t}~ \max_{\tb{\ti{a}}^{-i}_t \in \mathcal{A}^{-i}_t} {Q}^i_{j_i}(\tb{\ti{s}}_t,\tb{\ti{a}}, \tb{\ti{a}}^{-i}_t)$ 
\ELSE 
\STATE $\tb{\ti{a}}^i_t = g$ 
\ENDIF
\ENDFOR
\STATE Set $\tb{\ti{a}}_t = (\tb{\ti{a}}^1_t,\tb{\ti{a}}^2_t,\cdots, \tb{\ti{a}}^n_t)$
\STATE receive cost $\bar{c}_t (\tb{\ti{s}}_t, \tb{\ti{a}}_t; \tb{\ti{w}}) =  \frac{1}{n} \sum_{i=1}^n \bar{c}(\tb{\ti{s}}_t, \tb{\ti{a}}_t; \tb{\ti{w}}^i_t)$
\STATE next state $\tb{\ti{s}}_{t+1} \sim \mathbb{P}(\cdot | \tb{\ti{s}}_t, \tb{\ti{a}}_t)$ 
\FOR{$i\in N$}
\STATE Set $\widetilde{\tb{\ti{w}}}^i_t \leftarrow \tb{\ti{w}}^i_t + \gamma_t [c^i(\tb{\ti{s}}_t, \tb{\ti{a}}_t) - \bar{c}(\tb{\ti{s}}_t,\tb{\ti{a}}_t ;\tb{\ti{w}}^i_t)] \cdot \nabla_\tb{\ti{w}} \bar{c}(\tb{\ti{s}}_t,\tb{\ti{a}}_t; \tb{\ti{w}}^i_t )$
\STATE Set $\Sigma^i_t \leftarrow \Sigma^i_{t-1} + \phi_{{V}^i_{j_i}}(\tb{\ti{s}}_t, \tb{\ti{a}}_t)\phi_{{V}^i_{j_i}}(\tb{\ti{s}}_t, \tb{\ti{a}}_t)^{\top}$
\STATE Set $b^i_t \leftarrow b^i_{t-1} + \phi_{{V}^i_{j_i}}(\tb{\ti{s}}_t, \tb{\ti{a}}_t) {V}^i_{j_i}(\tb{\ti{s}}_{t+1})$ 
\ENDFOR
\STATE Set $S_t = \emptyset, S_t^c = N$
\FOR{$i\in N$}
\IF{$det(\Sigma^i_t) \geq 2 det(\Sigma^i_{t^i_{j_i}})$ or $t \geq 2 t^i_{j_i}$ } 
\STATE Set $S_t = S_t \cup \{i\}$; and $S_t^c = S_t^c \setminus \{i\}$
\ENDIF
\ENDFOR
\FOR{$i\in S_t$}
\STATE Set $j_i\leftarrow j_i+1$; $t^i_{j_i} \leftarrow t$, and $\epsilon_{j_i} \leftarrow \frac{1}{t^i_{j_i}}$
\STATE $\hat{\theta}^i_{j_i} \leftarrow \Sigma^{i^{-1}}_t b^i_t$
\STATE Set $\mathcal{C}^i_{j_i} \leftarrow \left\lbrace  \theta : || \Sigma_{t^i_{j_i}}^{i^{1/2}} (\theta -\hat{\theta}^i_{j_i})||_2 \leq \beta_{t_{j_i}} \right\rbrace$
\STATE Set ${Q}^i_{j_i}(\cdot, \cdot) \leftarrow$ MAEVI$(\mathcal{C}^i_{j_i}, \epsilon_{j_i}, \frac{1}{t^i_{j_i}}, \tb{\ti{w}}^i_t)$
\STATE Set $ V^i_{j_i}(\cdot) = \min_
{\tb{\ti{a}}^i\in \mathcal{A}^i } Q^i_{j_i}(\cdot, \tb{\ti{a}}^i, \tb{\ti{a}}^{-i})$
\ENDFOR
\STATE Send parameters to the neighbors 
\STATE Get $\tb{\ti{w}}^i_{t+1} = \sum_{j\in N} l_t(i,j) \widetilde{\tb{\ti{w}}}^j_{t}$
\STATE Set $t\leftarrow t+1$
\ENDWHILE
\ENDFOR
\caption{MACCM
} 
\label{algo: main_algo_multi_agent}
\end{algorithmic}
\end{algorithm}

Let $t$ be the global time index, and $K$ be the number of episodes. Each episode starts at fixed state $\tb{\ti{s}}_{init} = (s_{init}, s_{init}, \dots, s_{init})$ and ends when all the agents reach to the goal state $\tb{\ti{g}} = (g, g, \dots, g)$. 
An episode $k$ is decomposed into many epochs; let $j_i$ denote the $j$-th epoch of the agent $i$. Within this epoch, agent $i$ uses $Q^i_{j_i}$ as an optimistic estimator of the global state-action value function. 

Initially, for each agent $i\in N$, the estimate of the global state value function $V^i$ and the global state-action value function $Q^i$ is taken as 
$1$ for all $\tb{\ti{s}} \neq \tb{\ti{g}}$, and $0$ for $\tb{\ti{s}} = \tb{\ti{g}}$. In each episode $k$, if an agent $i\in N$ is not at the goal node, it takes action using the current optimistic estimator of the global state-action value $Q^i_{j_i}$. In particular, the action taken by agent $i$ is according to the $\min \max$ criteria that captures its best action against the worst possible action in terms of congestion by other agents. However, if agent $i$ has reached the goal node, it will stay there until the episode gets over, \textit{i.e.}, till all other agents reach the goal node (lines 5-12).

(Lines 16-20) Apart from executing a policy that uses an optimistic estimator, each agent $i\in N$ also updates the $\Sigma_t$ and $b_t$. These updates are used to estimate the true model parameters. 
$\Sigma_t$ and $b_t$ together are inspired from the ridge regression based minimizer of the model parameters; similar updates were used by \cite{zhou2021nearly,abbasi2011improved} in single agent model. The doubling criteria are used to update the optimistic estimator of the state-action value function. The determinant doubling criteria reflects the diminishing returns of the underlying transition. However, more than this update is required as it cannot guarantee the finite length of the epoch, so a simple time doubling criteria is used.  Moreover, the cost function parameters are also updated as given in Eq. \eqref{eqn: w_update}. 

In lines 21-26 of the algorithm, we maintain a set $S_t$ containing those agents for whom the doubling criteria are satisfied at time $t$. The determinant doubling criteria is used in many previous works of linear bandits, and RL \citep{abbasi2011improved,zhou2021nearly}. It is often referred to as lazy policy update. It reflects the diminishing returns of learning the underlying transitions. However, the determinant doubling criteria alone is insufficient and cannot guarantee the finite length of each epoch since feature norm $|| \phi_{V^i}(\cdot, \cdot)||$ are not bounded from below. So, a simple time doubling criteria is introduced.
This criterion possesses many excellent properties, including the easiness of implementation and the low space and time complexity. 

(Lines 27-33) We switch the epoch for all agents $i\in S_t$ and update their optimistic estimator of the state-action value function using the multi-agent extended value iteration (MAEVI) subroutine.
Moreover, we set the MAEVI error parameter $\epsilon_{j_i} = \frac{1}{t_{j_i}}$ to bound the cumulative error from the value iterations by a constant, i.e., $(2t_{j_i} - t_{j_i})\cdot \epsilon_{j_i} = 1$. 

The optimism in the MACCM algorithm in period $t$ is due to the construction of confidence set $\mathcal{C}^i_{j_i}$ for all the agents $i\in S_t$, which is input to the MAEVI sub-routine. The MAEVI sub-routine requires a confidence ellipsoid $\mathcal{C}^i_{j_i}$ containing true model parameters. This confidence set is constructed in line 30 of the MACCM algorithm, which is obtained via minimization of a suitable ridge regression problem with confidence radius $\beta_t$. Moreover, we construct a set $\mathcal{B}$ to ensure that the model parameters form a valid transition probability function. In particular, the model parameters are taken from $\mathcal{C}^i_{j_i} \cap \mathcal{B}$. Here $\mathcal{B} \coloneqq \{\boldsymbol{\theta} : \forall (\tb{\ti{s}},\tb{\ti{a}}), \left\langle \phi(\cdot | \tb{\ti{s}},\tb{\ti{a}}), \boldsymbol{\theta} \right\rangle$ is probability distribution and $\left\langle \phi(\tb{\ti{s}}^{\prime} | \tb{\ti{s}},\tb{\ti{a}})\right\rangle =  \mathds{1}_{\{\tb{\ti{s}}^{\prime} \neq \tb{\ti{g}}\}} \}$. In Theorem \ref{thm: MAEVI_analysis} we prove that the true model parameters  $\boldsymbol{\theta}^{\star}$ are in the set $\mathcal{C}^i_{j_i} \cap \mathcal{B}$ with high probability. 

The MAEVI algorithm uses a discount term $q$, which is key in the convergence of the  MAEVI sub-routine. This is because $\langle \cdot, \phi_{V^i}(\cdot, \cdot) \rangle$ is not a contractive map, so  we use an extra discount term $(1-q)$ that provides the contraction property. This may lead to an additional bias that can be suppressed by suitably choosing a $q$. Particularly, we choose $q = \frac{1}{t_{j_i}}$, and it will yield an additional regret of $\mathcal{O}(\log T)$ in the final regret. 
The term $(1-q)$ biases the estimated transition kernel towards the goal state $\tb{\ti{g}}$, which also encourages further optimism. Similar design is also available in \cite{min2022learning,tarbouriech2021stochastic}. 

It is important to note that we use the stochastic approximation based rule to update the cost function parameters $\tb{\ti{w}}^i$. We want to emphasize that MAEVI uses the most recent cost function parameters. These parameters are updated at each time period $t$ via the consensus matrix in line 35 of the MACCM algorithm. The convergence of the cost function parameters ensures the convergence of state-action value function 
and these are used in regret analysis of the MACCM algorithm
(Theorems \ref{thm: main_thm_regret_analysis} and \ref{thm: weight_convergence}).

\begin{algorithm}[h!]
\begin{algorithmic}[1]
    \STATE {\textbf{Input:}} Set of agents $S$; Confidence set $\mathcal{C}^k;\epsilon^k; \tb{\ti{w}}^k, ~\forall~k\in S$; discount term $q$.
    \STATE{\textbf{Initialize}:} Set ${Q}^{k,(0)}(
    \cdot, \cdot), {V}^{k,(0)}(\cdot) = 0, {V}^{k,(-1)}(\cdot) = \infty, ~\forall~k \in S$ 
    \FOR{$k \in S$}
    \IF{$\mathcal{C}^k \cap \mathcal{B} \neq \emptyset$}
    \WHILE{$\lvert\lvert {V}^{k,(l)}  - {V}^{k,(l-1)} \rvert\rvert_{\infty} \geq \epsilon^k$}
    \STATE ${Q}^{k,(l+1)}(\cdot,\cdot)$ = $\bar{c}(\cdot,\cdot;\tb{\ti{w}}^k)$ + $(1-q)$  $\min_{\boldsymbol{\theta} \in \mathcal{C}^k \cap \mathcal{B}}$
    $\left\langle \boldsymbol{\theta}, \phi_{{V}^{k,(l)}}(\cdot,\cdot)  \right\rangle$ 
    \STATE ${V}^{k,(l+1)}(\cdot) = \min_{\tb{\ti{a}} \in \mathcal{A}} {Q}^{k,(l+1)}(\cdot, \tb{\ti{a}}) $
    \ENDWHILE
    \STATE Set $l \leftarrow l+1$
    \ENDIF
    \ENDFOR
    \STATE Set ${Q}^k(\cdot, \cdot) \leftarrow {Q}^{k,(l+1)} (\cdot, \cdot),~ \forall k\in S$
    \STATE{\textbf{Output:}} ${Q}^k(\cdot, \cdot), ~\forall k \in S$
    \caption{Multi-Agent EVI routine}
    \label{alg: maevi}
\end{algorithmic}
\end{algorithm}

\section{MAIN RESULTS AND OUTLINE OF PROOFS}
\label{sec: main_results}
In this Section, we outline the main results. Due to space considerations, we defer all the proof details to the Appendix \ref{app: proofs}.
The following Theorem provides the bound on the regret $R_K$ given in Eq. \eqref{eqn: R_K} for the MACCM algorithm.

\begin{theorem}
	Under the Assumptions \ref{ass: existence_proper_policy}, \ref{ass: tpm_approx}, for any $\delta > 0$, let $\beta_t = B \sqrt{nd \log \left( \frac{4}{\delta} \left(n t^2 +  \frac{n t^3 B^2}{\lambda} \right) \right)} + \sqrt{\lambda nd }$, for all $t\geq 1$, where $B \geq B_{\star}$ and $\lambda \geq 1$. Then, with a probability of at least $1-\delta$, the regret of the MACCM algorithm satisfies 
	\begin{equation}
	\begin{split}
	R_K = \widetilde{\mathcal{O}} \bigg( B^{1.5} d \sqrt{nK/ c_{\min}} \cdot \log^2 \left(\frac{KBnd}{c_{\min}\delta} \right)  
	\\
	+ \frac{B^2 n d^2}{c_{\min}} \log^2 \left(\frac{KBnd}{c_{\min} \delta} \right) \bigg).
	\end{split}
	\end{equation}
	\label{thm: main_thm_regret_analysis}
\end{theorem}
If $B = O(B_{\star})$, then the regret is $\widetilde{O}(B_{\star}^{1.5}d \sqrt{nK/c_{\min}})$.
The proof of Theorem \ref{thm: main_thm_regret_analysis} includes three major steps: 1) convergence of cost function parameters (Theorem \ref{thm: weight_convergence}); 2) MAEVI analysis (Theorem \ref{thm: MAEVI_analysis}); 3) regret decomposition (Theorem \ref{thm: regret_decomposition}). The proof is available in Appendix \ref{app: main-thm-regret-analysis}

\textbf{Convergence of cost function parameters:} We next show the convergence of the cost function parameters $\tb{\ti{w}}^i$. To this end, let $d(\tb{\ti{s}})$ be the probability and stationary distribution of the Markov chain $\{\tb{\ti{s}}_t\}_{t\geq 0}$ under policy $\pi$, and $\pi(\tb{\ti{s}}, \tb{\ti{a}})$ be the probability of taking action $\tb{\ti{a}}$ in state $\tb{\ti{s}}$. Moreover, let $D^{\tb{\ti{s}}, \tb{\ti{a}}} = diag[d(\tb{\ti{s}}) \cdot \pi(\tb{\ti{s}}, \tb{\ti{a}}),~\tb{\ti{s}}\in \mc{S}, \tb{\ti{a}}\in \mc{A}]$ be the diagonal matrix with $d(\tb{\ti{s}}) \cdot \pi(\tb{\ti{s}}, \tb{\ti{a}})$ as diagonal entries. 
\begin{theorem}
	Under assumptions \ref{ass: comm_matrix} and \ref{ass: full_rank}, with sequence $\{\tb{\ti{w}}^i_t\}$, we have $\lim_t~ \tb{\ti{w}}^i_t = \tb{\ti{w}}^{\star}$ almost surely for each agent $i\in N$, where $ \tb{\ti{w}}^{\star}$ is unique solution to 
	\begin{equation}
	\Psi^{\top}D^{s,a}( \Psi\tb{\ti{w}}^{\star} - \bar{c}) = 0.
	\end{equation}
		\label{thm: weight_convergence}
\end{theorem}
The proof of this theorem uses the stochastic approximations of the single time-scale algorithms of \cite{borkarbook2edition}. The detailed proof is deferred to the Appendix \ref{app: weight-convergence}. 
The equation in above theorem is obtained by taking the first-order derivative of the least square minimization of the difference between the actual cost function and its linear parameterization.

\textbf{Multi-Agent EVI analysis:} Next, we show that the MAEVI algorithm converges in finite time to the optimistic estimator of the state-action value function. Specifically, we have the following theorem.
\begin{theorem}
\label{thm: MAEVI_analysis}
	Let $\beta_t = B \sqrt{nd \log \left( \frac{4}{\delta} \left(n t^2 +  \frac{n t^3 B^2}{\lambda} \right) \right)} + \sqrt{\lambda nd }$, for all $t\geq 1$. Then with probability at least $1-\delta/2$, and for each agent $i\in N$, for all $j_i \geq 1$, MAEVI converges in finite time, and the following hold: $\boldsymbol{\theta}^{\star} \in \mathcal{C}^i_{j_i} \cap \mathcal{B}$, $0 \leq Q^i_{j_i} (\cdot, \cdot) \leq Q^{i\star}(\cdot,\cdot; \tb{\ti{w}}^i)$, and $0 \leq V^i_{j_i} (\cdot) \leq V^{i\star}(\cdot; \tb{\ti{w}}^i)$.
\end{theorem}
The proof of this theorem is deferred to Appendix \ref{app: evi_analysis} due to space considerations. 

\textbf{Regret Decomposition:} Next, we give the details of the regret decomposition. To this end, we first show that the total number of calls $J$ to the MAEVI algorithm in the entire analysis is bounded. Let $J^i$ be the total number of calls to the MAEVI algorithm made by agent $i$. Note that $J\leq \sum_{i\in N} J^i$. An agent $i\in N$ makes a call to the MAEVI algorithm if either the determinant doubling criteria or the time doubling is satisfied. Let $J^i_1$ be the number of calls to MAEVI made via determinant doubling criteria, and $J^i_2$ be the calls to the EVI algorithm via the time doubling criteria. Therefore, $J^i \leq J^i_1 + J^i_2$. 
\begin{lemma}
	\label{lemma: calls_to_evi}
	The total number of calls to the MAEVI algorithm in the entire analysis, $J$, is bounded as
	\begin{equation}
	J \leq 2 n^2d \log \left( 1 + \frac{T B_{\star}^2 nd}{\lambda} \right) + 2n \log(T).
	\end{equation}
\end{lemma}
The proof of the above lemma is deferred to the Appendix \ref{app: number_calls_evi}. For the regret decomposition, we divide the time horizon into disjoint intervals $m=1, 2, \dots, M$. The endpoint of an interval is decided by one of the two conditions: 1) MAEVI is triggered for at least one agent, and 2) all the agents have reached the goal node. This decomposition is explicitly used in the regret analysis only and not in the actual algorithm implementation. It is easy to observe that the interval length (number of periods in the interval) can vary; let $H_m$ denote the length of the interval $m$. Moreover, at the end of the $M$th interval, all the  $K$ episodes are over. Therefore, the total length of all the intervals is $\sum_{m=1}^M H_m$, which is the same as $\sum_{k=1}^K T_k$, where $T_k$ is the time to finish the episode $k$. Hence, both representations reflect the total time $T$ to finish all the $K$ episodes. Using the above interval decomposition, we write the regret $R_K$ as 
\begin{equation}
\label{eqn: regret-eqn-temp}
\begin{split}
R_K = R(M) \leq \sum_{m=1}^M \sum_{h=1}^{H_m}  \frac{1}{n} \sum_{i=1}^n \bar{c}(\tb{\ti{s}}_{m,h}, \tb{\ti{a}}_{m,h}, \tb{\ti{w}}^i)
\\
+ 1 - \sum_{m\in \mathcal{M}(M)} \frac{1}{n} \sum_{i=1}^n V^i_{j_i(m)}(\tb{\ti{s}}_{init}),
\end{split}
\end{equation}
here $\mathcal{M}(M)$ is the set of all intervals that are the first intervals in each episode. In RHS we add 1 because  $|V^i_0| \leq 1$. In the Theorem below, we decompose the regret $R_K$ as 
\begin{theorem}
	\label{thm: regret_decomposition}
	Assume that the event in Theorem \ref{thm: MAEVI_analysis} holds, then we have the following upper bound on the regret
	\begin{equation}
	\begin{aligned}
	R(M)  &\leq E_1 +  E_2
	+2n^2dB_{\star} \log \left(1+ \frac{TB_{\star}^2 nd}{\lambda} \right)
	\\
	& ~~~~ + 2n B_{\star} \log(T) + 2
	\end{aligned}
	\end{equation}
	where $E_1$ and $E_2$ are defined as
	\begin{align*}
	E_1 &= \sum_{m=1}^M \sum_{h=1}^{H_m} \bigg[ \frac{1}{n} \sum_{i=1}^n \bigg\lbrace \bar{c}(\tb{\ti{s}}_{m,h}, \tb{\ti{a}}_{m,h}, \tb{\ti{w}}^i)
	\\
	& ~~~~~~~ +  \mathbb{P}V^i_{j_i(m)}(\tb{\ti{s}}_{m,h}, \tb{\ti{a}}_{m,h}) -  V^i_{j_i(m)}(\tb{\ti{s}}_{m,h}) \bigg\rbrace \bigg],
	\\
	E_2 &= \sum_{m=1}^M \sum_{h=1}^{H_m} \bigg[ \frac{1}{n} \sum_{i=1}^n V^i_{j_i(m)}(\tb{\ti{s}}_{m,h+1}) 
	\\
	& ~~~~~~~~~~~~~~~~~~~~~~~ - \frac{1}{n} \sum_{i=1}^n \mathbb{P}V^i_{j_i(m)}(\tb{\ti{s}}_{m,h}, \tb{\ti{a}}_{m,h}) \bigg].
	\end{align*}
\label{eqn: regret_decomposition}
\end{theorem}
The proof of this theorem is deferred to the Appendix \ref{app: regret-decomposition}.  
To complete the proof of Theorem \ref{thm: main_thm_regret_analysis}, we bound $E_1$ and $E_2$ separately (Appendix \ref{app: bound_E1}, \ref{app: bound_E2}). Bounding $E_1$ uses all the intrinsic properties of MACCM algorithm \ref{algo: main_algo_multi_agent}. Unlike the single-agent setup, here, we separately consider the set of all agents for whom the MAEVI is triggered or not. Thus, $E_1$ is decomposed into $E_1(S_m)$ and $E_1(S^c_m)$ where $S_m$ is the set of agents for whom MAEVI is triggered in $m$-th interval, and $S^c_m$ are remaining agents. The bounds on $E_1(S_m)$ and $E_1(S^c_m)$ are specifically required in our multi-agent setup and are novel.
Moreover, $E_2$ is the martingale difference sum, so it is bounded using the  concentration inequalities.

\section{COMPUTATIONAL EXPERIMENTS}
\label{sec: computations}
In this Section, we provide the details of the computations to validate the usefulness of our MACCM algorithm. Consider a network with two nodes $\{s_{init}, g\}$. Thus, the number of states is $2^n$, where $n$ is the number of agents.
In each state the actions available to each agent are $\mathcal{A}^i = \{-1, 1\}^{d-1}$, for some given $d\geq 2$. So, the total number of actions is $2^{nd}$. 
Since the number of states and actions is exponentially large, we parameterize the transition probability. For each $(\tb{\ti{s}}^{\prime}, \tb{\ti{a}}, \tb{\ti{s}}) \in \mathcal{S} \times \mathcal{A} \times \mathcal{S}$, the global transition probability is parameterized as $\mathbb{P}_{\boldsymbol{\theta}}(\tb{\ti{s}}^{\prime} | \tb{\ti{s}}, \tb{\ti{a}}) = \langle \phi(\tb{\ti{s}}^{\prime} | \tb{\ti{s}},\tb{\ti{a}}), {\boldsymbol{\theta}} \rangle$. The features $\phi(\tb{\ti{s}}^{\prime} | \tb{\ti{s}},\tb{\ti{a}})$ are described below. 
\begin{equation*}
\label{eqn: features_global}
\begin{cases} (\phi(s^{\prime^1} | s^1, \tb{\ti{a}}^1), \dots, \phi(s^{\prime^n} | s^n, \tb{\ti{a}}^n)), & if ~\tb{\ti{s}} \neq \tb{\ti{g}},
\\
\mb{0}_{nd}, & if ~ \tb{\ti{s}} = \tb{\ti{g}},~ \tb{\ti{s}}^{\prime} \neq \tb{\ti{g}},
\\
(\mb{0}_{nd-1}, 2^{n-1}), & if ~ \tb{\ti{s}} = \tb{\ti{g}},~ \tb{\ti{s}}^{\prime}  = \tb{\ti{g}},
\end{cases}
\end{equation*}
where $\phi(s^{\prime^{i}} | s^i, \tb{\ti{a}}^i)$ is defined as 
\begin{equation*}
\phi(s^{\prime^{i}} | s^i, \tb{\ti{a}}^i) = \begin{cases}  
\left(-\tb{\ti{a}}^i, \frac{1-\delta}{n} \right)^{\top}, & if ~ {s}^i = s^{\prime^{i}} = s_{init}
\\
\left(\tb{\ti{a}}^i, \frac{\delta}{n} \right)^{\top}, & if ~ {s}^i = s_{init},~s^{\prime^{i}} = g
\\
\mathbf{0}_{d}^{\top}, & if ~ {s}^i = g,~s^{\prime^{i}} = s_{init}
\\
\left(\mathbf{0}_{d-1}, \frac{1}{n} \right)^{\top}, & if ~ {s}^i = g,~ s^{\prime^{i}} = g.
\end{cases}
\end{equation*}
Here $\mathbf{0}_d^{\top} = (0,0, \dots, 0)^{\top}$ is a vector of $d$ dimension with all zeros.
Thus, the features $\phi(s^{\prime^{i}} | s^i, \tb{\ti{a}}^i) \in \mathbb{R}^{nd}$. Moreover, the transition probability parameters are taken as $\boldsymbol{\theta} = \left( \boldsymbol{\theta}^1, \frac{1}{2^{n-1}},  \boldsymbol{\theta}^2, \frac{1}{2^{n-1}} \dots, \boldsymbol{\theta}^n, \frac{1}{2^{n-1}} \right)$ where $\boldsymbol{\theta}^i \in \left\lbrace -\frac{\Delta}{n(d-1)}, \frac{\Delta}{n(d-1)}  \right\rbrace^{d-1} $, and $\Delta < \delta$. 
\begin{lemma}
	\label{lemma: feature_properties}
	The features $\phi(\tb{\ti{s}}^{\prime} | \tb{\ti{s}},\tb{\ti{a}})$ satisfy the following: (a) $\sum_{\tb{\ti{s}}^{\prime}} \langle \phi(\tb{\ti{s}}^{\prime} | \tb{\ti{s}}, \tb{\ti{a}}), \boldsymbol{\theta}  \rangle = 1, ~ \forall ~\tb{\ti{s}}, \tb{\ti{a}}$; (b) $\langle \phi(\tb{\ti{s}}^{\prime} = \tb{\ti{g}} | \tb{\ti{s}} = \tb{\ti{g}}, \tb{\ti{a}}), \boldsymbol{\theta}  \rangle = 1, ~\forall~\tb{\ti{a}}$; (c) $\langle \phi(\tb{\ti{s}}^{\prime} \neq \tb{\ti{g}} | \tb{\ti{s}} = \tb{\ti{g}}, \tb{\ti{a}}), \boldsymbol{\theta}  \rangle = 0,~ \forall ~\tb{\ti{a}}$. 
\end{lemma}
The proof is deferred to the Appendix \ref{app: feature_properties}. Recall, from Assumption \ref{ass: full_rank}, $\bar{c}(\tb{\ti{s}}, \tb{\ti{a}}; \tb{\ti{w}}) = \langle \psi(\tb{\ti{s}}, \tb{\ti{a}}), \tb{\ti{w}} \rangle$. We take the features as $\psi(\tb{\ti{s}}, \tb{\ti{a}}) = (\psi(s^1, \tb{\ti{a}}^1), \psi(s^2, \tb{\ti{a}}^2), \dots, \psi(s^n, \tb{\ti{a}}^n))$, where  $\psi(s_{init}, \tb{\ti{a}}^i) = \sum_{j=1}^n \mathds{1}_{\{\tb{\ti{a}}^i = \tb{\ti{a}}^j | s^j = s_{init}\}}$, and $\psi(g, \tb{\ti{a}}^i) = 0$ for any $\tb{\ti{a}}^i \in \mathcal{A}^i$. That is the feature $\psi(s_{init}, \tb{\ti{a}}^i)$ captures the congestion realized by any agent $i$ present at $s_{init}$. For each agent $i\in N$ and for each $\tb{\ti{s}}, \tb{\ti{a}}$ pair the private component  $K^i(\tb{\ti{s}}, \tb{\ti{a}}) \sim Uniform(c_{\min},1)$. 
Further, each entry of the consensus matrix is taken as $1/n$, where $n$ is the number of agents.

The intent of this small 2 node network is 3 fold: \textbf{(1)} This 2 node network is common in RL literature \cite{min2022learning}, as it depicts the worst-case performance in the form of a lower bound on the regret. \textbf{(2)} Though the network seems very small, the number of states is $2^n$, and both the number of actions and model parameters are $2^{nd}$, i.e., exponential in the number of agents, and the feature dimension.
So, the model complexity increases exponentially in the number of agents offering a computationally challenging model. \textbf{(3)} The major computational head in our algorithm is due to the MAEVI sub-routine; in each step, we solve a computationally challenging discrete combinatorial optimization problem in model parameters. Given these constraints, we consider a small network with a low number of agents;  however, our algorithm achieves sub-linear regret even in this hard instance. 
For a general network with any number of agents, a separate feature design and a suitable model parameters choice can make the MAEVI algorithm easy; such a feature design in itself is a complex problem.

Next, we compute the value of the optimal policy for the above model. We require it in the regret computations. First, note that the value of the optimal policy is defined as the sum of expected costs if all agents reach the goal state exactly in $1$ time period, in 2 time period, and so on. Let $x_j$ be the number of agents move to the goal node at each period $j=1,2, \dots, t-1$, and remaining $x_t = n-\sum_{j=1}^{t-1} x_j$ agents move to the goal node by $t$-th period. We call this sequence of departures of agents to the goal node as the `departure sequence'. For the above departure sequence, the cost incurred is $C_{\alpha}(x_1, \dots, x_{t-1}, x_t) =$
\begin{equation}
\label{eqn: C_t_steps}
\alpha  \sum_{j=1}^{t-1} x_j^2 + \sum_{j=1}^{t-1} \Big( n - \sum_{i = 1}^{j} x_i \Big) \cdot c_{\min} + \alpha  x_t^2,
\end{equation}
where $\alpha$ is the mean of the uniform distribution $\mathcal{U}(c_{\min}, 1)$, i.e., $\alpha = \frac{c_{\min} + 1 }{2}$. We use this $\alpha$ instead of the private cost to compute the optimal value. The first term in the above equation is because all $x_j$ agents have moved to the goal node in time period $j$; hence the congestion is $x_j$. Moreover, each agent incurs the private cost $\alpha$ in this period. 
So, the cost incurred to any agent as per Eq. \eqref{eqn: agent_i_cost} is $\alpha x_j$, and the number of agents moved are $x_j$, and hence the total cost incurred is $(\alpha x_j)\times  x_j = \alpha x_j^2$. This happens for all the time periods $j =1, 2, \dots, t-1$, so we sum this for $(t-1)$ periods. The remaining agents at each time period incurred a waiting cost of $c_{\min}$; thus, we have a second term. Finally, the third term is because the remaining agents move to the goal node at the last period $t$.
So, the optimal value, $V^{\star}$ is
\begin{equation}
\label{eqn: v_pi_star_expression}
V^{\star} = \sum_{t=1}^{\infty} \mathbb{P}[x_1^{\star}, \dots,x_t^{\star}] \cdot C_{\alpha}(x_1^{\star}, \dots, x_t^{\star})
\end{equation}
where $x_1^{\star}, \dots, x_t^{\star}$ are the optimal departure sequence. These $x_j^{\star}$ are obtained by minimizing the cost function $C_{\alpha}(x_1, \dots,  x_t)$  in Eq. \eqref{eqn: C_t_steps}. Moreover, $\mathbb{P}[x_1^{\star}, \dots, x_{t}^{\star}]$ is the probability of occurrence of this optimal departure sequence. Note that, unlike the regret defined in Eq. \eqref{eqn: R_K} that use the cost function parameters, here we take the value of the optimal policy defined above.
The following Theorem provides $x_1^{\star}, \dots, x_{t}^{\star}$ and its  cost $C_{\alpha}(x_1^{\star}, \dots, x_{t}^{\star})$. 

\begin{theorem}
	\label{thm: optimal_x_j_star}
	The optimal departure sequence is given by 
	\begin{align}
	x_j^{\star}  &= \left\lfloor \frac{n}{t} + \left(\frac{t + 1}{2} - j  \right) \cdot \frac{c_{\min}}{2 \alpha} \right\rfloor, ~\forall ~j\in[t-1], \nonumber
	\\
	x_t^{\star} &= n - \sum_{j=1}^{t-1} x_j^{\star} 
	\end{align}
	The cost 
	$C_{\alpha}(x_1^{\star}, x_2^{\star}, \dots, x_{t}^{\star}) $ of using the above optimal departure sequence is
	\begin{equation}
	\label{eqn: optimal_cost_computation}
	\alpha t \left(\frac{n}{t}\right)^2 + n(t-1) \cdot \frac{c_{\min}}{2 \alpha} - \frac{t(t-1)(t+1)}{12} \cdot \frac{c_{\min}^2}{4\alpha^2}.
	\end{equation}
\end{theorem}
The proof of above theorem is deferred to the Appendix \ref{app: optimal_x_j_star}. The above cost captures the trade-off between the minimum cost to an agent for staying at the initial node and the congestion cost. In particular, the first term captures all agents' total cost of going to the goal node. The remaining terms capture the minimum cost of staying at $s_{init}$.

Apart from the cost of the optimal departure sequence, we also require the probability of this departure sequence to compute $V^{\star}$. To this end, we recall the feature design of transition probability that allows an agent to stay or depart from the initial node $s_{init}$. Note that agent $i$ stays or departs from initial state iff the sign of the action $\tb{\ti{a}}^i$ matches the sign of the transition probability function parameter $\boldsymbol{\theta}^i$, i.e,  $sgn(\tb{\ti{a}}^i_j) = sgn(\boldsymbol{\theta}^i_j)$ for all $j=1, 2, \dots, d-1$ (more details are available in SM).  Using this sign matching property, we have the following theorem for the transition probability.
\begin{theorem}
	The transition probability $\mathbb{P}[x_1^{\star}, \dots, x_{t}^{\star}]$ is given by
	\begin{equation}
	\label{eqn: prob_dept_sequence}
	\prod_{k=1}^{t-1}  \big( 1 -  \gamma n + (\gamma -\eta) \sum_{j=1}^{k} x_{j}^{\star} \big) \times \big( \gamma n- (\gamma - \eta) \sum_{j=1}^{t-1} x_{j}^{\star} \big),
	\end{equation}
	where $\gamma = \left(\frac{\Delta}{n} + \frac{\delta}{n \cdot 2^{n-1}} \right) $ and $\eta = \frac{1}{n\cdot 2^{n-1}}$.
\label{thm: transition_prob_optimal}
\end{theorem} 
The proof of this Theorem is deferred to the Appendix \ref{app: tranisition_prob_optimal}. Using the above transition probability along with cost in Eq. \eqref{eqn: optimal_cost_computation} get $V^{\star}$. However, it is hard to get the closed form expression of $V^{\star}$, so we use its approximate value for regret computation. The approximate optimal value $V^{\star}_{T}$ in terms of a given large $T < \infty$ is
\begin{equation}
\label{eqn: v_pi_star_expression_estimate}
V^{\star}_{T} = \sum_{t=1}^{
	T} \mathbb{P}[x_1^{\star}, \dots, x_{t}^{\star}] \cdot C_{\alpha}(x_1^{\star}, \dots, x_{t}^{\star}),
\end{equation}
To obtain a better approximation of the optimal value $V^{\star}$, we tune $T$. 
\begin{figure}[h!]
	\centering
	\includegraphics[scale = 0.45]{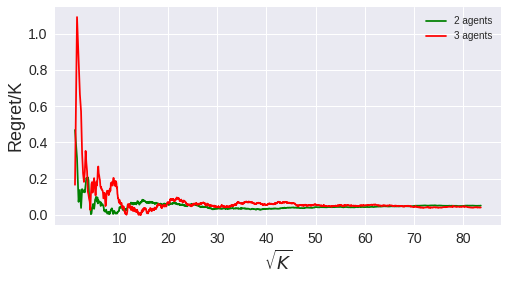}
	\caption{Average regret for $n=2$ agents in green and $n=3$ agents in red. Here $d = 2, \delta = 0.1, \Delta = 0.2, K = 7000$.  ${V}^{\star}_{T} = 2.15$ and $4.365$ for 2 and 3 agents respectively. All the values are averaged over 15 runs.}
	\label{fig: avg_reg_2_agents}
\end{figure} 
Moreover, note that we are approximating the private component of the cost by  $\alpha$. Thus, the above $V^{\star}_{T}$ will have some error; we minimize it in our computations by running the MAACM algorithm for multiple runs and average the regret over these runs. The average regret for $2$ and $3$ agents are very close to zero, as  shown in Figure \ref{fig: avg_reg_2_agents}. 

\begin{remark}
	\label{rem: single_agent_reduction}
	Suppose $n=1$ and $c(s,a) = c_{\min} = 1$ for all $s,a$. Then, the optimal departure sequence is $x_1^{\star} = \dots = x_{t-1}^{\star} = 0, x_t^{\star} = 1$. So, from Eq. \eqref{eqn: C_t_steps}, the optimal cost is $t$. Moreover, the probability in Eq. \eqref{eqn: prob_dept_sequence} reduces to $(1-\Delta - \delta)^{t-1} (\Delta + \delta)$ and hence $V^{\star} = \sum_{t=1}^{\infty} (1-\Delta - \delta)^{t-1} (\Delta + \delta) \cdot t = \frac{1}{\delta + \Delta}$. Thus, we recover \cite{min2022learning} results for 2  node network with 1 agent (and hence with no congestion cost). Also, the features used in \cite{min2022learning} and \cite{vial2022regret} are interchangeable. So, we have more general results with extra complexity regarding congestion and privacy.
\end{remark}

\section{RELATED WORK}
\label{sec: related_work}
The single agent SSP is well known for decades \citep{bertsekas2012dynamic}. However, these assume the knowledge of the transition probabilities and the cost of each edge.
Recently, there has been much work on the online SSP problem when the transition or the cost is not known or random. In such cases, the RL based algorithms are proposed \citep{min2022learning,vial2022regret,tarbouriech2020no,tarbouriech2021stochastic}. Many instances of SSP are run over multiple episodes using these algorithm, and the regret over $K$ episodes of the SSP is defined.

The online SSP problem is first described in \cite{tarbouriech2020no} with $\widetilde{O}(K^{2/3})$ regret. Later this is improved in \cite{rosenberg2020near}. They gave a upper bound of $\widetilde{O}(B_{\star} |S| \sqrt{|A|K})$, and a lower bound of $\Omega(B_{\star}  \sqrt{|S||A|K})$ where $B_{\star}$ is an upper bound on the expected cost of the optimal policy.  However, they assume that the cost functions are known and deterministic. Authors in \cite{cohen2021minimax} assume the cost function to be i.i.d. and initially unknown. They give an upper and a lower bound proving the optimal regret of $\widetilde{\Theta}(\sqrt{(B_{\star}^2 + B_{\star}) |S||A| K)}$ 
The algorithms proposed by \cite{rosenberg2020near} and \cite{tarbouriech2020no} uses the ``optimism in the face of uncertainty (OFU)'' principle that in-turn uses the ideas from UCRL2 algorithm of \cite{jaksch2010near} for average reward MDPs. \cite{cohen2021minimax} uses a black-box reduction of SSP to finite horizon MDPs. Similar reduction is used in \cite{chen2021finding,chen2021minimax} for SSP with adversarially changing costs. \cite{cohen2021minimax} gave a new algorithm named ULCVI for regret minimization in finite-horizon MDPs. \cite{tarbouriech2021stochastic} extended the work of \cite{cohen2021minimax} to obtain a comparable regret bound for SSP without prior knowledge of expected time to reach the goal state.

Often the cost and the transition probabilities are unknown, and the state and action space is humongous. To this end, many researchers use function approximations of the transition probabilities, the per-period cost, or both \citep{wang2020optimism,yang2020reinforcement}. Some recent works in this direction are \cite{jia2020model,min2022learning,jin2020provably,zhou2021nearly}. 
In particular, \cite{min2022learning} proposes a LEVIS algorithm that uses the optimistic update of the estimated $Q$ function using the extended value iteration (EVI) algorithm.
Unlike \cite{min2022learning}, authors in \cite{vial2022regret} uses OFU principle and parameterize the cost function also. Moreover, the feature design of the transition probabilities is also different; however, the features in these works are interchangeable. The regrets obtained in these works depend $K$, $B_{\star}$, and $c_{\min}$. In our work, we simultaneously use the linear function approximations of the transition probabilities and the global cost function. We also provide a decentralized multi-agent algorithm incorporating congestion cost and agents' privacy of cost.

\section{DISCUSSION}
\label{sec: discussion}
This work considers a multi-agent variant of the optimal path-finding problem on a given network with pre-specified initial and goal nodes.
The cost of traversing a link of the network depends on the private cost of the agent (capturing the agent's travel efficiency) and the congestion on the link. The unknown transition probability is the linear function of a given basis function. Moreover, each agent maintains an estimate of global cost function parameters, that are shared among the agents via a communication matrix. 

We propose a fully decentralized multi-agent congestion cost minimization (MACCM) algorithm that achieves a sub-linear regret.
In every episode of the MACCM algorithm, each agent maintains an optimistic estimate of the state-action value function; this estimate is updated according to a MAEVI sub-routine. The update happens if the `doubling criteria' is triggered for any agent.
To our knowledge, this is the first work that considers the multi-agent version of the congestion cost minimization problem over a network with linear function approximations. It has broader applicability in many real-life scenarios, such as decentralized fleet management. Experiments for 2 and 3 agent cases on a network validate our results.

The work we consider possesses many challenges and future directions, and we mention some of these here. The current algorithm is based on the optimistic state-action value function updated according to doubling criteria.
A better optimistic estimator with tighter regret bounds can be tried. For example, one can use a parameterized policy space and incorporate the feedback from the policy parameter in the state-action value function. For large networks we can explore a distinct feature design and suitable model parameters to address the scalability of computations. Further a lower bound on the regret of our multi-agent congestion cost minimization, MACCM, algorithm is desirable.

\subsubsection*{Acknowledgements}
We would like to thank the anonymous Reviewers and Meta Reviewer for their useful comments and suggestions. While working at this problem Prashant Trivedi was partially supported by the Teaching Assistantship offered by Government of India. Some part of this work was done when Nandyala Hemachandra was visiting IIM Bangalore on a sabbatical leave.

\subsubsection*{Societal Impact} 
This work can only have a positive societal impact  because it can be a good data-driven (RL) decentralised model for fleet management in transport.

\subsubsection*{Code Release}
Some details of the code for the MACCM algorithm implementations are available at \url{https://github.com/PRA1TRIVEDI/MACCM}. Refer to the Readme.md file for the description about how to run the code.

\bibliography{References}
\bibliographystyle{apalike}

\appendix

\section*{APPENDIX}

In this appendix we give the details of proofs omitted in the main results in Section \ref{app: proofs}; further details of the feature design and the computations in Section \ref{app: details-computations}; proofs of the intermediate lemmas and propositions in Section \ref{app: inter-lemmas}; and some useful results that we use from the existing literature in Section \ref{app: additional-useful-results}. 

For better readability, we first reiterate the result, and then give its proof. 

\section{PROOFS AND DETAILS OF THE MAIN RESULTS}
\label{app: proofs}

First, we give proofs of the results we omitted in the main paper.

\subsection{Proof of Proposition \ref{prop: equiv-opti-problem}}
\label{app: equiv-opti-problem}
We first provide the equivalence of the optimization problems \eqref{eqn: op} and \eqref{eqn: op_equivalent} obtain from the least square minimizer of the global cost function.  Recall the optimization problem \eqref{eqn: op} is
\begin{equation}
\tag{OP 1}
\min_{\tb{\ti{w}}}~~ \mathbb{E}_{\tb{\ti{s}},\tb{\ti{a}}} [\bar{c}(\tb{\ti{s}},\tb{\ti{a}}) - \bar{c}(\tb{\ti{s}},\tb{\ti{a}};\tb{\ti{w}})]^2.
\end{equation}

Recall the proposition: The optimization problem in \eqref{eqn: op} is equivalently characterized as (both have the same stationary points)
	\begin{equation*}
	\tag{OP 2}
	\min_{\tb{\ti{w}}} \sum_{i=1}^n \mathbb{E}_{\tb{\ti{s}},\tb{\ti{a}}} [c^i(\tb{\ti{s}},\tb{\ti{a}}) - \bar{c}(\tb{\ti{s}},\tb{\ti{a}};\tb{\ti{w}})]^2.
	\end{equation*}

\begin{proof}
	Taking the first order derivative of the objective function in optimization problem \eqref{eqn: op} w.r.t. $\tb{\ti{w}}$, we have: 
	\begin{eqnarray*}
		-2 \times \mathbb{E}_{\tb{\ti{s}},\tb{\ti{a}}} [ \bar{c}(\tb{\ti{s}},\tb{\ti{a}}) - \bar{c}(\tb{\ti{s}},\tb{\ti{a}}; \tb{\ti{w}})] \times \nabla_{\tb{\ti{w}}} \bar{c}(\tb{\ti{s}},\tb{\ti{a}}; \tb{\ti{w}}),
		&=& -2 \times \mathbb{E}_{\tb{\ti{s}},\tb{\ti{a}}} \left[ \frac{1}{n}\sum_{i\in {N}} c^i(\tb{\ti{s}},\tb{\ti{a}}) - \bar{c}(\tb{\ti{s}},\tb{\ti{a}}; \tb{\ti{w}})\right] \times \nabla_{\tb{\ti{w}}} \bar{c}(\tb{\ti{s}},\tb{\ti{a}}; \tb{\ti{w}}),
		\\
		&=& -\frac{2}{n} \times \mathbb{E}_{\tb{\ti{s}},\tb{\ti{a}}} \left[ \sum_{i\in {N}} c^i(\tb{\ti{s}},\tb{\ti{a}}) - n\cdot\bar{c}(\tb{\ti{s}},\tb{\ti{a}}; \tb{\ti{w}})\right] \times \nabla_{\tb{\ti{w}}} \bar{c}(\tb{\ti{s}},\tb{\ti{a}}; \tb{\ti{w}}),
		\\
		&=& -\frac{2}{n} \times \mathbb{E}_{\tb{\ti{s}},\tb{\ti{a}}} \left[ \sum_{i\in {N}} \left( c^i(\tb{\ti{s}},\tb{\ti{a}}) - \bar{c}(\tb{\ti{s}},\tb{\ti{a}}; \tb{\ti{w}})\right)\right] \times \nabla_{\tb{\ti{w}}} \bar{c}(\tb{\ti{s}},\tb{\ti{a}}; \tb{\ti{w}}).
	\end{eqnarray*}
	Ignoring the factor $\frac{1}{n}$ in the above equation, we exactly have the first order derivative of the objective function in \eqref{eqn: op_equivalent}. Thus, both optimization problems have the same stationary points. Hence, \eqref{eqn: op} is an \textit{equivalent characterization} of the optimization problem  \eqref{eqn: op_equivalent}.
\end{proof}

\subsection{Proof of Theorem \ref{thm: main_thm_regret_analysis}}
\label{app: main-thm-regret-analysis}
In this section, we give the proof of our main result (Theorem \ref{thm: main_thm_regret_analysis}). It provides the upper bound on the regret of our MACCM algorithm. The proof of this theorem relies on an intermediate Lemma \ref{lemma: interim_regret_thm_regret} (given below).

Recall the theorem: Under the Assumptions \ref{ass: existence_proper_policy}, \ref{ass: tpm_approx}, for any $\delta > 0$, let $\beta_t = B \sqrt{nd \log \left( \frac{4}{\delta} \left(n t^2 +  \frac{n t^3 B^2}{\lambda} \right) \right)} + \sqrt{\lambda nd }$, for all $t\geq 1$, where $B \geq B_{\star}$ and $\lambda \geq 1$. Then, with probability at least $1-\delta$, the regret of the MACCM algorithm satisfies 
	\begin{equation}
	R_K = \mathcal{O} \bigg( B^{1.5} d \sqrt{nK/ c_{\min}} \cdot \log^2 \left(\frac{KBnd}{c_{\min}\delta} \right)  
	+ \frac{B^2 n d^2}{c_{\min}} \log^2 \left(\frac{KBnd}{c_{\min} \delta} \right) \bigg)
	\label{eqn: regret_expression_main_thm}
	\end{equation}

\begin{proof}
	Note that the total cost incurred by each agent in $K$ episodes is upper bounded by $R_K + KB_{\star}$ and it is lower bounded by  $c_{\min} \cdot n \cdot T$. This provides the relation between a fixed quantity $K$ and the random quantity $T$. To complete the proof we use the following lemma.
\begin{lemma}
	\label{lemma: interim_regret_thm_regret}
	Under Assumptions \ref{ass: existence_proper_policy} and \ref{ass: tpm_approx}, for any $\delta>0$, let $\beta_t = B \sqrt{nd \log \left( \frac{4}{\delta} \left(n t^2 +  \frac{n t^3 B^2}{\lambda} \right) \right)} + \sqrt{\lambda nd }$ for all $t\geq 0$, where  $B \geq B_{\star}$ where $\lambda \geq 1$. Then with the probability of at least $1-\delta$, the regret of the MACCM algorithm satisfies 
	\begin{equation*}
	R_K \leq10 \beta_T \sqrt{T d \log \left( 1 +\frac{ T B_{\star}^2}{\lambda} \right) } 
	+ 16n^2dB_{\star} \log \left(T+ \frac{T^2B_{\star}^2 nd}{\lambda} \right),
	\end{equation*}
	where $T$ is the total number of time periods.
\end{lemma}
The proof of this lemma is given in Appendix \ref{sec: regret-proposition}. Using the above lemma, with probability at least $1-\delta$, we have
	\begin{equation*}
	c_{\min} \cdot n \cdot T \leq 10 \beta_T \sqrt{T d \log \left( 1 +\frac{ T B_{\star}^2}{\lambda} \right) } 
	+ 16n^2dB_{\star} \log \left(T+ \frac{T^2B_{\star}^2 nd}{\lambda} \right)+  K B_{\star}.
	\end{equation*}
	Solving the above equation for $T$, we have 
	\begin{equation*}
	T = \mathcal{O} \left( \log^2 \left( \frac{n}{\delta} \right) \cdot \left( \frac{K B_{\star}}{n c_{min}} + \frac{B^2_{\star} d^2}{c_{\min}^2} \right) \right).
	\end{equation*}
	Plugging this back into Lemma \ref{lemma: interim_regret_thm_regret}, we have the desired result of Theorem \ref{thm: main_thm_regret_analysis}. 
\end{proof}

To complete the proof of above theorem, we need to prove the above Lemma \ref{lemma: interim_regret_thm_regret}. To this end, we require the convergence of the cost function parameters as stated in Theorem \ref{thm: weight_convergence}. Let $d(\tb{\ti{s}})$ be the probability and stationary distribution of the Markov chain $\{\tb{\ti{s}}_t\}_{t\geq 0}$ under policy $\pi$, and $\pi(\tb{\ti{s}}, \tb{\ti{a}})$ be the probability of taking action $\tb{\ti{a}}$ is state $\tb{\ti{s}}$. Moreover, let $D^{\tb{\ti{s}}, \tb{\ti{a}}} = diag[d(\tb{\ti{s}}) \cdot \pi(\tb{\ti{s}}, \tb{\ti{a}})]$ be the diagonal matrix with $d(\tb{\ti{s}}) \cdot \pi(\tb{\ti{s}}, \tb{\ti{a}})$ as diagonal elements. 
	
\subsection{Proof of Theorem \ref{thm: weight_convergence}}
\label{app: weight-convergence}
Recall the theorem: Under the Assumptions \ref{ass: comm_matrix} and \ref{ass: full_rank}, with sequence $\{\tb{\ti{w}}^i_t\}$, we have $\lim_t~ \tb{\ti{w}}^i_t = \tb{\ti{w}}^{\star}$ a.s. for each agent $i\in N$, where
		$ \tb{\ti{w}}^{\star}$ is unique solution to 
		\begin{equation*}
		\Psi^{\top}D^{s,a}( \Psi\tb{\ti{w}}^{\star} - \bar{c}) = 0.
		\end{equation*}
	\begin{proof}
		\label{proof: critic_convergence_fisher}
		To prove the convergence of the cost function parameters, we use the following proposition to give bounds on $\tb{\ti{w}}^i_t$ for all $i\in N$. For proof, we refer to \cite{zhang2018fully}.
		
		\begin{prop}
			\label{prop: w_boundedness}
			Under Assumptions \ref{ass: comm_matrix}, and \ref{ass: full_rank} the sequence $\{\tb{\ti{w}}^i_t\}$ satisfy $sup_t~||\tb{\ti{w}}^i_t||<\infty$ a.s., for all $i\in {N}$.
		\end{prop}
		Let $\mathcal{F}_{t} = \sigma(c_{\tau}, \tb{\ti{w}}_{\tau}, \tb{\ti{s}}_{\tau},  \tb{\ti{a}}_{\tau}, L_{\tau - 1}, \tau \leq t)$ be the filtration which is an increasing $\sigma$-algebra over time $t$. Define the following for notation convenience. Let $c_t = [c^1_t, \dots, c^n_t]^{\top} \in \mathbb{R}^n$ , and $\tb{\ti{w}}_t = [(\tb{\ti{w}}^1_t)^{\top}, \dots, (\tb{\ti{w}}^n_t)^{\top}]^{\top} \in \mathbb{R}^{nk}$. Moreover, let $A \otimes B$ represent the Kronecker product of any two matrices $A$ and $B$. Let $y_t = [(y^1_t)^{\top}, \dots, (y^n_t)^{\top}]^{\top}$, where $y^i_{t+1} = [(c^i_{t+1} - \psi_t^{\top}\tb{\ti{w}}^i_t)\psi_t^{\top}]^{\top}$. Recall, $\psi_t = \psi(\tb{\ti{s}}_t,\tb{\ti{a}}_t)$. Let $I$ be the identity matrix of the dimension $k \times k$. Then update of $\tb{\ti{w}}_t$ can be written as 
		\begin{equation}
		\label{eqn: z_recursion}
		\tb{\ti{w}}_{t+1} = (L_t \otimes I)(\tb{\ti{w}}_t + \gamma_{t}\cdot y_{t+1}).
		\end{equation}
		Let $\mathds{1} = (1,\dots, 1)$ represents the vector of all 1's. We define the operator $\left\langle \tb{\ti{w}} \right\rangle = \frac{1}{n} (\mathds{1}^{\top} \otimes I) \tb{\ti{w}} = \frac{1}{n} \sum_{i\in {N}} \tb{\ti{w}}^i $. This $\langle \tb{\ti{w}} \rangle \in \mathbb{R}^{k}$ represents the average of the vectors in $\{\tb{\ti{w}}^1, \tb{\ti{w}}^2, \dots, \tb{\ti{w}}^n \}$.
		Moreover, let $\mathcal{J} = (\frac{1}{n} \mathds{1}\mathds{1}^{\top})\otimes I \in \mathbb{R}^{nk \times nk}$ is the projection operator that projects a vector into the consensus subspace $\{\mathds{1}\otimes u: u\in \mathbb{R}^{k}\}$. Thus $\mathcal{J}\tb{\ti{w}} = \mathds{1}\otimes \langle \tb{\ti{w}} \rangle$. Now define the 
		disagreement vector $\tb{\ti{w}}_{\perp} = \mathcal{J}_{\perp}\tb{\ti{w}} = \tb{\ti{w}}- \mathds{1}\otimes \langle \tb{\ti{w}} \rangle$, where $\mathcal{J}_{\perp} = I-\mathcal{J}$. Here $I$ is $nk \times nk$ dimensional identity matrix. The iteration $\tb{\ti{w}}_t$ can be decomposed 
		as the sum of a vector in disagreement space and a vector in consensus space, i.e., $\tb{\ti{w}}_t = \tb{\ti{w}}_{\perp,t} + \mathds{1} \otimes \langle \tb{\ti{w}}_t \rangle$. The proof of convergence consists of two steps.
		
		
		\textbf{Step 01:} To show $\lim_t~ \tb{\ti{w}}_{\perp,t} = 0$ a.s. From Proposition \ref{prop: w_boundedness} we have $\mathbb{P}[sup_t ||\tb{\ti{w}}_t|| < \infty] = 1$, i.e., $\mathbb{P}[\cup_{K_1 \in \mathbb{Z}^{+}}~\{sup_t ||\tb{\ti{w}}_t|| < K_1\}] = 1$. It suffices to show that $\lim_t~ \tb{\ti{w}}_{\perp, t} \mathds{1}_{\{sup_t ||\tb{\ti{w}}_t|| < K_1\}} = 0$ for any $K_1\in \mathbb{Z}^+$. Lemma 5.5 in \cite{zhang2018fully} proves the boundedness of  $\mathbb{E} \left[||\beta_{t}^{-1}\tb{\ti{w}}_{\perp,t}||^2\right]$ over the set  $\{sup_t ||\tb{\ti{w}}_t|| \leq K_1\}$, for any $K_1>0$. We state the lemma here. 
		\begin{prop}[Lemma 5.5 in \cite{zhang2018fully}]
			\label{prop: sup_algo1}
			Under assumptions \ref{ass: comm_matrix}, and  \ref{ass: full_rank} for any $K_1>0$, we have 
			\begin{equation*}
			sup_t ~\mathbb{E}[||\beta_{t}^{-1}\tb{\ti{w}}_{\perp,t}||^2 \mathds{1}_{\{sup_t ||\tb{\ti{w}}_t|| \leq K\}}] < \infty.
			\end{equation*}
		\end{prop}
		From Proposition \ref{prop: sup_algo1} we obtain that for any $K_1>0, ~\exists ~ K_2< \infty$ such that for any $t\geq 0,~ \mathbb{E}[||\tb{\ti{w}}_{\perp,t}||^2] < K_2 \gamma_{t}^2 $ over the set $\{sup_t~ ||\tb{\ti{w}}_t|| < K_1\}$. Since $\sum_t \gamma_{t}^2 < \infty$, by Fubini's theorem we have $\sum_t \mathbb{E}(||\tb{\ti{w}}_{\perp, t}||^2 \mathds{1}_{\{sup_{t} \ ||\tb{\ti{w}}_t|| < K_1 \}}) < \infty$. Thus, $\sum_t ||\tb{\ti{w}}_{\perp, t}||^2 \mathds{1}_{\{sup_{t} \   ||\tb{\ti{w}}_t|| < K_1\}} < \infty$ a.s. Therefore, $\lim_t ~ \tb{\ti{w}}_{\perp, t} \mathds{1}_{\{sup_t ||\tb{\ti{w}}_t|| < K_1\}} =0 $ a.s. Since $\{sup_t ||\tb{\ti{w}}_t||  < \infty\}$ with probability 1, thus $\lim_t~ \tb{\ti{w}}_{\perp,t} = 0$ a.s. This ends the proof of Step 01.
		
		\textbf{Step 02:} To show the convergence of the consensus vector $\mathds{1}\otimes \langle \tb{\ti{w}}_t \rangle$, 
		first note that the iteration of  $\langle \tb{\ti{w}}_t \rangle$ (Equation (\ref{eqn: z_recursion})) can be written as
		\begin{eqnarray}
		\langle \tb{\ti{w}}_{t+1} \rangle  &=& \frac{1}{N} (\mathds{1}^{\top} \otimes I)(L_t \otimes I)(\mathds{1}\otimes \langle \tb{\ti{w}}_t \rangle + \tb{\ti{w}}_{\perp, t} + \gamma_{t}~ y_{t+1}) \nonumber
		\\
		&=& \langle \tb{\ti{w}}_t \rangle + \gamma_{t} \langle (L_t \otimes I)(y_{t+1} + \gamma_{t}^{-1} \tb{\ti{w}}_{\perp,t}) \rangle \nonumber
		\\
		&=&  \langle \tb{\ti{w}}_{t} \rangle + \gamma_{t}~ \mathbb{E}( \langle y_{t+1} \rangle | \mathcal{F}_{t}) + \beta_{t}\xi_{t+1}, \label{eqn: z_update}
		\end{eqnarray}
		where 
		\begin{eqnarray*}
			\xi_{t+1} &=& \langle (L_t \otimes I)(y_{t+1} + \gamma_{t}^{-1}\tb{\ti{w}}_{\perp,t}) \rangle - \mathbb{E}( \langle y_{t+1} \rangle| \mathcal{F}_{t}), ~~ and
			\\
			\label{y}
			\langle y_{t+1} \rangle &=& [(\bar{c}_{t+1} - \psi_t^{\top} \langle \tb{\ti{w}}_t \rangle)\psi_t^{\top}]^{\top}.
		\end{eqnarray*}
		Note that $\mathbb{E}(\langle y_{t+1} \rangle | \mathcal{F}_{t})$ is Lipschitz continuous in $\langle \tb{\ti{w}}_t \rangle$. Moreover, $\xi_{t+1}$ is a martingale difference sequence and satisfies 
		\begin{equation}
		\label{eqn: xi-martingale}
		\mathbb{E}[||\xi_{t+1}||^2 ~|~ \mathcal{F}_{t}] \leq  \mathbb{E}[||y_{t+1} + \gamma_{t}^{-1}\tb{\ti{w}}_{\perp,t}||^2_{R_t}~|~\mathcal{F}_{t}] +  || \mathbb{E}(\langle y_{t+1} \rangle ~|~ \mathcal{F}_{t}) ||^2,
		\end{equation}
		where $R_t = \frac{L_t^{\top} \mathds{1} \mathds{1}^{\top}L_t \otimes I}{ n^{2}}$ has bounded spectral norm. Bounding first and second terms in RHS of Equation (\ref{eqn: xi-martingale}), we have, for any $K_1 >0$
		\begin{equation*}
		\mathbb{E}(||\xi_{t+1}||^2 | \mathcal{F}_{t}) \leq K_3 (1+ ||\langle \tb{\ti{w}}_t \rangle||^2),
		\end{equation*}
		over the set  $\{sup_t ~ ||\tb{\ti{w}}_t||\leq K_1\}$ for some $K_3 < \infty$. Thus condition (3) of Assumption \ref{ass: K-C_lemma} is satisfied. The ODE associated with the Equation (\ref{eqn: z_update}) has the form
		\begin{equation}
		\label{eqn: z_ode}
		\langle \dot{\tb{\ti{w}}} \rangle 
		= 
		-\Psi^{\top}D^{\tb{\ti{s}},\tb{\ti{a}}}\Psi 
		\langle \tb{\ti{w}} \rangle
		+
		\Psi^{\top}D^{\tb{\ti{s}},\tb{\ti{a}}}\bar{c}.
		\end{equation}
		Let the RHS of Equation (\ref{eqn: z_ode}) be $h(\langle \tb{\ti{w}} \rangle)$. Note that $h(\langle \tb{\ti{w}} \rangle)$ is Lipschitz  continuous in $\langle \tb{\ti{w}} \rangle$. Also, recall that $D^{\tb{\ti{s}},\tb{\ti{a}}} = diag[d(\tb{\ti{s}}) \cdot \pi(\tb{\ti{s}}, \tb{\ti{a}}), \tb{\ti{s}}\in\mathcal{S}, \tb{\ti{a}}\in \mathcal{A}]$. Hence the ODE given in Equation (\ref{eqn: z_ode}) has unique globally asymptotically stable equilibrium $\tb{\ti{w}}^{\star}$ satisfying 
		\begin{eqnarray*}
			\Psi^{\top}D^{\tb{\ti{s}},\tb{\ti{a}}}(\bar{c} - \Psi\tb{\ti{w}}^{\star}) = 0.
		\end{eqnarray*}
		Moreover, from Propositions \ref{prop: w_boundedness}, and \ref{prop: sup_algo1}, the sequence $\{\tb{\ti{w}}_t\}$ is bounded almost surely, so is the sequence $\{\langle \tb{\ti{w}}_t \rangle\}$. Specializing Corollary 8.1 and Theorem 8.3 on page 114-115 in \cite{borkarbook2edition} we have $\lim_t~ \langle \tb{\ti{w}}_t \rangle = \tb{\ti{w}}^{\star}$ a.s. over the set $\{sup_t~||\tb{\ti{w}}_t|| \leq K_1\}$ for any $K_1>0$. This concludes the proof of Step 02. 
		
		The proof follows from Proposition \ref{prop: w_boundedness} and results from Step 01. Thus, we have $\lim_t~ \tb{\ti{w}}^i_t = \tb{\ti{w}}^{\star}$ a.s. for each $i\in {N}$.
	\end{proof}
	
Apart from the convergence of the cost function parameter, we also require the MAEVI analysis; in particular, we now prove the Theorem \ref{thm: MAEVI_analysis}. 

\subsection{Proof of Theorem \ref{thm: MAEVI_analysis}}
\label{app: evi_analysis}
	Recall the theorem: Let $\beta_t = B \sqrt{nd \log \left( \frac{4}{\delta} \left(n t^2 +  \frac{n t^3 B^2}{\lambda} \right) \right)} + \sqrt{\lambda nd }$, for all $t\geq 1$. Then with probability at least $1-\delta/2$, and for each agent $i\in N$, for all $j_i \geq 1$, MAEVI converges in finite time, and the following holds
		\begin{equation*}
		\boldsymbol{\theta}^{\star} \in \mathcal{C}^i_{j_i} \cap \mathcal{B}, ~ 0 \leq Q^i_{j_i} (\cdot, \cdot) \leq Q^{i\star}(\cdot,\cdot; \tb{\ti{w}}^i), 0 \leq V^i_{j_i} (\cdot) \leq V^{i\star}(\cdot; \tb{\ti{w}}^i)
		\end{equation*}
	\begin{proof}
		To prove this Theorem, for each agent $i\in N$, we decompose $t$ into different rounds. Each round $j_i \geq 1$ of agent $i$ corresponds to $t \in [t^i_{j_i} + 1, t^i_{j_i+1}]$, during which the action-value function estimator is the output $Q^i_{j_i}$ of MAEVI sub-routine by agent $i$. We will apply the induction argument on the rounds to show that optimism holds for all $j_i \geq 1$.
		
		Consider round $j_i = 1$ for agent $i\in N$. In this round, let $t \in [1, t^i_1]$. We have $V^i_0 \leq B_{\star}$, from the algorithm's initialization. Define $\eta^i_{t} = V^i_0(\tb{\ti{s}}_{t+1}) - \langle \phi_{V^i_0}(\tb{\ti{s}}_{t}, \tb{\ti{a}}_{t}), \boldsymbol{\theta}^{\star} \rangle$ for $t \in [1,t^i_1]$.  Then $\{\eta^i_{t}\}_{t=1}^{t^i_1}$ are $B_{\star}$-sub-Gaussian.
		
		Applying Theorem \ref{thm: abbasi_yadkori_thm_1} of  \cite{abbasi2011improved} in our case, with probability at least $\left( 1- \frac{\delta}{n \cdot t^i_1(t^i_1+1)} \right)$ we have,
		\begin{align}
		\left\| \mathbf{\Sigma}_{t}^{i^{-1/2}} \sum_{k=1}^{t} \phi_{V^i_0}(\tb{\ti{s}}_k, \tb{\ti{a}}_k) \eta^i_k \right\|_2 
		& \leq  B_{\star} \sqrt{2 \log \left( \frac{\det(\mathbf{\Sigma}^i_{t})^{1/2}}{\delta/(n\cdot t^i_1(t^i_1+1)) \cdot \det(\mathbf{\Sigma}^i_{t-1})^{1/2}}
			\right)} \nonumber
		\\ 
		&\overset{(i)}{=} B_{\star} \sqrt{2 \log \left( \frac{\det(\mathbf{\Sigma}^i_{t})^{1/2}}{\delta/(n \cdot t^i_1(t^i_1+1)) \cdot \lambda^{nd/2}}
			\right)} \nonumber
		\\
		& \overset{(ii)}{\leq}  B_{\star} \sqrt{2 \log \left( \frac{ \left(\lambda + \frac{t B_{\star}^2 nd}{nd} \right)^{nd/2}}{\delta/(n \cdot t^i_1(t^i_1+1)) \cdot \lambda^{nd/2}}  \right)}  \nonumber
		\\
		& =  B_{\star} \sqrt{2 \log \left( \frac{ \lambda^{nd/2} \cdot \left(1+ \frac{t B_{\star}^2 }{\lambda} \right)^{nd/2}}{\delta/(n \cdot t^i_1(t^i_1+1)) \cdot \lambda^{nd/2}}  \times \frac{(\delta/(n \cdot t^i_1(t^i_1+1)))^{nd/2}}{(\delta/(n \cdot t^i_1(t^i_1+1)))^{nd/2}}  \right)}  \nonumber
		\\
		& =  B_{\star} \sqrt{2 \log \left( \frac{ \left(1+ \frac{t B_{\star}^2}{\lambda} \right)^{nd/2}}{(\delta/(n \cdot t^i_1(t^i_1+1)))^{nd/2}}  \times \frac{(\delta/(n \cdot t^i_1(t^i_1+1)))^{nd/2}}{\delta/(n \cdot t^i_1(t^i_1+1))} \right)} \nonumber
		\\
		& =  B_{\star} \sqrt{2 \log \left( \frac{ 1+ \frac{t B_{\star}^2 }{\lambda}}{\delta/(n \cdot t^i_1(t^i_1+1))} \right)^{nd/2} + 2 \log \left( \frac{\delta}{n \cdot t^i_1(t^i_1+1)} \right)^{nd/2 - 1}} \nonumber
		\\
		& \overset{(iii)}{\leq}  B_{\star} \sqrt{2 \log \left( \frac{ 1+ \frac{t B_{\star}^2}{\lambda}}{\delta/(n \cdot t^i_1(t^i_1+1))} \right)^{nd/2} } \nonumber
		\\ 
		&= B_{\star} \sqrt{nd \log \left( \frac{ 1+ \frac{t B_{\star}^2}{\lambda}}{\delta/(n \cdot t^i_1(t^i_1+1))} \right) } \nonumber
		\\
		& =  B_{\star} \sqrt{nd \log \left( \frac{n \cdot t^i_1(t^i_1+1) +  \frac{n \cdot t \cdot t^i_1(t^i_1+1) B_{\star}^2}{\lambda}}{\delta} \right)},
		\label{eqn: MAEVI_lhs_part1_round1}
		\end{align}
		
		where $(i)$ follows from the fact that $det (\mathbf{\Sigma}^i_{t - 1}) = \lambda^{nd}$; $(ii)$ uses the determinant trace inequality (Theorem \ref{thm: det_trace_inequality}) along with the assumption that $|| \phi_{V^i_0} || \leq B_{\star} \sqrt{nd}$; and $(iii)$ uses the fact that $0< \frac{\delta}{n \cdot t^i_1 (t^i_1 + 1)} \leq 1$ hence $\log \left( \frac{\delta}{n \cdot t^i_1 (t^i_1 + 1)} \right) <0$. 
		
		Next, we consider the LHS of Equation \eqref{eqn: MAEVI_lhs_part1_round1} and give the lower bound for the same.
		\begin{align}
		\left\| \mathbf{\Sigma}_{t}^{i^{-1/2}} \sum_{k=1}^{t} \phi_{V^i_0}(\tb{\ti{s}}_k, \tb{\ti{a}}_k) \eta^i_k \right\|_2 
		&\overset{(i)}{=}
		\left\| \mathbf{\Sigma}_{t}^{i^{-1/2}} \sum_{k=1}^{t} \phi_{V^i_0}(\tb{\ti{s}}_k, \tb{\ti{a}}_k) \cdot (V^i_0(\tb{\ti{s}}_{k+1}) - \langle \phi_{V^i_0}(\tb{\ti{s}}_{k}, \tb{\ti{a}}_{k}), \boldsymbol{\theta}^* \rangle)  \right\|_2 \nonumber
		\\
		& =  \left\| \mathbf{\Sigma}_{t}^{i^{1/2}} \mathbf{\Sigma}_{t}^{i^{-1}} \sum_{k=1}^{t} \phi_{V^i_0}(\tb{\ti{s}}_k, \tb{\ti{a}}_k) \cdot V^i_0(\tb{\ti{s}}_{k+1})  
		- \mathbf{\Sigma}_{t}^{i^{1/2}} \mathbf{\Sigma}_{t}^{i^{-1}} (\mathbf{\Sigma}_{t}^{i} - \lambda \mathbf{I}) \boldsymbol{\theta}^{\star} \right\|_2  \nonumber
		\\
		&\overset{(ii)}{=} \left\| \mathbf{\Sigma}_{t}^{i^{1/2}} \hat{\boldsymbol{\theta}}^i_{t} 
		- \mathbf{\Sigma}_{t}^{i^{1/2}}  \boldsymbol{\theta}^{\star} + \lambda \mathbf{\Sigma}_{t}^{i^{-1/2}} \boldsymbol{\theta}^{\star} \right\|_2  \nonumber
		\\
		&\overset{(iii)}{\geq}  \left\| \mathbf{\Sigma}_{t}^{i^{1/2}} (\hat{\boldsymbol{\theta}}^i_{t} - \boldsymbol{\theta}^{\star} ) \right\|_2 - \left\|
		\lambda \mathbf{\Sigma}_{t}^{i^{-1/2}} \boldsymbol{\theta}^{\star} \right\|_2  \nonumber
		\\
		&\overset{(iv)}{\geq}  \left\| \mathbf{\Sigma}_{t}^{i^{1/2}} (\hat{\boldsymbol{\theta}}^i_{t} - \boldsymbol{\theta}^{\star} ) \right\|_2 - 
		\sqrt{\lambda nd }, 
		\label{eqn: MAEVI_lhs_part2_round1}
		\end{align}
		where $(i)$ follows by the definition of $\eta^i_k$; $(ii)$ uses the updates of $\Sigma^i_t$ and  definition of $\hat{\boldsymbol{\theta}}^i_{t}$ in the algorithm; $(iii)$ is the consequence of the triangle inequality, i.e., $||a|| -||b|| \leq ||a+b||$
		; and finally $(iv)$ follows from the fact that $|| \boldsymbol{\theta}^{\star} || 
		\leq \sqrt{nd}$.
		
		From  Equations \eqref{eqn: MAEVI_lhs_part1_round1} and \eqref{eqn: MAEVI_lhs_part2_round1} we have the following:
		
		\begin{equation*}
		B_{\star} \sqrt{nd \log \left( \frac{n \cdot t^i_1(t^i_1+1) +  \frac{n \cdot t \cdot t^i_1(t^i_1+1) B_{\star}^2}{\lambda}}{\delta} \right)}
		\geq 
		\left\| \mathbf{\Sigma}_{t}^{i^{1/2}} (\hat{\boldsymbol{\theta}}^i_{t} - \boldsymbol{\theta}^{\star} ) \right\|_2 - 
		\sqrt{\lambda nd }. 
		\end{equation*}
		From the definition of $\beta_t$ in this Theorem, we have
		\begin{equation*}
		\left\| \mathbf{\Sigma}_{t}^{i^{1/2}} (\hat{\boldsymbol{\theta}}^i_{t} - \boldsymbol{\theta}^{\star} ) \right\|_2 \leq B_{\star} \sqrt{nd \log \left( \frac{n \cdot t^i_1(t^i_1+1) +  \frac{n \cdot t \cdot t^i_1(t^i_1+1) B_{\star}^2}{\lambda}}{\delta} \right)} + \sqrt{\lambda nd } \leq \beta_{t^i_1}.
		\end{equation*}
		Since above holds for all $t\in [1, t^i_1]$ with probability at least $1 - \frac{\delta}{n \cdot t^i_1(t^i_1 + 1)}$, the true parameters $\boldsymbol{\theta}^{\star} \in \mathcal{C}^i_1 \cap \mathcal{B}$. This implies in round 1, for each agent $i\in N$, the true parameters are in set $\mathcal{C}^i_1 \cap \mathcal{B}$. To complete round 1, we need to show that the output $Q^i_1$ and $V^i_1$ of MAEVI are optimistic for each agent $i\in N$. This will be done using the second induction argument on the loop of MAEVI. For  the base step, it follows from the non-negativity of the $Q^{i\star}(\cdot, \cdot; \tb{\ti{w}}^i)$ and $V^{i\star}(\cdot; \tb{\ti{w}}^i)$ that $Q^{i, (0)}(\cdot, \cdot) \leq Q^{i\star}(\cdot, \cdot; \tb{\ti{w}}^i)$ and $V^{i,(0)}(\cdot) \leq V^{i\star}(\cdot; \tb{\ti{w}}^i)$.
		
		Now, assume that $Q^{i, (l)}(\cdot, \cdot)$ and $V^{i, (l)}(\cdot, \cdot)$ are optimistic. For the $(l+1)$-th iterate in MAEVI algorithm, we have
		\begin{align*}
		Q^{i,(l+1)}(\cdot,\cdot) 
		&=
		\bar{c}(\cdot,\cdot;\tb{\ti{w}}^i) + (1-q)~ \min_{\boldsymbol{\theta} \in \mathcal{C}^i_1 \cap \mathcal{B}} ~\left\langle \boldsymbol{\theta}, \phi_{V^{i,(l)}}(\cdot,\cdot)  \right\rangle
		\\
		& \overset{(i)}{=} 
		\bar{c}(\cdot,\cdot;\tb{\ti{w}}^i) + (1-q)~ \left\langle \boldsymbol{\theta}^{\star}, \phi_{V^{i,(l)}}(\cdot,\cdot)  \right\rangle
		\\
		& \overset{(ii)}{=}  \bar{c}(\cdot,\cdot;\tb{\ti{w}}^i) + (1-q)~ \mathbb{P} V^{i, (l)} (\cdot, \cdot)
		\\
		& \overset{(iii)}{\leq}  \bar{c}(\cdot,\cdot;\tb{\ti{w}}^i) + \mathbb{P} V^{i, (l)} (\cdot, \cdot)
		\\
		& \overset{(iv)}{\leq}  Q^{i\star} (\cdot, \cdot; \tb{\ti{w}}^i), 
		\end{align*}
		where $(i)$ holds because $\boldsymbol{\theta}^{\star}$ is the minimizer; $(ii)$ is by definition  of the linear function approximation of the cost-to-go function. $(iii)$ is because $(1-q)$ is a positive fraction. The last inequality uses the definition of $Q^{i\star}(\cdot, \cdot; \tb{\ti{w}}^i)$ and the induction hypothesis on $l$, $V^{i,(l)}(\cdot)$ is optimistic, i.e., $V^{i,(l)}(\cdot) \leq V^{i\star}(\cdot; \tb{\ti{w}}^i)$. Therefore, by induction $Q^{i, (l+1)}(\cdot,\cdot)$ is also optimistic for all $l$, and hence the final output $Q^i_1(\cdot, \cdot)$ and $V^{i}_1(\cdot)$ are optimistic. This finishes the proof of round 1.
		
		For the outer induction, suppose that the event in Theorem \ref{thm: MAEVI_analysis} holds for round $1$ to $j_i -1$ for each agent $i\in N$ with high probability. Therefore, $\boldsymbol{\theta}^{\star} \in \mathcal{C}^i_k \cap \mathcal{B}$ for rounds $k = 1, 2, \dots, j_i - 1$. And in each round $k=1,2, \dots, j_i -1$, the output of the MAEVI algorithms are optimistic, i.e., $Q^i_{k}(\cdot, \cdot) \leq Q^{i \star}_k (\cdot, \cdot; \tb{\ti{w}}^i)$ and $V^i_{k}(\cdot) \leq V^{i \star}_k (\cdot; \tb{\ti{w}}^i),~\forall ~i\in N$. So, for each agent $i\in N$, we define the following event 
		\begin{equation*}
		\mathcal{E}^i_{j_i -1} = \{ \boldsymbol{\theta}^{\star} \in \mathcal{C}^i_{k} \cap \mathcal{B}; ~ Q^i_{k}(\cdot, \cdot) \leq Q^{i \star} (\cdot, \cdot; \tb{\ti{w}}^i), ~ V^i_{k}(\cdot) \leq V^{i \star} (\cdot; \tb{\ti{w}}^i)~ \text{for each} ~k=1,2, \dots, j_i - 1\}.
		\end{equation*}
		and assume that the $\mathbb{P}(\mathcal{E}^i_{j_i -1}) \geq 1 - \frac{\delta^{\prime}}{n}$ for some $\delta^{\prime} >0$. We now show that the event $\mathcal{E}^i_{j_i}$ also holds with high probability. To do this, we will construct the auxiliary sequence of functions for each agent $i\in N$ as follows:
		$$
		\widetilde{V}^i_{k}(\cdot) \coloneqq \min\{ V^i_k(\cdot), B_{\star}\}, ~\forall ~k=1, 2,\dots, j_i -1.
		$$
		Moreover, for any $k = 1,2,\dots, j_i$ and for any $t\in [t^i_{k-1} + 1 , t^i_k]$, define the following: 
		\begin{eqnarray*}
			\widetilde{\eta}^i_t &=& V^i_{k-1} (\tb{\ti{s}}_{t+1}) - \left\langle \phi_{\widetilde{V}^i_{k-1}} (\tb{\ti{s}}_t, \tb{\ti{a}}_t), \boldsymbol{\theta}^{\star} \right\rangle
			\\
			\mb{\widetilde{\Sigma}}^i_t &=& \lambda \mb{I} + \sum_{r=1}^t \phi_{\widetilde{V}^i_{k(r) - 1}}(\tb{\ti{s}}_r, \tb{\ti{a}}_r) \phi_{\widetilde{V}^i_{k(r) - 1}}  (\tb{\ti{s}}_r, \tb{\ti{a}}_r)^{\top}
			\\
			\widetilde{\boldsymbol{\theta}}^i_t &=& {\mb{\widetilde{\Sigma}}^{i^{-1}}_t} \sum_{r=1}^t \phi_{\widetilde{V}^i_{k(r) - 1}}(\tb{\ti{s}}_r, \tb{\ti{a}}_r) \widetilde{V}^i_{k(r) - 1}(\tb{\ti{s}}_{r+1})
			\\
			\mathcal{\widetilde{C}}^i_k &=& \{ \boldsymbol{\theta} \in \mathbb{R}^{nd} : \left\| \widetilde{\mathbf{\Sigma}}_{t^i_k}^{i^{1/2}} (\widetilde{\boldsymbol{\theta}}^i_{t^i_k} - \boldsymbol{\theta} ) \right\|_2 \leq \beta_{t^i_k} \},
		\end{eqnarray*}
		where $k(r)$ is the round that contains the time period $r$, i.e., $r \in [t^i_{k-1}+1, t^i_k]$.
		
		By construction $\{\widetilde{\eta}^i_t\}_{t=1}^{t^i_{j_i}}$ are almost surely $B_{\star}$ sub-Gaussian. Moreover, using the similar computations as above and \cite{abbasi2011improved} result (Theorem \ref{thm: abbasi_yadkori_thm_1}), we can say that event $\widetilde{\mathcal{E}}^i_{j_i}$ will hold with high probability where 
		\begin{equation*}
		\mathcal{\widetilde{E}}^i_{j_i} = \{ \boldsymbol{\theta}^{\star} \in \widetilde{\mathcal{C}}^i_{j_i} \cap \mathcal{B}; ~ Q^i_{j_i}(\cdot, \cdot) \leq Q^{i \star} (\cdot, \cdot; \tb{\ti{w}}^i); ~ V^i_{j_i}(\cdot) \leq V^{i \star} (\cdot; \tb{\ti{w}}^i)\},
		\end{equation*}
		and $Q^i_{j_i}$ is the output of 
		$ MAEVI( \widetilde{\mathcal{C}}^i_{j_i}, \epsilon^i_{j_i}, \frac{1}{t^i_{j_i}}, \tb{\ti{w}}^i)$. 
		
		Moreover, under the event $\mathcal{E}^i_{j_i -1}$, the optimism implies that $\widetilde{V}^i_k = V^i_k$ for all $k=1,2, \dots, j_i-1$ and for all $i\in N$. Also, under the event  $\mathcal{E}^i_{j_i -1}$, we have $\widetilde{\eta}^i_t = \eta^i_t, \widetilde{\mb{\Sigma}}^i_t  = \mb{\Sigma}^i_t, \widetilde{\boldsymbol{\theta}}^i_t = \hat{\boldsymbol{\theta}}^i_t$ for all $t\leq t^i_{j_i}$, and thus $\widetilde{\mathcal{C}}^i_{j_i} = \mathcal{C}^i_{j_i}$. So, for each agent $i\in N$, we have 
		\begin{equation*}
		\mathcal{E}^i_{j_i} = \mathcal{E}^i_{j_i - 1} \cap \mathcal{\widetilde{E}}^i_{j_i}, ~~\forall ~   i\in N.
		\end{equation*}
		Therefore, using the union bound, we have $\mathbb{P} (\mathcal{E}^i_{j_i}) \geq 1 - \frac{\delta^{\prime}}{n} - \frac{\delta}{n\cdot t^i_{j_i} (t^i_{j_i} + 1)}$. Now, by induction argument and taking the union bound, we have 
		\begin{equation*}
		\sum_{j_i=1}^{J^i} \frac{\delta}{n\cdot t^i_{j_i} (t^i_{j_i} + 1)} = \sum_{j_i=1}^{J^i} \frac{\delta}{n}\cdot \left(\frac{1}{t^i_{j_i}} - \frac{1}{(t^i_{j_i} + 1)} \right) \leq \frac{\delta}{n},
		\end{equation*}
		from here we conclude that with a probability of at least $1 - \frac{\delta}{n}$, the good event holds for all the $j_i \leq J^i$, where $J^i$ is the number of calls to the MAEVI algorithm by agent $i\in N$.  
		
		Next, it remains to show that the MAEVI will converge in finite time with $\epsilon^i$ tolerances for any agent $i\in N$. To do so, it suffices to show that $\|V^{i, (l)} - V^{i, (l-1)}\|_\infty$ shrinks exponentially. We now claim that $\|Q^{i, (l)} - Q^{i, (l-1)}\|_\infty$ shrinks exponentially, which together with $Q^i$ update in algorithm gives the desired result since $\|V^{i,(l)}-V^{i,(l-1)} \|_\infty \leq \|Q^{i, (l)} - Q^{i, (l-1)}\|_\infty$. To show this, first note that for any $(\tb{\ti{s}},\tb{\ti{a}})$ pair we have,
		\begin{align*}
		|Q^{i,(l)} (\tb{\ti{s}}, \tb{\ti{a}}) - Q^{i, (l-1)} (\tb{\ti{s}}, \tb{\ti{a}}) | & = (1-q) \cdot \left| \min_{\boldsymbol{\theta} \in \mathcal{C}^i \cap \mathcal{B}} \langle \boldsymbol{\theta},\phi_{V^{i,(l-1)}}(\tb{\ti{s}}, \tb{\ti{a}}) \rangle - \min_{\boldsymbol{\theta} \in \mathcal{C}^i \cap \mathcal{B}} \langle \boldsymbol{\theta},\phi_{V^{i,(l-2)}}(\tb{\ti{s}}, \tb{\ti{a}}) \rangle \right| 
		\\
		&= (1-q) \cdot \left| \min_{\boldsymbol{\theta} \in \mathcal{C}^i \cap \mathcal{B}} \langle \boldsymbol{\theta},\phi_{V^{i,(l-1)}}(\tb{\ti{s}}, \tb{\ti{a}}) \rangle + \max_{\boldsymbol{\theta} \in \mathcal{C}^i \cap \mathcal{B}} \langle \boldsymbol{\theta}, - \phi_{V^{i,(l-2)}}(\tb{\ti{s}}, \tb{\ti{a}}) \rangle \right|
		\\
		&= (1-q) \cdot \left|  - \max_{\boldsymbol{\theta} \in \mathcal{C}^i \cap \mathcal{B}} \langle \boldsymbol{\theta}, -\phi_{V^{i,(l-1)}}(\tb{\ti{s}}, \tb{\ti{a}}) \rangle + \max_{\boldsymbol{\theta} \in \mathcal{C}^i \cap \mathcal{B}} \langle \boldsymbol{\theta}, - \phi_{V^{i,(l-2)}}(\tb{\ti{s}}, \tb{\ti{a}}) \rangle \right|
		\\
		&\overset{(i)}{\leq} (1-q) \cdot \max_{\boldsymbol{\theta} \in \mathcal{C}^i \cap \mathcal{B}} \left|  - \langle \boldsymbol{\theta}, -\phi_{V^{i,(l-1)}}(\tb{\ti{s}}, \tb{\ti{a}}) \rangle +  \langle \boldsymbol{\theta}, - \phi_{V^{i,(l-2)}}(\tb{\ti{s}}, \tb{\ti{a}}) \rangle \right|
		\\ 
		& =  (1-q)\cdot \max_{\boldsymbol{\theta} \in \mathcal{C}^i \cap \mathcal{B} } \left| \langle \boldsymbol{\theta} , \phi_{V^{i, (l-1)}} (\tb{\ti{s}}, \tb{\ti{a}}) - \phi_{V^{i,(l-2)}} (\tb{\ti{s}}, \tb{\ti{a}}) \rangle \right|
		\\ 
		& \overset{(ii)}{=} (1-q)\cdot \left| \langle \bar{\boldsymbol{\theta}} , \phi_{V^{i,(l-1)}} (\tb{\ti{s}}, \tb{\ti{a}}) - \phi_{V^{i, (l-2)}} (\tb{\ti{s}}, \tb{\ti{a}}) \rangle \right|
		\\ 
		& = (1-q) \cdot |\mathbb{P}_{\bar{\boldsymbol{\theta}}}(V^{i, (l-1)} - V^{i, (l-2)})(\tb{\ti{s}}, \tb{\ti{a}}) |
		\\ 
		& \overset{(iii)}{\leq}  (1-q) \cdot \max_{\tb{\ti{s}}^{\prime} \in \mathcal{S}} \left|V^{i, (l-1)} (\tb{\ti{s}}^{\prime}) - V^{i, (l-2)}(\tb{\ti{s}}^{\prime}) \right|
		\\
		& =  (1-q) \cdot \max_{\tb{\ti{s}}^{\prime} \in \mathcal{S}} \left|\min_{\tb{\ti{a}}^{\prime}}Q^{i, (l-1)} (\tb{\ti{s}}^{\prime}, \tb{\ti{a}}^{\prime}) - \min_{\tb{\ti{a}}^{\prime}}Q^{i, (l-2)}(\tb{\ti{s}}^{\prime}, \tb{\ti{a}}^{\prime}) \right|
		\\
		&\overset{(iv)}{\leq} (1-q) \cdot \max_{\tb{\ti{s}}^{\prime} \in \mathcal{S}, \tb{\ti{a}}^{\prime} \in \mathcal{A} } \left|Q^{i, (l-1)} (\tb{\ti{s}}^{\prime}, \tb{\ti{a}}^{\prime}) - Q^{i, (l-2)}(\tb{\ti{s}}^{\prime}, \tb{\ti{a}}^{\prime}) \right|
		\\
		& =  (1-q)\cdot || Q^{i, (l-1)} - Q^{i, (l-2)} ||_\infty,
		\end{align*} 
		where $(i)$ holds because max is a contraction. $(ii)$  holds because $\bar{\boldsymbol{\theta}}$ is the $\boldsymbol{\theta}$ in the non-empty set $\mathcal{C}^i \cap \mathcal{B}$ that achieves the maximum. Since, $\mathbb{P}_{\bar{\boldsymbol{\theta}}}(\cdot|s,a)$ is a probability distribution we have $(iii)$. Finally, $(iv)$ holds again because of the contraction property of the maximum function.
		Since $(\tb{\ti{s}},\tb{\ti{a}})$ are arbitrary in the above, we conclude that $\|Q^{i, (l)} - Q^{i, (l-1)}\|_\infty \leq (1-q) \|Q^{i, (l-1)} - Q^{i, (l-2)}\|_\infty$. Applying this recursively, we have a term that is exponentially decaying and hence $|| Q^{i, (l)} - Q^{i, (l-1)} ||_\infty$ shrinks exponentially implying $\|V^{i,(l)}-V^{i,(l-1)} \|_\infty$ also shrinks exponentially.
	\end{proof}
	This concludes the MAEVI analysis. Next, we show that the total number of calls, $J$ to the MAEVI algorithm in the entire analysis is bounded. Note that an agent $i\in N$ calls the MAEVI algorithm if at least one of
	the doubling criteria is satisfied, i.e., either the determinant doubling criteria or the time doubling criteria is satisfied. Let $J^i$ be the total number of calls to the MAEVI algorithm made by agent $i$. We will show that $J^i$ is bounded. For agent $i$, let $J^i_1$ be the number of calls to MAEVI made via determinant doubling criteria, and $J^i_2$ be the calls to the MAEVI algorithm via the time doubling criteria. Note that $J^i \leq J^i_1 + J^i_2$. The inequality is taken because there are cases when both the doubling criteria are satisfied. In particular, we use Lemma \ref{lemma: calls_to_evi} that ensures the number of calls, $J$ to the MAEVI algorithm is bounded. 

\subsection{Proof of Lemma \ref{lemma: calls_to_evi}}
\label{app: number_calls_evi}

Recall the lemma: The total number of calls to the MAEVI algorithm in the entire analysis $J$, is bounded as
\begin{equation*}
J \leq 2 n^2d \log \left( 1 + \frac{T B_{\star}^2 nd}{\lambda} \right) + 2n \log(T).
\end{equation*}

\begin{proof}
	To prove this result, let us consider an agent $i\in N$. We bound the number of calls to MAEVI algorithm $J^i$, for each agent $i\in N$ and use the fact that $J \leq \sum_{i\in N} J^i$.
	
	Consider agent $i\in N$, since $V^i_0 \leq B_{\star}$. It is easy to see that
	\begin{eqnarray*}
		|| \Sigma^i_T ||_2 &=& \left\Vert \lambda \mb{I} + \sum_{j=0}^{J^i} \sum_{t=t^i_{j}+1}^{t_{j+1}} \phi_{V^i_j}(\tb{\ti{s}}_t, \tb{\ti{a}}_t) \phi_{V^i_j}(\tb{\ti{s}}_t, \tb{\ti{a}}_t)^{\top} \right\Vert
		\\
		&\leq& \lambda + \sum_{j=0}^{J^i} \sum_{t=t^i_{j}+1}^{t^i_{j+1}} || \phi_{V^i_j}(\tb{\ti{s}}_t, \tb{\ti{a}}_t) ||^2 
		\\
		&\leq& \lambda + T B_{\star}^2 nd.
	\end{eqnarray*}
	The first inequality is because of the triangle inequality. The second one uses the fact that $||\phi_V|| \leq B_{\star}\sqrt{nd}$ and $V^i_j \leq B_{\star}$ for all $j \geq 0$ (from Assumption \ref{ass: tpm_approx}). 
	
	For the determinant doubling criteria, we have $det(\Sigma_T) \leq (\lambda + T B_{\star}^2 nd)^{nd}$.  This implies
	\begin{equation*}
	(\lambda + T B_{\star}^2 nd)^{nd} \geq 2^{J^i_1} \cdot det(\Sigma_0) = 2^{J^i_1} \cdot \lambda^{nd},
	\end{equation*}
	taking log on both sides, we have
	\begin{equation*}
	J^i_1 \leq nd \log_2 \left( 1 + \frac{T B_{\star}^2 nd}{\lambda} \right) \leq 2 nd \log \left( 1 + \frac{T B_{\star}^2 nd}{\lambda} \right).
	\end{equation*}
	This bounds the number of calls to the MAEVI algorithm when the determinant doubling criteria is satisfied. Next, we consider the number of calls to the MAEVI algorithm when the time doubling criteria is satisfied, i.e., we will bound $J^i_2$. Note that $t_0 = 1$, so we have $2^{J^i_2} \leq T$, this implies $J^i_2 \leq \log_2 (T) \leq 2 \log(T)$. Now, summing up the bounds for $J^i_1$ and $J^i_2$ we have the bounds for $J^i$, i.e., 
	\begin{equation*}
	J^i \leq J^i_1 + J^i_2 \leq 2 nd \log \left( 1 + \frac{T B_{\star}^2 nd}{\lambda} \right) + 2 \log(T).
	\end{equation*}
	This proves that the number of calls to the MAEVI algorithm made by an agent $i\in N$ is bounded. 
	So, the total number of calls to MAEVI in the entire analysis is bounded as
	\begin{equation*}
	J \leq \sum_{i\in N} J^i = \sum_{i\in N} 2 nd \log \left( 1 + \frac{T B_{\star}^2 nd}{\lambda} \right) + 2 \log(T) = 2 n^2 d \log \left( 1 + \frac{T B_{\star}^2 nd}{\lambda} \right) + 2n \log(T).
	\end{equation*}
	This ends the proof.
\end{proof}

The next step is to decompose the regret and bound each term. To this end, we prove Theorem \ref{thm: regret_decomposition}. First, we divide the time horizon into disjoint intervals $m=1, 2, \dots, M$, where the endpoint of each interval is decided by one of the two conditions:  1) at least for one agent, the MAEVI is triggered, and 2) all the agents have reached the goal state. Note that this decomposition is explicitly used in the regret analysis only and not in the actual algorithm implementation. Let $H_m$ denote the length of the interval $m$. Moreover, at the end of the $M$-th interval, all the $K$ episodes are over. Therefore, we have the total length of all the intervals as $\sum_{m=1}^M H_m$, which is the same as $\sum_{k=1}^K T_k$, where $T_k$ is the time to finish the episode $k$. 
Within an interval, the optimistic estimator of the state-value function for each agent remains the same. An agent for whom the MAEVI is not triggered in the $m$-th interval will continue using the same optimistic estimator of the state-action value function in the $(m+1)$-th interval. Later we separate the set of agents depending on whether the MAEVI is triggered for them or not. Let $S_m$ be the set of agents for whom the MAEVI is triggered in the $m$-th interval. Using the above interval decomposition, the regret $R_K$ can be written as 
\begin{equation}
\label{eqn: regret_redefine}
R_K = R(M) \leq \sum_{m=1}^M \sum_{h=1}^{H_m}  \frac{1}{n} \sum_{i=1}^n \bar{c}(\tb{\ti{s}}_{m,h}, \tb{\ti{a}}_{m,h}; \tb{\ti{w}}^i) + 1 - \sum_{m\in \mathcal{M}(M)} \frac{1}{n} \sum_{i=1}^n V^i_{j_i(m)}(\tb{\ti{s}}_{init}),
\end{equation}
here $\mathcal{M}(M)$ is the set of all intervals that are the first intervals in each episode. In RHS we add 1 because  $|V^i_0| \leq 1$. Recall, the above equation is the same as Equation \eqref{eqn: regret-eqn-temp} in the main paper.

\subsection{Proof of Theorem \ref{thm: regret_decomposition}}
\label{app: regret-decomposition}
	
	Recall the theorem: Assume that the event in Theorem \ref{thm: MAEVI_analysis} holds, then we have the following upper bound on the regret given in Equation \eqref{eqn: regret_redefine},
	\begin{equation}
	\begin{aligned}
	R(M)  &\leq \underbrace{\sum_{m=1}^M \sum_{h=1}^{H_m} \bigg[ \frac{1}{n} \sum_{i=1}^n \bar{c}(\tb{\ti{s}}_{m,h}, \tb{\ti{a}}_{m,h}; \tb{\ti{w}}^i) + \frac{1}{n} \sum_{i=1}^n \mathbb{P}V^i_{j_i(m)}(\tb{\ti{s}}_{m,h}, \tb{\ti{a}}_{m,h}) - \frac{1}{n} \sum_{i=1}^n V^i_{j_i(m)}(\tb{\ti{s}}_{m,h}) \bigg]}_{E_1} 
	\\
	&~~~ + \underbrace{\sum_{m=1}^M \sum_{h=1}^{H_m} \bigg[ \frac{1}{n} \sum_{i=1}^n V^i_{j_i(m)}(\tb{\ti{s}}_{m,h+1}) - \frac{1}{n} \sum_{i=1}^n \mathbb{P}V^i_{j_i(m)}(\tb{\ti{s}}_{m,h}, \tb{\ti{a}}_{m,h}) \bigg]}_{E_2}
	\\
	& ~~~+ 2n^2dB_{\star} \log \left(1+ \frac{TB_{\star}^2 nd}{\lambda} \right) + 2n B_{\star} \log(T) + 2.
	\end{aligned}
	\end{equation} 
	
	\begin{proof} 
The proof relies on the following proposition that is key result for the regret decomposition. 
	\begin{prop}
		\label{prop: regret_decom_lemma}
		Conditioned on the event given in Theorem \ref{thm: MAEVI_analysis}, for the above mentioned interval decomposition, we have the following:
		\begin{equation*}
		\begin{split}
		\sum_{m=1}^M \left( \sum_{h=1}^{H_m} \left\lbrace \frac{1}{n} \sum_{i=1}^n V^i_{j_i(m)} (\tb{\ti{s}}_{m,h}) - \frac{1}{n} \sum_{i=1}^n V^i_{j_i(m)}(\tb{\ti{s}}_{m,h+1}) \right\rbrace \right) - \sum_{m\in \mathcal{M}(M)} \frac{1}{n} \sum_{i=1}^n V^i_{j_i(m)}(\tb{\ti{s}}_{init})
		\\
		\leq 1 + 2n^2dB_{\star} \log \left(1+ \frac{TB_{\star}^2 nd}{\lambda} \right) + 2n B_{\star} \log(T).
		\end{split}
		\end{equation*}
	\end{prop}
	The proof this proposition is given in 
	Section \ref{proof: reg_dec_initial_lemma} of this appendix. Using this Proposition in the regret expression given in Equation \eqref{eqn: regret_redefine}, we get the regret decomposition as desired.
	\end{proof}

In the next Section, we provide the details of feature design of the transition probabilities and the details of the optimal policy value used in the computations. 

\section{DETAILS OF THE FEATURE DESIGN AND THE OPTIMAL POLICY FOR COMPUTATIONAL EXPERIMENTS}
\label{app: details-computations}
In this section, we will provide the details of the optimal policy and related results. Recall, for each $(\tb{\ti{s}}^{\prime}, \tb{\ti{a}}, \tb{\ti{s}}) \in \mathcal{S} \times \mathcal{A} \times \mathcal{S}$, the global transition probability is parameterized as $\mathbb{P}_{\boldsymbol{\theta}}(\tb{\ti{s}}^{\prime} | \tb{\ti{s}}, \tb{\ti{a}}) = \langle \phi(\tb{\ti{s}}^{\prime} | \tb{\ti{s}},\tb{\ti{a}}), {\boldsymbol{\theta}} \rangle$. The features $\phi(\tb{\ti{s}}^{\prime} | \tb{\ti{s}},\tb{\ti{a}})$ are 
\begin{equation*}
\phi(\tb{\ti{s}}^{\prime} | \tb{\ti{s}},\tb{\ti{a}}) = \begin{cases} (\phi(s^{\prime^{1}} | s^1, \tb{\ti{a}}^1), \dots, \phi(s^{\prime^{n}} | s^n, \tb{\ti{a}}^n)), & if ~\tb{\ti{s}} \neq \tb{\ti{g}},
\\
\mb{0}_{nd}, & if ~ \tb{\ti{s}} = \tb{\ti{g}},~ \tb{\ti{s}}^{\prime} \neq \tb{\ti{g}},
\\
(\mb{0}_{nd-1}, 2^{n-1}), & if ~ \tb{\ti{s}} = \tb{\ti{g}},~ \tb{\ti{s}}^{\prime}  = \tb{\ti{g}},
\end{cases}
\end{equation*}
where $\phi(s^{\prime^{i}} | s^i, \tb{\ti{a}}^i)$ are defined as
\begin{equation*}
\phi(s^{\prime^{i}} | s^i, \tb{\ti{a}}^i) = \begin{cases}  
\left(-\tb{\ti{a}}^i, \frac{1-\delta}{n} \right)^{\top}, & if ~ {s}^i = s^{\prime^{i}} = s_{init}
\\
\left(\tb{\ti{a}}^i, \frac{\delta}{n} \right)^{\top}, & if ~ {s}^i = s_{init},~s^{\prime^{i}} = g
\\
\mathbf{0}_{d}^{\top}, & if ~ {s}^i = g,~s^{\prime^{i}} = s_{init}
\\
\left(\mathbf{0}_{d-1}, \frac{1}{n} \right)^{\top}, & if ~ {s}^i = s^{\prime^{i}} = g.
\end{cases}
\end{equation*}
Here $\mathbf{0}_d^{\top} = (0,0, \dots, 0)^{\top}$ is a vector of $d$ dimension with all zeros. Thus, the features $\phi(s^{\prime^{i}} | s^i, \tb{\ti{a}}^i) \in \mathbb{R}^{nd}$. Moreover, the transition probability parameters are taken as $\boldsymbol{\theta} = \left( \boldsymbol{\theta}^1, \frac{1}{2^{n-1}},  \boldsymbol{\theta}^2, \frac{1}{2^{n-1}} \dots, \boldsymbol{\theta}^n, \frac{1}{2^{n-1}} \right)$, where $\boldsymbol{\theta}^i \in \left\lbrace -\frac{\Delta}{n(d-1)}, \frac{\Delta}{n(d-1)}  \right\rbrace^{d-1} $, and $\Delta < \delta$. 

We first proof Lemma \ref{lemma: feature_properties} to show that these transition probability function features satisfy some basic properties.

\subsection{Proof of Lemma \ref{lemma: feature_properties}}
\label{app: feature_properties}

Recall the lemma: The features $\phi(\tb{\ti{s}}^{\prime} | \tb{\ti{s}},\tb{\ti{a}})$ satisfy the following: (a) $\sum_{\tb{\ti{s}}^{\prime}} \langle \phi(\tb{\ti{s}}^{\prime} | \tb{\ti{s}}, \tb{\ti{a}}), \boldsymbol{\theta}  \rangle = 1, ~ \forall ~\tb{\ti{s}}, \tb{\ti{a}}$; (b) $\langle \phi(\tb{\ti{s}}^{\prime} = \tb{\ti{g}} | \tb{\ti{s}} = \tb{\ti{g}}, \tb{\ti{a}}), \boldsymbol{\theta}  \rangle = 1, ~\forall~\tb{\ti{a}}$; (c) $\langle \phi(\tb{\ti{s}}^{\prime} \neq \tb{\ti{g}} | \tb{\ti{s}} = \tb{\ti{g}}, \tb{\ti{a}}), \boldsymbol{\theta}  \rangle = 0,~ \forall ~\tb{\ti{a}}$.

\begin{proof}
	To prove this lemma, we consider two cases. In case 1, $\tb{\ti{s}} \neq \tb{\ti{g}}$ and case 2, $\tb{\ti{s}} = \tb{\ti{g}}$.
	
	\textbf{Case 01:} ($\tb{\ti{s}} \neq \tb{\ti{g}}$ ). Without loss of generality we consider the following state $\tb{\ti{s}} = (\underbrace{s_{init}, s_{init}, \dots, s_{init}}_{k~ times}, \underbrace{g,g, \dots, g}_{(n-k)~ times})$, i.e., $k \neq 0$ agents are at $s_{init}$ and remaining $n-k$ are at $g$. 
	Consider an agent $i$, who is at $s_{init}$ node. Out of total $2^n$ next possible states, there are exactly $\frac{2^n}{2}$ states in which agent $i$ will remain at $s_{init}$, and in $\frac{2^n}{2}$ states in which the agents move to goal node. 
	The probability that the next node of agent $i$ is $s_{init}$ given that the current node of agent $i$ is $s_{init}$ is given by $- \langle \tb{\ti{a}}^i, \boldsymbol{\theta}^i \rangle + \frac{1-\delta}{n} \times \frac{1}{2^{n-1}}$. And the probability that the next node of agent $i$ is $g$ given that the current node of agent $i$ is $s_{init}$ is  $ \langle \tb{\ti{a}}^i, \boldsymbol{\theta}^i \rangle + \frac{\delta}{n} \times \frac{1}{2^{n-1}}$. These probabilities are obtained using the features defined for our two agent model.
	Since, this is true for all the agents  $1,2,\dots, k$ which are at $s_{init}$. So, the contribution to the probability term from these $k$ agents who are at $s_{init}$ is
	\begin{eqnarray*}
		\sum_{i=1}^k \left\lbrace \left( -\langle \tb{\ti{a}}^i, \boldsymbol{\theta}^i \rangle + \frac{1-\delta}{n} \times \frac{1}{2^{n-1}} \right) \times \frac{2^n}{2} \right\rbrace + \sum_{i=1}^k  \left\lbrace \left(\langle \tb{\ti{a}}^i, \boldsymbol{\theta}^i \rangle + \frac{\delta}{n} \times \frac{1}{2^{n-1}} \right) \times \frac{2^n}{2} \right\rbrace = \frac{k}{n}.
	\end{eqnarray*}
	Next, consider an agent whose state is $g$; for this agent, there are two possibilities, stay at $g$ or go to $s_{init}$. The probability of going to $s_{init}$ is zero, however if the agent stays at $g$ then the probability corresponding to that is $\frac{1}{n} \times \frac{1}{2^{n-1}}$. So the total probability of going to the next state for the agent whose
	state is $g$ is 
	$$
	0 \times \frac{2^n}{2} + \left( \frac{1}{n} \times \frac{1}{2^{n-1}} \right) \frac{2^n}{2}.
	$$
	Again this is true for the agents $k+1, k+2, \dots, n$ who are at $g$, so the overall probability is 
	\begin{equation*}
	\sum_{i = k+1}^n \left\lbrace \left( 0 \times \frac{2^n}{2} + \left( \frac{1}{n} \times \frac{1}{2^{n-1}} \right) \frac{2^n}{2} \right)  \right\rbrace = \frac{n-k}{n}.
	\end{equation*}
	So, the sum of these linearly approximated probabilities is
	\begin{eqnarray*}
		\sum_{\tb{\ti{s}}^{\prime}} \langle \phi(\tb{\ti{s}}^{\prime} | \tb{\ti{s}}, \tb{\ti{a}}), \boldsymbol{\theta}  \rangle &=& \frac{k}{n} + \frac{n-k}{n} = 1.
	\end{eqnarray*}
	This ends the proof of the first case. 
	
	\textbf{Case 02:} ($\tb{\ti{s}} = \tb{\ti{g}}$). For this case, the probability is 
	\begin{eqnarray*}
		\sum_{\tb{\ti{s}}^{\prime}} \langle \phi(\tb{\ti{s}}^{\prime} | \tb{\ti{s}} = \tb{\ti{g}}, \tb{\ti{a}}), \boldsymbol{\theta}  \rangle &=& \sum_{\tb{\ti{s}}^{\prime} \neq \tb{\ti{g}}} \langle \phi(\tb{\ti{s}}^{\prime} | \tb{\ti{s}} = \tb{\ti{g}}, \tb{\ti{a}}), \boldsymbol{\theta}  \rangle + \langle \phi(\tb{\ti{s}}^{\prime} = \tb{\ti{g}} | \tb{\ti{s}} = \tb{\ti{g}}, \tb{\ti{a}}), \boldsymbol{\theta}  \rangle
		\\
		&=& \langle \mb{0}, \boldsymbol{\theta} \rangle + \langle (\mb{0}_{nd-1}, 2^{n-1}), \boldsymbol{\theta}  \rangle = 1.
	\end{eqnarray*}
	Therefore, in both cases, we have
	\begin{equation*}
	\sum_{\tb{\ti{s}}^{\prime}} \langle \phi(\tb{\ti{s}}^{\prime} | \tb{\ti{s}} = \tb{\ti{g}}, \tb{\ti{a}}), \boldsymbol{\theta}  \rangle  = 1, ~~\forall~ \tb{\ti{s}}, \tb{\ti{a}}. 
	\end{equation*}
	The other two statements of the lemma follow by feature design and model parameter space.
\end{proof}

Next, we provide the details of the value of the optimal policy for the above model. We require it in the regret computations. First, note that the value of the optimal policy can be defined in terms of expected cost if all agents reach the goal state exactly in a $1$ time period, in a $2$ time period, and so on. Suppose the agents reach to the goal state exactly by $t$ steps such that $x_j$ agents move to the goal node at each step $j=1,2, \dots, t-1$ and remaining $x_t = n-\sum_{j=1}^{t-1} x_j$ agents move to the goal node by $t$-th step. We call this sequence of departures of agents to the goal node as the `departure sequence'. For the above departure sequence, the cost incurred is 
\begin{equation}
\begin{aligned}
C_{\alpha}(x_1, \dots, x_{t}) &= \alpha \times \sum_{j=1}^{t-1} x_j^2 + \sum_{j=1}^{t-1} \bigg( n - \sum_{i = 1}^{j} x_i \bigg) \cdot c_{\min} + \alpha \times x_t^2
\\
&= \alpha \times \sum_{j=1}^{t-1} x_j^2 + \sum_{j=1}^{t-1} \bigg( n - \sum_{i = 1}^{j} x_i \bigg) \cdot c_{\min} + \alpha \times \bigg(n- \sum_{j=1}^{t-1} x_j \bigg)^2.
\end{aligned}
\end{equation}
where $\alpha$ is the mean of the uniform distribution $\mathcal{U}(c_{\min}, 1)$, i.e., $\alpha = \frac{c_{\min} + 1 }{2}$. We use this $\alpha$ instead of the private cost to compute the optimal value. The first term in the above equation is because all $x_j$ agents have moved to the goal node in time period $j$; hence the congestion is $x_j$. Moreover, each agent incurs the private cost $\alpha$ in this period. 
So, the cost incurred to any agent is $\alpha x_j$, and the number of agents moved are $x_j$, and hence the total cost incurred is $(\alpha x_j)\times  x_j = \alpha x_j^2$. This happens for all the time periods $j =1, 2, \dots, t-1$, so we sum this for $(t-1)$ periods. The remaining agents at each time period incurred a waiting cost of $c_{\min}$; thus, we have a second term. Finally, the third term is because the remaining agents, $x_t$ will move to the goal node at the last time period $t$.
The optimal value of a policy $\pi^{\star}$, $ V^\star \coloneqq V^{\pi^\star}$, can be written as
\begin{equation}
V^{\star} = \sum_{t=1}^{\infty} \mathbb{P}[x_1^{\star}, \dots, x_{t}^{\star}] \cdot C_{\alpha}(x_1^{\star}, \dots, x_{t}^{\star}),
\end{equation}
where $x_1^{\star}, x_2^{\star}, \dots, x_{t-1}^{\star}, x_t^{\star}$ is the optimal departure sequence. These $x_j^{\star}$ are obtained by minimizing the cost of function $C_{\alpha}(x_1, \dots, x_{t})$. Moreover, $\mathbb{P}[x_1^{\star}, \dots, x_{t}^{\star}]$ is the probability of occurrence of this optimal departure sequence. In Theorem \ref{thm: optimal_x_j_star} we provide the optimal departure sequence $x_1^{\star}, \dots, x_{t}^{\star}$, and the corresponding value $C_{\alpha}(x_1^{\star}, \dots, x_{t}^{\star})$.

\subsection{Proof of Theorem \ref{thm: optimal_x_j_star}}
\label{app: optimal_x_j_star}
Recall the theorem: The optimal departure sequence is given by 
	\begin{equation*}
	x_j^{\star}  = \left\lfloor \frac{n}{t} + \left(\frac{t + 1}{2} - j  \right) \cdot \frac{c_{\min}}{2 \alpha}\right\rfloor, ~\forall ~j=1,2, \dots, t-1; ~~~~    x_t^{\star} = n -\sum_{j=1}^{t-1} x_j^{\star}.
	\end{equation*}
	Moreover, the optimal cost 
	$C_{\alpha}(x_1^{\star},  \dots, x_{t}^{\star}) $ of using the above optimal departure sequence is
	\begin{equation*}
	\alpha t \left(\frac{n}{t}\right)^2 + n(t-1) \cdot \frac{c_{\min}}{2 \alpha} - \frac{t(t-1)(t+1)}{12} \cdot \frac{c_{\min}^2}{4\alpha^2}.
	\end{equation*}
	
\label{proof: optimal_x_j_star}

\begin{proof}
	First, recall for the departure sequence $x_1, \dots, x_t$, the cost function is given by
	\begin{equation*}
	\begin{aligned}
	C_{\alpha}(x_1, \dots, x_{t}) &= \alpha \times \sum_{j=1}^{t-1} x_j^2 + \sum_{j=1}^{t-1} \bigg( n - \sum_{i = 1}^{j} x_i \bigg) \cdot c_{\min} + \alpha \times x_t^2
	\\
	&= \alpha \times \sum_{j=1}^{t-1} x_j^2 + \sum_{j=1}^{t-1} \bigg( n - \sum_{i = 1}^{j} x_i \bigg) \cdot c_{\min} + \alpha \times \bigg(n- \sum_{j=1}^{t-1} x_j \bigg)^2.
	\end{aligned}
	\end{equation*}
	The proof of this theorem follows by taking the partial derivative of the cost function with respect to the $x_j$'s and equating them to zero. The first order conditions are necessary and sufficient because the Hessian of the above cost function is positive definite (shown below), and hence the minima exist. The partial derivative of the cost function with respect to $x_j$ is given by
	\begin{equation}
	\label{eqn: derv_at_j}
	\frac{\partial{C}}{\partial{x_j}} = 2 \alpha x_j - 2 \alpha \left(n- \sum_{i=1}^{t-1} x_i \right) - (t-j) c_{\min} = 0, ~~\forall~ j=1,2, \dots, t-1.
	\end{equation}
	From the above equations, we have 
	\begin{equation*}
	x_{j} - x_{j+1} = \frac{c_{\min}}{2 \alpha}, ~\forall ~j=1,2, \dots, t-2.
	\end{equation*}
	Converting all the variables $x_2, x_3, \dots, x_{t-1}$ in terms of $x_1$, we have
	\begin{equation}
	\label{eqn: x_j_recursion}
	x_j = x_1 - (j-1) \cdot \frac{c_{\min}}{2 \alpha}, ~~\forall~j=2, 3, \dots, t-1.
	\end{equation}
	Also, from Equation \eqref{eqn: derv_at_j}, for $j=1$, we have 
	\begin{equation*}
	\frac{\partial{C}}{\partial{x_1}} =  2 \alpha x_1 - 2 \alpha \left(n- \sum_{i=1}^{t-1} x_i \right) - (t-1) c_{\min} = 0.
	\end{equation*}
	Substituting the variables $x_2, x_3, \dots, x_{t-1}$ in terms of $x_1$ in above equation, and solving for $x_1$ using Equation \eqref{eqn: x_j_recursion}, we have 
	\begin{equation*}
	2 \alpha x_1 - 2 \alpha \left(n- \sum_{i=1}^{t-1} x_i \right) - (t-1) c_{\min} = 2 \alpha x_1 - 2 \alpha \left(n- \sum_{i=1}^{t-1} \left(x_1 - (i-1) \cdot \frac{c_{\min}}{2\alpha} \right) \right) - (t-1) c_{\min} = 0.
	\end{equation*}
	Solving for $x_1$, we have
	\begin{equation}
	x_1^{\star} = \frac{n}{t} + \frac{t-1}{2} \cdot \frac{c_{\min}}{2 \alpha}.
	\end{equation}
	Putting them in Equation \eqref{eqn: x_j_recursion}, we have
	\begin{equation*}
	\begin{aligned}
	x_j^{\star} &= \frac{n}{t} + \left(\frac{t-1}{2} - (j-1) \right) \cdot \frac{c_{\min}}{2 \alpha}
	= \left\lfloor\frac{n}{t} + \left(\frac{t + 1}{2} - j  \right) \cdot \frac{c_{\min}}{2 \alpha} \right\rfloor, ~\forall ~j=1,2, \dots, t-1.
	\\
	x_t^{\star} &= n - \sum{j=1}^{t-1} x_j^{\star}
	\end{aligned}
	\end{equation*}
	Next, we show that the Hessian of the cost function is positive definite. It is easy to see that the Hessian is given by
	\begin{equation*}
	\nabla^2 C_{\alpha}(x_1, \dots, x_{t-1}) =\left[ \begin{matrix} 
	4\alpha & 2\alpha & 2\alpha &\dots & 2\alpha
	\\
	2\alpha & 4\alpha & 2\alpha & \dots & 2\alpha
	\\
	\vdots & \vdots & \vdots & \vdots & \vdots
	\\
	2\alpha & 2\alpha &  2\alpha & \dots & 4\alpha
	\end{matrix}
	\right],
	\end{equation*}
	i.e., all the diagonal elements are $4\alpha$, and all off-diagonal elements are $2\alpha$. For such a matrix with the $n \times n$ order one eigenvalue is $2(n+1)\alpha$, and the remaining eigenvalues are $2\alpha$. So, all the eigenvalues are positive; hence Hessian is positive definite. So, the function is convex. 
	
	To complete the proof, we put back all $x_j^{\star}$'s 
	in the cost function and simplify it further
	\begin{equation*}
	C(x_j^{\star}; j=1,2, \dots, t-1) = \alpha t \left(\frac{n}{t}\right)^2 + n(t-1) \cdot \frac{c_{\min}}{2 \alpha} - \frac{t(t-1)(t+1)}{12} \cdot \frac{c_{\min}^2}{4\alpha^2}.
	\end{equation*}
	This ends the proof.
\end{proof}

Apart from the cost of the optimal departure sequence, we also require the probability of this departure sequence to compute $V^{\star}$. To this end, we recall the feature design of transition probability that allows an agent to stay or depart from the initial node $s_{init}$ to the goal node $g$ iff the sign of the action $\tb{\ti{a}}^i$ taken by agent $i$ matches the sign of the transition probability function parameter $\boldsymbol{\theta}^i$, i.e,  $sgn(\tb{\ti{a}}^i_j) = sgn(\boldsymbol{\theta}^i_j)$ for all $j=1, 2, \dots, d-1$. This is because for each agent $i\in N$, we have $\langle \phi(s^{\prime^{i}} | s^i, \tb{\ti{a}}^i), (\boldsymbol{\theta}^i, \frac{1}{2^{n-1}}) \rangle$ as follows:
\begin{equation}
\left\langle \phi(s^{\prime^{i}} | s^i, \tb{\ti{a}}^i), \left(\boldsymbol{\theta}^i, \frac{1}{2^{n-1}} \right) \right\rangle =
\begin{cases}  
-\langle \tb{\ti{a}}^i, \boldsymbol{\theta}^i \rangle + \frac{1-\delta}{n} \times \frac{1}{2^{n-1}},& ~if ~ {s}^i = s^{\prime^{i}} = s_{init}
\\
\langle \tb{\ti{a}}^i, \boldsymbol{\theta}^i \rangle + \frac{\delta}{n} \times \frac{1}{2^{n-1}},&~ if ~ {s}^i = s_{init},~s^{\prime^{i}} = g
\\
0,&~ if ~ {s}^i = g,~s^{\prime^{i}} = s_{init}
\\
\frac{1}{n} \times \frac{1}{2^{n-1}} & ~ if ~ {s}^i = s^{\prime^{i}} = g.
\end{cases}
\end{equation}
Hence the transition probability is as follows
\begin{equation}
\mathbb{P}(\tb{\ti{s}}^{\prime} | \tb{\ti{s}}, \tb{\ti{a}}) = \sum_{i=1}^n \left\langle \phi(s^{\prime^{i}} | s^i, \tb{\ti{a}}^i), \left(\boldsymbol{\theta}^i, \frac{1}{2^{n-1}} \right) \right\rangle.
\end{equation}

Using the sign matching property, we have the following theorem for the transition probability.

\subsection{Proof of Theorem \ref{thm: transition_prob_optimal}}
\label{app: tranisition_prob_optimal}

Recall the theorem: The transition probability $\mathbb{P}[x_1 =x_1^{\star}, \dots, x_{t}=x_{t}^{\star}]$ is given by
\begin{equation*}
\mathbb{P}[x_1 =x_1^{\star}, \dots, x_{t}=x_{t}^{\star}] = \prod_{k=1}^{t-1}  \Big( 1 - \Big( \gamma n- (\gamma -\eta) \sum_{j=1}^{k} x_{j}^{\star} \Big) \Big) \times \Big( \gamma n- (\gamma - \eta) \sum_{j=1}^{t-1} x_{j}^{\star} \Big),
\end{equation*}
where $\gamma = \left(\frac{\Delta}{n} + \frac{\delta}{n \cdot 2^{n-1}} \right) $ and $\eta = \frac{1}{n\cdot 2^{n-1}}$.

\begin{proof}
	To find the probability of optimal departure sequence, we first find the probability of all agents reaching the goal state $\tb{\ti{g}}$ exactly by $t$ time period starting from initial state $\tb{\ti{s}}_{init}$. 
	Formally, we find the following probability
	\begin{equation*}
	\mathbb{P}[x_1 =x_1^{\star}, \dots, x_{t-1}=x_{t-1}^{\star}, x_t = x_t^{\star}] = \mathbb{P}(\tb{\ti{s}}_{t+1} = \tb{\ti{g}}, \tb{\ti{s}}_t \neq \tb{\ti{g}}, \tb{\ti{s}}_{t-1} \neq \tb{\ti{g}}, \dots, \tb{\ti{s}}_2 \neq \tb{\ti{g}} | \tb{\ti{s}}_1 = \tb{\ti{s}}_{init}, \tb{\ti{a}}_1).
	\end{equation*}
	This is because, the agents will reach to the goal state in exactly $t$ time periods while using $x_1^{\star}, \dots,  x_{t-1}^{\star}, x_t^{\star}$ is the optimal departure sequence. So, $t$ time periods will end if $\tb{\ti{s}}_{t+1} = \tb{\ti{g}}$.
	
	The above probability can be written as
	\begin{eqnarray*}
		& &\mathbb{P}(\tb{\ti{s}}_{t+1} =\tb{\ti{g}}, \tb{\ti{s}}_{t} \neq \tb{\ti{g}}, \dots, \tb{\ti{s}}_2 \neq \tb{\ti{g}} | \tb{\ti{s}}_1 = \tb{\ti{s}}_{init}, \tb{\ti{a}}_1) = \mathbb{P}(\tb{\ti{s}}_2 \neq \tb{\ti{g}} | \tb{\ti{s}}_1 = \tb{\ti{s}}_{init}, \tb{\ti{a}}_1) \times \mathbb{P}( \tb{\ti{s}}_3 \neq \tb{\ti{g}} | \tb{\ti{s}}_1 = \tb{\ti{s}}_{init}, \tb{\ti{s}}_2,  \tb{\ti{a}}_1, \tb{\ti{a}}_2) 
		\\
		& & ~~~\times \dots \times 
		\mathbb{P}(\tb{\ti{s}}_{t} \neq \tb{\ti{g}}| \tb{\ti{s}}_1 = \tb{\ti{s}}_{init}, \tb{\ti{s}}_2, \dots, \tb{\ti{s}}_{t-1}, \tb{\ti{a}}_1, \dots, \tb{\ti{a}}_{t-1}) \times \mathbb{P}(\tb{\ti{s}}_{t+1} =\tb{\ti{g}}| \tb{\ti{s}}_1 = \tb{\ti{s}}_{init}, \tb{\ti{s}}_2, \dots, \tb{\ti{s}}_{t}, \tb{\ti{a}}_1, \dots, \tb{\ti{a}}_{t}).
	\end{eqnarray*}
	This can be written  as follows
	\begin{eqnarray}
	\mathbb{P}(\tb{\ti{s}}_{t+1} =\tb{\ti{g}}, \tb{\ti{s}}_{t} \neq \tb{\ti{g}}, \dots, \tb{\ti{s}}_2 \neq \tb{\ti{g}} | \tb{\ti{s}}_1 = \tb{\ti{s}}_{init}, \tb{\ti{a}}_1) 
	&=& \prod_{k=1}^{t-1}  \mathbb{P}(\tb{\ti{s}}_{k+1} \neq \tb{\ti{g}}| \tb{\ti{s}}_1 = \tb{\ti{s}}_{init}, \tb{\ti{s}}_2, \dots, \tb{\ti{s}}_{k}, \tb{\ti{a}}_1, \dots, \tb{\ti{a}}_k) \nonumber
	\\
	& & \times \mathbb{P}(\tb{\ti{s}}_{t+1} =\tb{\ti{g}}| \tb{\ti{s}}_1 = \tb{\ti{s}}_{init}, \tb{\ti{s}}_2, \dots, \tb{\ti{s}}_{t}, \tb{\ti{a}}_1, \dots, \tb{\ti{a}}_{t}) \nonumber
	\\
	&=& \prod_{k=1}^{t-1}  (1 -\mathbb{P}(\tb{\ti{s}}_{k+1} = \tb{\ti{g}}| \tb{\ti{s}}_1 = \tb{\ti{s}}_{init}, \tb{\ti{s}}_2, \dots, \tb{\ti{s}}_{k}, \tb{\ti{a}}_1, \dots, \tb{\ti{a}}_k)) \nonumber
	\\
	& & \times \mathbb{P}(\tb{\ti{s}}_{t+1} =\tb{\ti{g}}| \tb{\ti{s}}_1 = \tb{\ti{s}}_{init}, \tb{\ti{s}}_2, \dots, \tb{\ti{s}}_{t}, \tb{\ti{a}}_1, \dots, \tb{\ti{a}}_{t}). \label{eqn: prob_1}
	\end{eqnarray}
	To find this probability first consider the following probability for any $k = 1,2, \dots, t-1$
	\begin{align}
	\mathbb{P}(\tb{\ti{s}}_{k+1} = \tb{\ti{g}}| \tb{\ti{s}}_1 = \tb{\ti{s}}_{init}, \tb{\ti{s}}_2, \dots, \tb{\ti{s}}_{k}, \tb{\ti{a}}_1, \dots, \tb{\ti{a}}_{k})) 
	&= \sum_{\{i: s^i_k = s_{init}\}} \left( \langle \tb{\ti{a}}^i_k, \boldsymbol{\theta}^i_k \rangle + \frac{\delta}{n \cdot 2^{n-1}} \right) + \sum_{\{i:s^i_k = g\}} \left( \frac{1}{n\cdot 2^{n-1}} \right) \nonumber
	\\
	& = \sum_{\{i: s^i_k = s_{init}\}} \left( \frac{\Delta}{n} + \frac{\delta}{n \cdot 2^{n-1}} \right) + \sum_{\{i:s^i_k = g\}} \left( \frac{1}{n\cdot 2^{n-1}} \right). \label{eqn: prob_2}
	\end{align}
	The above equation uses the definition of transition probability, depending on the number of agents that have moved from the initial node to the goal node and the number of agents already at the goal node. Moreover, we also note that the agent $i$ will move to goal node from the initial node with highest probability if the sign of $\tb{\ti{a}}^i_k$ matches the sign of $\boldsymbol{\theta}^i_k$ for each component and for all $k=1, 2, \dots, t-1$. Thus, $\langle  \tb{\ti{a}}^i_k, \boldsymbol{\theta}^i_k \rangle = \frac{\Delta}{n}$, for all $k=1, 2, \dots, t-1$.
	
	Substituting Eq. \eqref{eqn: prob_2} in Eq. \eqref{eqn: prob_1}, we have the following
	\begin{align}
	& \mathbb{P}(\tb{\ti{s}}_{t+1} =\tb{\ti{g}}, \tb{\ti{s}}_{t} \neq \tb{\ti{g}}, \dots, \tb{\ti{s}}_2 \neq \tb{\ti{g}} | \tb{\ti{s}}_1 = \tb{\ti{s}}_{init}, \tb{\ti{a}}_1)  \nonumber
	\\
	&= \prod_{k=1}^{t-1}  (1 -\mathbb{P}(\tb{\ti{s}}_{k+1} = \tb{\ti{g}}| \tb{\ti{s}}_1 = \tb{\ti{s}}_{init}, \tb{\ti{s}}_2, \dots, \tb{\ti{s}}_{k}, \tb{\ti{a}}_1, \dots, \tb{\ti{a}}_{k})) \nonumber
	\\
	&~~~~~  \times \mathbb{P}(\tb{\ti{s}}_{t+1} =\tb{\ti{g}}| \tb{\ti{s}}_1 = \tb{\ti{s}}_{init}, \tb{\ti{s}}_2, \dots, \tb{\ti{s}}_{t}, \tb{\ti{a}}_1, \dots, \tb{\ti{a}}_{t}) \nonumber
	\\
	&= \prod_{k=1}^{t-1}  \left (1 - \left\lbrace \sum_{\{i: s^i_k = s_{init}\}} \left( \frac{\Delta}{n} + \frac{\delta}{n \cdot 2^{n-1}} \right) + \sum_{\{i:s^i_k = g\}} \left( \frac{1}{n\cdot 2^{n-1}} \right) \right\rbrace \right) \nonumber
	\\
	&~~~~~  \times \sum_{\{i: s^i_{t-1} = s_{init}\}} \left( \frac{\Delta}{n} + \frac{\delta}{n \cdot 2^{n-1}} \right) + \sum_{\{i:s^i_{t-1} = g\}} \left( \frac{1}{n\cdot 2^{n-1}} \right) \nonumber
	\\
	&= \prod_{k=1}^{t-1}  \left( 1 - \left\lbrace \left( n- \sum_{j=1}^{k} x_{j} \right) \cdot \left( \frac{\Delta}{n} + \frac{\delta}{n \cdot 2^{n-1}} \right) + \left( \sum_{j=1}^{k} x_j \right) \cdot \left(\frac{1}{n\cdot 2^{n-1}} \right) \right\rbrace \right)  \nonumber
	\\
	&~~~~~  \times \left\lbrace \left( n- \sum_{j=1}^{t-1} x_{j} \right) \cdot  \left(\frac{\Delta}{n} + \frac{\delta}{n \cdot 2^{n-1}} \right) + \left( \sum_{j=1}^{k} x_j \right) \cdot \left(\frac{1}{n\cdot 2^{n-1}} \right) \right\rbrace, \nonumber
	\end{align}
	where in the last inequality, we use the fact that the departure sequence is $x_1, x_2, \dots, x_{t-1}$. 
	
	The above probability is the same as the probability of optimal departure sequence $x_j^{\star}$, i.e., 
	\begin{equation*}
	\mathbb{P}[x_1 =x_1^{\star}, \dots, x_{t-1}=x_{t-1}^{\star}, x_t = x_t^{\star}] = \prod_{k=1}^{t-1}  \Big( 1 - \Big( \gamma n- (\gamma -\eta) \sum_{j=1}^{k} x_{j}^{\star} \Big) \Big) \times \Big( \gamma n- (\gamma - \eta) \sum_{j=1}^{t-1} x_{j}^{\star} \Big),
	\end{equation*}
	where $\gamma = \left(\frac{\Delta}{n} + \frac{\delta}{n \cdot 2^{n-1}} \right) $ and $\eta = \frac{1}{n\cdot 2^{n-1}}$.
\end{proof}
Since it is hard to get the closed form expression of the above optimal value
we obtain an approximation of the optimal value function for regret computation. So, we define the approximate optimal value in terms of a given large $T < \infty$, as 
\begin{equation*}
V^{\star}_{T} = \sum_{t=1}^{
	T} \mathbb{P}[x_1^{\star}, \dots, x_{t}^{\star}] \cdot C_{\alpha}(x_1^{\star}, \dots, x_{t}^{\star}).
\end{equation*}
We use this approximate value in the computations, where $T$ is tuned suitably.

\section{PROOF OF THE INTERMEDIATE LEMMAS AND PROPOSITIONS}
\label{app: inter-lemmas}
In this section, we will provide the proof details of Lemmas and Propositions  stated in this appendix.

\subsection{Proof of Lemma \ref{lemma: interim_regret_thm_regret}}
\label{sec: regret-proposition}

\begin{proof}
	The proof of this lemma involves three major steps: 1) the convergence of the cost function parameters (Theorem \ref{thm: weight_convergence}); 2) convergence of the optimistic estimator of the state-action value function whenever MAEVI is triggered (Theorem \ref{thm: MAEVI_analysis}); 3) the regret given in Equation \eqref{eqn: regret_expression_main_thm} will be decomposed into two components (Theorem \ref{thm: regret_decomposition}), and each component will be bounded (Propositions \ref{prop: E1_bound}, \ref{prop: E2_bound}). This decomposition further depends on the agents for whom the MAEVI is triggered or not. The decomposition of regret
	in this manner is novel to our  multi-agent congestion cost minimization. To best of our knowledge this is not available in literature.
	
	To complete the proof, we need to bound both $E_1$ and $E_2$. The following proposition gives the bound to $E_1$, which uses \emph{all the
		intrinsic properties} of our algorithm design. 
	\begin{prop}
		\label{prop: E1_bound}
		The term $E_1$ is upper bounded as follows
		\begin{align}
		E_1 \leq 
		8 \beta_T \sqrt{2 T d \log \left( 1 +\frac{ T B_{\star}^2}{\lambda} \right) } + 8 nd (B_{\star}+1) \left[\log \left( 1 + \frac{T B_{\star}^2 nd}{\lambda} \right) + 2 \log(T) \right] + 4 (n+1).
		\label{eqn: E1_bound_final}
		\end{align}
	\end{prop}
	Proof of this proposition is available in Section \ref{app: bound_E1} of this appendix. Next, we bound the other term $E_2$, which uses the Azuma-Hoeffding inequality. The following proposition provides the bound to $E_2$.
	\begin{prop}
		\label{prop: E2_bound}
		The term $E_2$ is upper bounded as follows
		\begin{equation}
		E_2 = \frac{1}{n} \sum_{i=1}^n 2B_{\star} \sqrt{2T \log \left(\frac{2nT}{\delta} \right)} = 2B_{\star} \sqrt{2T \log \left(\frac{2nT}{\delta} \right)}.
		\label{eqn: bound_E2_final}
		\end{equation}
	\end{prop}
	The proof of this lemma is present in Section \ref{app: bound_E2}. The proof of Lemma \ref{lemma: interim_regret_thm_regret} follows by substituting bounds of $E_1$ and $E_2$ from Equations \eqref{eqn: E1_bound_final} and \eqref{eqn: bound_E2_final} in Theorem \ref{thm: regret_decomposition},
	\begin{equation*}
	\begin{aligned}
	R(M)  &\leq 8 \beta_T \sqrt{2 T d \log \left( 1 +\frac{ T B_{\star}^2}{\lambda} \right) } + 8 nd (B_{\star}+1) \left[\log \left( 1 + \frac{T B_{\star}^2 nd}{\lambda} \right) + 2 \log(T) \right] + 4 (n+1)
	\\
	& ~~~ +2B_{\star} \sqrt{2T \log \left(\frac{2nT}{\delta} \right)} + 2n^2dB_{\star} \log \left(1+ \frac{TB_{\star}^2 nd}{\lambda} \right) + 2n B_{\star} \log(T) + 2.
	\end{aligned}
	\end{equation*}
	Combining the lower order terms, we have
	\begin{equation*}
	R_K \leq 10 \beta_T \sqrt{T d \log \left( 1 +\frac{ T B_{\star}^2}{\lambda} \right) } + 16n^2dB_{\star} \log \left(T+ \frac{T^2B_{\star}^2 nd}{\lambda} \right).
	\end{equation*}
	This ends the proof of Lemma \ref{lemma: interim_regret_thm_regret}.
\end{proof}


\subsection{Proof of Proposition \ref{prop: regret_decom_lemma}}
\label{proof: reg_dec_initial_lemma}
\begin{proof}
	The proof uses the ideas for the single agent SSP in \cite{tarbouriech2020no} and \cite{min2022learning}; however, for the multi-agent setup, we have some extra challenges. Consider the following
	\begin{eqnarray}
	& & \sum_{m=1}^M \left( \sum_{h=1}^{H_m} \left\lbrace \frac{1}{n} \sum_{i=1}^n V^i_{j_i(m)} (\tb{\ti{s}}_{m,h}) - \frac{1}{n} \sum_{i=1}^n V^i_{j_i(m)}(\tb{\ti{s}}_{m,h+1}) \right\rbrace \right) \nonumber
	\\
	&\overset{(i)}{=}& \sum_{m=1}^M \left( \frac{1}{n} \sum_{i=1}^n V^i_{j_i(m)} (\tb{\ti{s}}_{m,1}) - \frac{1}{n} \sum_{i=1}^n V^i_{j_i(m)}(\tb{\ti{s}}_{m,H_{m}+1})\right) \nonumber
	\\
	&\overset{(ii)}{=}& \sum_{m=1}^{M-1} \left( \frac{1}{n} \sum_{i=1}^n V^i_{j_i(m+1)} (\tb{\ti{s}}_{m+1,1}) - \frac{1}{n} \sum_{i=1}^n V^i_{j_i(m)}(\tb{\ti{s}}_{m,H_{m}+1}) \right) \nonumber
	\\
	& & + \sum_{m=1}^{M-1} \left( \frac{1}{n} \sum_{i=1}^n V^i_{j_i(m)} (\tb{\ti{s}}_{m,1}) - \frac{1}{n} \sum_{i=1}^n V^i_{j_i(m+1)} (\tb{\ti{s}}_{m+1,1}) \right)  \nonumber
	\\
	& & +\frac{1}{n} \sum_{i=1}^n V^i_{j_i(M)}(\tb{\ti{s}}_{M,1}) - \frac{1}{n} \sum_{i=1}^n V^i_{j_i(M)} (\tb{\ti{s}}_{M, H_M + 1}) \nonumber
	\\
	&\overset{(iii)}{=}& \sum_{m=1}^{M-1} \left( \frac{1}{n} \sum_{i=1}^n V^i_{j_i(m+1)} (\tb{\ti{s}}_{m+1,1}) - \frac{1}{n} \sum_{i=1}^n V^i_{j_i(m)}(\tb{\ti{s}}_{m,H_{m}+1}) \right) \nonumber
	\\
	& & +\frac{1}{n} \sum_{i=1}^n V^i_{j_i(1)}(\tb{\ti{s}}_{1,1}) - \frac{1}{n} \sum_{i=1}^n V^i_{j_i(M)}(\tb{\ti{s}}_{M,1}) \nonumber
	\\
	& & + \frac{1}{n} \sum_{i=1}^n V^i_{j_i(M)}(\tb{\ti{s}}_{M,1}) - \frac{1}{n} \sum_{i=1}^n V^i_{j_i(M)} (\tb{\ti{s}}_{M, H_M + 1}) \nonumber
	\\
	&=& \sum_{m=1}^{M-1} \left( \frac{1}{n} \sum_{i=1}^n V^i_{j_i(m+1)} (\tb{\ti{s}}_{m+1,1}) - \frac{1}{n} \sum_{i=1}^n V^i_{j_i(m)}(\tb{\ti{s}}_{m,H_{m}+1}) \right) \nonumber
	\\
	& & + \frac{1}{n} \sum_{i=1}^n V^i_{j_i(1)}(\tb{\ti{s}}_{1,1})  - \frac{1}{n} \sum_{i=1}^n V^i_{j_i(M)} (\tb{\ti{s}}_{M, H_M + 1}) \nonumber
	\\
	&\overset{(iv)}{\leq}& \sum_{m=1}^{M-1} \left( \frac{1}{n} \sum_{i=1}^n V^i_{j_i(m+1)} (\tb{\ti{s}}_{m+1,1}) - \frac{1}{n} \sum_{i=1}^n V^i_{j_i(m)}(\tb{\ti{s}}_{m,H_{m}+1}) \right) + \frac{1}{n} \sum_{i=1}^n V^i_{j_i(1)}(\tb{\ti{s}}_{1,1}),  \label{eqn: regret_decomposition_first_step}  
	\end{eqnarray}
	where $(i)$ follows from the telescopic summation over $h$; in $(ii)$ we add and subtract $\frac{1}{n} \sum_{i=1}^n V^i_{j_i(m+1)} (\tb{\ti{s}}_{m+1,1})$ inside the summation. $(iii)$ again uses the telescopic summation. Finally, $(iv)$ follows by dropping a non-negative term with a negative sign.
	
	Now consider the first term of the RHS of the Equation \eqref{eqn: regret_decomposition_first_step}. Note that the interval ends if either of the two conditions are satisfied: 1) MAEVI is triggered by at least one agent $i\in N$; 2) all the agents reach the goal state.
	Let us suppose, all the agents reach the goal state hence $\tb{\ti{s}}_{m+1,1} = \tb{\ti{s}}_{init}$, and $\tb{\ti{s}}_{m, H_m + 1} = \tb{\ti{g}}$. This implies 
	\begin{eqnarray}
	\frac{1}{n} \sum_{i=1}^n V^i_{j_i(m+1)} (\tb{\ti{s}}_{m+1,1}) - \frac{1}{n} \sum_{i=1}^n V^i_{j_i(m)}(\tb{\ti{s}}_{m,H_{m}+1})  
	&=& \frac{1}{n} \sum_{i=1}^n V^i_{j_i(m+1)} (\tb{\ti{s}}_{init}) - \frac{1}{n} \sum_{i=1}^n V^i_{j_i(m)}(\tb{\ti{g}})  \nonumber
	\\
	&=& \frac{1}{n} \sum_{i=1}^n V^i_{j_i(m+1)} (\tb{\ti{s}}_{init}).
	\label{eqn: regret_decomposition_interval_end_goal_state}
	\end{eqnarray}
	Next, suppose the interval ended because MAEVI is triggered for some agent $i\in N$. Then we apply a trivial upper bound, i.e., 
	\begin{equation}
	\label{eqn: regret_decomposition_interval_end_evi_triggered}
	\frac{1}{n} \sum_{i=1}^n V^i_{j_i(m+1)} (\tb{\ti{s}}_{m+1,1}) - \frac{1}{n} \sum_{i=1}^n V^i_{j_i(m)}(\tb{\ti{s}}_{m,H_{m}+1})   \leq \frac{1}{n} \sum_{i=1}^n \max_{j_i} ||V^i_{j_i}||_{\infty}. 
	\end{equation}
	Also recall the total number of calls to the MAEVI algorithm from Lemma \ref{lemma: calls_to_evi} is given by
	\begin{equation*}
	J \leq  2 n^2d \log \left( 1 + \frac{T B_{\star}^2 nd}{\lambda} \right) + 2n \log(T).
	\end{equation*}
	Thus, from Equation \eqref{eqn: regret_decomposition_first_step} we have,
	\begin{eqnarray}
	& &\sum_{m=1}^M \left( \sum_{h=1}^{H_m} \left\lbrace \frac{1}{n} \sum_{i=1}^n V^i_{j_i(m)} (\tb{\ti{s}}_{m,h}) - \frac{1}{n} \sum_{i=1}^n V^i_{j_i(m)}(\tb{\ti{s}}_{m,h+1}) \right\rbrace \right)  \nonumber
	\\
	& \leq & \sum_{m=1}^{M-1} \left( \frac{1}{n} \sum_{i=1}^n V^i_{j_i(m+1)} (\tb{\ti{s}}_{m+1,1}) - \frac{1}{n} \sum_{i=1}^n V^i_{j_i(m)}(\tb{\ti{s}}_{m,H_{m}+1}) \right) + \frac{1}{n} \sum_{i=1}^n V^i_{j_i(1)}(\tb{\ti{s}}_{1,1}) \nonumber
	\\
	& \overset{(i)}{\leq} &  \sum_{m=1}^{M-1} \left(  \frac{1}{n} \sum_{i=1}^n V^i_{j_i(m+1)} (\tb{\ti{s}}_{init}) \cdot \mathds{1}_{\{m+1 \in \mathcal{M}(M)\}} \right)  + \frac{1}{n} \sum_{i=1}^n V^i_{j_i(1)}(\tb{\ti{s}}_{1,1}) \nonumber
	\\
	& & + \left[ 2 n^2d \log \left( 1 + \frac{T B_{\star}^2 nd}{\lambda} \right) + 2n \log(T) \right] \cdot \frac{1}{n} \sum_{i=1}^n \max_{j_i} ||V^i_{j_i}||_{\infty}  \nonumber
	\\
	&\overset{(ii)}{\leq} & \sum_{m \in \mathcal{M}(M)} \left(  \frac{1}{n} \sum_{i=1}^n V^i_{j_i(m)} (\tb{\ti{s}}_{init}) \right) + \frac{1}{n} \sum_{i=1}^n V^i_{0}(\tb{\ti{s}}_{init}) \nonumber
	\\
	& & + \left[ 2 n^2d \log \left( 1 + \frac{T B_{\star}^2 nd}{\lambda} \right) + 2n \log(T) \right] \cdot \frac{1}{n} \sum_{i=1}^n B_{\star}  \nonumber
	\\
	& \overset{(iii)}{=} & \sum_{m \in \mathcal{M}(M)} \left(  \frac{1}{n} \sum_{i=1}^n V^i_{j_i(m)} (\tb{\ti{s}}_{init}) \right) + 1 + 2 n^2d B_{\star} \log \left( 1 + \frac{T B_{\star}^2 nd}{\lambda} \right) + 2nB_{\star} \log(T), \label{eqn: regret_decomposition_second_step}
	\end{eqnarray}
	where the first inequality is the same as Equation \eqref{eqn: regret_decomposition_first_step}. $(i)$ uses the combination of Equations \eqref{eqn: regret_decomposition_interval_end_goal_state} and \eqref{eqn: regret_decomposition_interval_end_evi_triggered} along with the fact that if the goal state is reached by all the agents, then $m+1 \in \mathcal{M}(M)$. In $(ii)$ we use the fact that $V^i_{j_i(1)}(\tb{\ti{s}}_{1,1}) = V^i_0(\tb{\ti{s}}_{init})$ for all agents $i\in N$. Also, $||V^i_{j_i}||_{\infty} \leq B_{\star}$ for all $i\in N$. Finally $(iii)$ follows from the fact that $V^i_0(\tb{\ti{s}}_{init}) = 1$ for all $i \in N$. The proof follows by arranging the terms of the above Equation \eqref{eqn: regret_decomposition_second_step}.
\end{proof}

\subsection{Proof of Proposition \ref{prop: E1_bound} (Bounding $E_1$)}
\label{app: bound_E1}

\begin{proof}
	Recall, from Equation \eqref{eqn: regret_decomposition} we have 
	\begin{equation*}
	E_1 = \sum_{m=1}^M \sum_{h=1}^{H_m} \left[ \frac{1}{n} \sum_{i=1}^n \bar{c}(\tb{\ti{s}}_{m,h}, \tb{\ti{a}}_{m,h}; \tb{\ti{w}}^i) + \frac{1}{n} \sum_{i=1}^n \mathbb{P}V^i_{j_i(m)}(\tb{\ti{s}}_{m,h}, \tb{\ti{a}}_{m,h}) - \frac{1}{n} \sum_{i=1}^n V^i_{j_i(m)}(\tb{\ti{s}}_{m,h}) \right].
	\end{equation*}
	First note that $V^i_{j_i(m)}(\tb{\ti{s}}_{m,h})  = \min_a Q^i_{j_i(m)}(\tb{\ti{s}}_{m,h}, a)  = Q^i_{j_i(m)}(\tb{\ti{s}}_{m,h}, \tb{\ti{a}}_{m,h})$, therefore $E_1$ can be written as
	\begin{equation*}
	E_1 = \sum_{m=1}^M \sum_{h=1}^{H_m} \left[ \frac{1}{n} \sum_{i=1}^n \bar{c}(\tb{\ti{s}}_{m,h}, \tb{\ti{a}}_{m,h}; \tb{\ti{w}}^i) + \frac{1}{n} \sum_{i=1}^n \mathbb{P}V^i_{j_i(m)}(\tb{\ti{s}}_{m,h}, \tb{\ti{a}}_{m,h}) - \frac{1}{n} \sum_{i=1}^n Q^i_{j_i(m)}(\tb{\ti{s}}_{m,h}, \tb{\ti{a}}_{m,h}) \right].
	\end{equation*}
	
	Since the interval, $m$ ends if the MAEVI is triggered for at least one agent or all the agents reach the goal state. Let $S_m$ be the set of agents for whom the MAEVI is triggered in the interval $m$. We can decompose the above summation over $S_m$ and $S^c_m$. In particular, we have
	\begin{eqnarray*}
		E_1 &=& \sum_{m=1}^M \sum_{h=1}^{H_m} \left[ \frac{1}{n} \sum_{i\in S_m} \bar{c}(\tb{\ti{s}}_{m,h}, \tb{\ti{a}}_{m,h}; \tb{\ti{w}}^i) + \frac{1}{n} \sum_{i\in S_m} \mathbb{P}V^i_{j_i(m)}(\tb{\ti{s}}_{m,h}, \tb{\ti{a}}_{m,h}) - \frac{1}{n} \sum_{i\in S_m} Q^i_{j_i(m)}(\tb{\ti{s}}_{m,h}, \tb{\ti{a}}_{m,h}) \right]
		\\
		& & + \sum_{m=1}^M \sum_{h=1}^{H_m} \left[ \frac{1}{n} \sum_{i\in S^c_m} \bar{c}(\tb{\ti{s}}_{m,h}, \tb{\ti{a}}_{m,h}; \tb{\ti{w}}^i) + \frac{1}{n} \sum_{i\in S^c_m} \mathbb{P}V^i_{j_i(m)}(\tb{\ti{s}}_{m,h}, \tb{\ti{a}}_{m,h}) - \frac{1}{n} \sum_{i\in S^c_m} Q^i_{j_i(m)}(\tb{\ti{s}}_{m,h}, \tb{\ti{a}}_{m,h}) \right]
		\\
		&=&\sum_{m=1}^M \sum_{h=1}^{H_m} [ E_1(S_m) + E_1(S^c_m)].
	\end{eqnarray*}
	To bound $E_1$, we need to bound both $E_1(S_m)$ and $E_1(S^c_m)$ for each $m$. \emph{As opposed to the single agent SSP of \cite{min2022learning}, we need separate bounds for $E_1(S_m)$ and $E_1(S^c_m)$.}  First consider $E_1(S_m)$.
	
	\textbf{Bounding $E_1(S_m)$}
	
	First consider $E_1(S_m)$. To bound this, we will use the MAEVI update for each agent $i\in S_m$. Recall that the MAEVI update is given by
	\begin{align*}
	Q^{i, (l)} (\tb{\ti{s}}_{m,h}, \tb{\ti{a}}_{m,h}) &= \bar{c}(\tb{\ti{s}}_{m,h}, \tb{\ti{a}}_{m,h}; \tb{\ti{w}}^i) + (1-q) \min_{\boldsymbol{\theta} \in \mathcal{C}^i_{j_i(m)} \cap \mathcal{B}} \left\langle \boldsymbol{\theta}, \phi_{V^{i, (l-1)}} (\tb{\ti{s}}_{m,h}) \right\rangle
	\\
	&= \bar{c}(\tb{\ti{s}}_{m,h}, \tb{\ti{a}}_{m,h}; \tb{\ti{w}}^i) + (1-q) \cdot \left\langle \boldsymbol{\theta}_{m,h}, \phi_{V^{i, (l-1)}} (\tb{\ti{s}}_{m,h}) \right\rangle
	\\
	&= \bar{c}(\tb{\ti{s}}_{m,h}, \tb{\ti{a}}_{m,h};\tb{\ti{w}}^i) + (1-q) \cdot \left\langle \boldsymbol{\theta}_{m,h}, \phi_{V^{i, (l)}} (\tb{\ti{s}}_{m,h}) \right\rangle + (1-q) \cdot \left\langle \boldsymbol{\theta}_{m,h}, [\phi_{V^{i, (l-1)}} - \phi_{V^{i, (l)}}]  (\tb{\ti{s}}_{m,h}) \right\rangle
	\\
	&= \bar{c}(\tb{\ti{s}}_{m,h}, \tb{\ti{a}}_{m,h}; \tb{\ti{w}}^i) + (1-q) \cdot \left\langle \boldsymbol{\theta}_{m,h}, \phi_{V^{i, (l)}} (\tb{\ti{s}}_{m,h}) \right\rangle + (1-q) \cdot \mathbb{P}_{m,h} [V^{i, (l-1)} - V^{i, (l)}]  (\tb{\ti{s}}_{m,h})
	\\
	&= \bar{c}(\tb{\ti{s}}_{m,h}, \tb{\ti{a}}_{m,h}; \tb{\ti{w}}^i) + (1-q) \cdot \mathbb{P}_{m,h} V^{i, (l)} (\tb{\ti{s}}_{m,h}) + (1-q) \cdot \mathbb{P}_{m,h} [V^{i, (l-1)} - V^{i, (l)}]  (\tb{\ti{s}}_{m,h})
	\\
	&\geq  \bar{c}(\tb{\ti{s}}_{m,h}, \tb{\ti{a}}_{m,h}; \tb{\ti{w}}^i) + (1-q) \cdot \mathbb{P}_{m,h} V^{i, (l)} (\tb{\ti{s}}_{m,h}) -  (1-q) \cdot \frac{1}{t_{j_i(m)}},
	\end{align*}
	where the last inequality follows from the fact that the stopping criteria for the MAEVI for agent $i\in S_m$ is $||V^{i, (l)} - V^{i, (l-1)} ||_{\infty} \leq \epsilon^i_{j_i} = \frac{1}{t_{j_i(m)}}$. So, the above inequality implies that
	\begin{equation*}
	\bar{c}(\tb{\ti{s}}_{m,h}, \tb{\ti{a}}_{m,h}; \tb{\ti{w}}^i) - Q^{i, (l)} (\tb{\ti{s}}_{m,h}, \tb{\ti{a}}_{m,h}) \leq  (1-q) \cdot \frac{1}{t_{j_i(m)}}   - (1-q) \cdot \mathbb{P}_{m,h} V^{i, (l)} (\tb{\ti{s}}_{m,h}). 
	\end{equation*}
	Adding $\mathbb{P}V^i_{j_i(m)}(\tb{\ti{s}}_{m,h}, \tb{\ti{a}}_{m,h})$  on both sides in the above equation, we have 
	\begin{align}
	& \bar{c}(\tb{\ti{s}}_{m,h}, \tb{\ti{a}}_{m,h}; \tb{\ti{w}}^i) + \mathbb{P}V^i_{j_i(m)}(\tb{\ti{s}}_{m,h}, \tb{\ti{a}}_{m,h}) - Q^{i, (l)} (\tb{\ti{s}}_{m,h}, \tb{\ti{a}}_{m,h})  \nonumber
	\\
	& \leq  \mathbb{P}V^i_{j_i(m)}(\tb{\ti{s}}_{m,h}, \tb{\ti{a}}_{m,h}) +
	(1-q) \cdot \frac{1}{t_{j_i(m)}}  - (1-q) \cdot \mathbb{P}_{m,h} V^{i, (l)} (\tb{\ti{s}}_{m,h}) \nonumber
	\\
	&= [\mathbb{P} - \mathbb{P}_{m,h}] V^i_{j_i(m)}(\tb{\ti{s}}_{m,h}, \tb{\ti{a}}_{m,h}) + q\cdot \mathbb{P}_{m,h} V^{i, (l)} (\tb{\ti{s}}_{m,h}) + (1-q) \cdot \frac{1}{t_{j_i(m)}} \nonumber
	\\
	&\overset{(i)}{\leq}  [\mathbb{P} - \mathbb{P}_{m,h}] V^i_{j_i(m)}(\tb{\ti{s}}_{m,h}, \tb{\ti{a}}_{m,h}) + qB_{\star} + (1-q) \cdot \frac{1}{t_{j_i(m)}} \nonumber
	\\
	&\overset{(ii)}{=} [\mathbb{P} - \mathbb{P}_{m,h}] V^i_{j_i(m)}(\tb{\ti{s}}_{m,h}, \tb{\ti{a}}_{m,h}) + \frac{B_{\star}}{t_{j_i(m)}} + (1-q) \cdot \frac{1}{t_{j_i(m)}} \nonumber
	\\
	&\overset{(iii)}{\leq}  \left\langle \boldsymbol{\theta}^{\star} - \boldsymbol{\theta}_{m,h}, \phi_{V^i_{j_i(m)}} (\tb{\ti{s}}_{m,h}, \tb{\ti{a}}_{m,h})  \right\rangle  + \frac{B_{\star} + 1}{t_{j_i(m)}},  \label{eqn: e_1_s_m_interim}
	\end{align}
	where $(i)$ uses the fact that $V^i_{j_i(m)} \leq V^{i\star} \leq B_{\star}$, $(ii)$ is consequence of the fact that $q = \frac{1}{t_{j_i(m)}}$, and $(iii)$ follows by dropping a  negative term. Now taking summation over $i\in S_m$, we have the following:
	\begin{eqnarray}
	& & \frac{1}{n} \sum_{i\in S_m} \bar{c}(\tb{\ti{s}}_{m,h}, \tb{\ti{a}}_{m,h}; \tb{\ti{w}}^i) + \frac{1}{n} \sum_{i\in S_m} \mathbb{P}V^i_{j_i(m)}(\tb{\ti{s}}_{m,h}, \tb{\ti{a}}_{m,h}) - \frac{1}{n} \sum_{i\in S_m} Q^{i, (l)} (\tb{\ti{s}}_{m,h}, \tb{\ti{a}}_{m,h}) \nonumber
	\\
	& & \leq \left\langle \boldsymbol{\theta}^{\star} - \boldsymbol{\theta}_{m,h}, \frac{1}{n} \sum_{i\in S_m} \phi_{V^i_{j_i(m)}} (\tb{\ti{s}}_{m,h}, \tb{\ti{a}}_{m,h})  \right\rangle  + \frac{1}{n} \sum_{i\in S_m} \frac{B_{\star} + 1}{t_{j_i(m)}}. \nonumber
	\end{eqnarray}
	
	Let $\mathcal{M}_0(M) = \{m \leq M: j_i(m) \geq 1,~\forall~i\in N \}$ be the set of all intervals for which the output of MAEVI algorithm is $Q^i_{j_i(m)}$ for all agents $i\in N$ rather than the output $Q^i_0$. We first consider the intervals from $\mathcal{M}_0(M)$. So, from the above equation, we have 
	\begin{eqnarray}
	& & \sum_{m\in \mathcal{M}_0(M)} \sum_{h=1}^{H_m} \left[\frac{1}{n} \sum_{i\in S_m} \bar{c}(\tb{\ti{s}}_{m,h}, \tb{\ti{a}}_{m,h}; \tb{\ti{w}}^i) + \frac{1}{n} \sum_{i\in S_m} \mathbb{P}V^i_{j_i(m)}(\tb{\ti{s}}_{m,h}, \tb{\ti{a}}_{m,h}) - \frac{1}{n} \sum_{i\in S_m} Q^{i, (l)} (\tb{\ti{s}}_{m,h}, \tb{\ti{a}}_{m,h}) \right] \nonumber
	\\
	& & \leq \sum_{m\in \mathcal{M}_0(M)} \sum_{h=1}^{H_m} \left[ \left\langle \boldsymbol{\theta}^{\star} - \boldsymbol{\theta}_{m,h}, \frac{1}{n} \sum_{i\in S_m} \phi_{V^i_{j_i(m)}} (\tb{\ti{s}}_{m,h}, \tb{\ti{a}}_{m,h})  \right\rangle  + \frac{1}{n} \sum_{i\in S_m} \frac{B_{\star} + 1}{t_{j_i(m)}} \right] \nonumber
	\\
	&=& \underbrace{\sum_{m\in \mathcal{M}_0(M)} \sum_{h=1}^{H_m} \left[ \left\langle \boldsymbol{\theta}^{\star} - \boldsymbol{\theta}_{m,h}, \frac{1}{n} \sum_{i\in S_m} \phi_{V^i_{j_i(m)}} (\tb{\ti{s}}_{m,h}, \tb{\ti{a}}_{m,h})  \right\rangle \right]}_{A_1} 
	+ \underbrace{\sum_{m\in \mathcal{M}_0(M)} \sum_{h=1}^{H_m} \left[ \frac{1}{n} \sum_{i\in S_m} \frac{B_{\star} + 1}{t_{j_i(m)}} \right]}_{A_2}. \label{eqn: E1(S_m)_part_1}
	\end{eqnarray}
	To bound the above, we need to bound $A_1$ and $A_2$. First consider $A_1$. 
	\begin{align}
	A_1 &= \sum_{m\in \mathcal{M}_0(M)} \sum_{h=1}^{H_m} \left[ \left\langle \boldsymbol{\theta}^{\star} - \boldsymbol{\theta}_{m,h}, \frac{1}{n} \sum_{i\in S_m} \phi_{V^i_{j_i(m)}} (\tb{\ti{s}}_{m,h}, \tb{\ti{a}}_{m,h})  \right\rangle \right]  \nonumber
	\\
	&= \sum_{m\in \mathcal{M}_0(M)} \sum_{h=1}^{H_m} \left[ \frac{1}{n} \sum_{i\in S_m} \left\langle \boldsymbol{\theta}^{\star} - \boldsymbol{\theta}_{m,h},   \phi_{V^i_{j_i(m)}} (\tb{\ti{s}}_{m,h}, \tb{\ti{a}}_{m,h})  \right\rangle \right]. \label{eqn: A1}
	\end{align}
	To bound the above term, consider the inner term in the above equation for agent $i\in S_m$, 
	\begin{align}
	&  \left\langle \boldsymbol{\theta}^{\star} - \boldsymbol{\theta}_{m,h},   \phi_{V^i_{j_i(m)}} (\tb{\ti{s}}_{m,h}, \tb{\ti{a}}_{m,h})  \right\rangle \nonumber
	\\
	&  \overset{(i)}{\leq} || \boldsymbol{\theta}^{\star} - \boldsymbol{\theta}_{m,h} ||_{{\mathbf{\Sigma}}^i_{t(m,h)}} \cdot || \phi_{V^i_{j_i(m)}} (\tb{\ti{s}}_{m,h}, \tb{\ti{a}}_{m,h})  ||_{{\mathbf{\Sigma}}^{i^{-1}}_{t(m,h)}}  \nonumber
	\\
	& = || \boldsymbol{\theta}^{\star} + \hat{\boldsymbol{\theta}}^i_{j_i(m)} - \hat{\boldsymbol{\theta}}^i_{j_i(m)} - \boldsymbol{\theta}_{m,h} ||_{{\mathbf{\Sigma}}^i_{t(m,h)}} \cdot || \phi_{V^i_{j_i(m)}} (\tb{\ti{s}}_{m,h}, \tb{\ti{a}}_{m,h})  ||_{{\mathbf{\Sigma}}^{i^{-1}}_{t(m,h)}}   \nonumber
	\\
	& \overset{(ii)}{\leq} \left( || \boldsymbol{\theta}^{\star} - \hat{\boldsymbol{\theta}}^i_{j_i(m)}||_{{\mathbf{\Sigma}}^i_{t(m,h)}} +  || \hat{\boldsymbol{\theta}}^i_{j_i(m)} - \boldsymbol{\theta}_{m,h} ||_{{\mathbf{\Sigma}}^i_{t(m,h)}} \right) \cdot || \phi_{V^i_{j_i(m)}} (\tb{\ti{s}}_{m,h}, \tb{\ti{a}}_{m,h})  ||_{{\mathbf{\Sigma}}^{i^{-1}}_{t(m,h)}}   \nonumber
	\\ 
	& \overset{(iii)}{\leq} 2 \left( || \boldsymbol{\theta}^{\star} - \hat{\boldsymbol{\theta}}^i_{j_i(m)}||_{{\mathbf{\Sigma}}^i_{t(m,h)}} +  || \hat{\boldsymbol{\theta}}^i_{j_i(m)} - \boldsymbol{\theta}_{m,h} ||_{{\mathbf{\Sigma}}^i_{t(m,h)}} \right) \cdot || \phi_{V^i_{j_i(m)}} (\tb{\ti{s}}_{m,h}, \tb{\ti{a}}_{m,h})  ||_{{\mathbf{\Sigma}}^{i^{-1}}_{t(m,h)}}   \nonumber
	\\
	& \overset{(iv)}{\leq} 4 \beta_T || \phi_{V^i_{j_i(m)}} (\tb{\ti{s}}_{m,h}, \tb{\ti{a}}_{m,h})  ||_{{\mathbf{\Sigma}}^{i^{-1}}_{t(m,h)}}.
	\label{eqn: inner_prod_first_bound}
	\end{align}
	where $(i)$ uses the Cauchy-Schwartz inequality. In $(ii)$ we apply the triangle inequality; $(iii)$ uses the following: recall $t_{j_i(m)}$ is the time at which the $j_i(m)$-th MAEVI is triggered by agent $i$, and $t(m,h)$ is the time period corresponding to the $h$-th step in the $m$-th interval. Therefore, $t(m,h) \geq t_{j_i(m)}$. Therefore, by determinant doubling criteria we must have $det(\mb{\Sigma}_{t(m,h)}) \leq 2 \det(\mb{\Sigma}_{t_{j_i(m)}})$ otherwise, $t(m,h)$ and $t_{j_i(m)}$ would not belong to the same interval $m$. The inequality follows from $\lambda_k(\mb{\Sigma}_{t(m,h)}) \leq 2 \lambda_k(\Sigma)_{t_{j_i(m)}}$ for all $k \in [nd]$, where $\lambda_k (\cdot)$ is the $k$-th eigenvalue. Finally, $(iv)$ follows from the fact that $\boldsymbol{\theta}^{\star}$ and $\boldsymbol{\theta}_{m,h}$ belong to the confidence ellipsoid $\mathcal{C}^i_{j_i(m)}$. 
	
	Moreover, we also have the following:
	\begin{eqnarray}
	\left\langle \boldsymbol{\theta}^{\star} - \boldsymbol{\theta}_{m,h},   \phi_{V^i_{j_i(m)}} (\tb{\ti{s}}_{m,h}, \tb{\ti{a}}_{m,h})  \right\rangle
	&\leq& 
	\left\langle \boldsymbol{\theta}^{\star},  \phi_{V^i_{j_i(m)}} (\tb{\ti{s}}_{m,h}, \tb{\ti{a}}_{m,h})  \right\rangle \nonumber 
	\\
	&=& \mathbb{P} V^i_{j_i(m)} (\tb{\ti{s}}_{m,h}, \tb{\ti{a}}_{m,h})  \nonumber
	\\
	&\leq& B_{\star}.
	\label{eqn: inner_prod_second_bound}
	\end{eqnarray}
	From Equations \eqref{eqn: inner_prod_first_bound} and \eqref{eqn: inner_prod_second_bound}, we have
	\begin{eqnarray}
	\left\langle \boldsymbol{\theta}^{\star} - \boldsymbol{\theta}_{m,h},   \phi_{V^i_{j_i(m)}} (\tb{\ti{s}}_{m,h}, \tb{\ti{a}}_{m,h})  \right\rangle
	&\leq&
	\min \{ B_{\star}, 4 \beta_T || \phi_{V^i_{j_i(m)}} (\tb{\ti{s}}_{m,h}, \tb{\ti{a}}_{m,h})  ||_{{\mathbf{\Sigma}}^{i^{-1}}_{t(m,h)}}\} \nonumber
	\\
	&\leq& 4 \beta_T \min \{ 1,  || \phi_{V^i_{j_i(m)}} (\tb{\ti{s}}_{m,h}, \tb{\ti{a}}_{m,h})  ||_{{\mathbf{\Sigma}}^{i^{-1}}_{t(m,h)}}\}, \label{eqn: inner_less_than_4beta}
	\end{eqnarray}
	where the second inequality is because of the fact that $B_{\star} \leq \beta_T \leq 4\beta_T$. This implies from Equation \eqref{eqn: A1} we have,
	\begin{eqnarray}
	A_1 &\leq& \sum_{m\in \mathcal{M}_0(M)} \sum_{h=1}^{H_m} \left[  \frac{4 \beta_T}{n} \sum_{i\in S_m} \min \{ 1, || \phi_{V^i_{j_i(m)}} (\tb{\ti{s}}_{m,h}, \tb{\ti{a}}_{m,h})  ||_{{\mathbf{\Sigma}}^{i^{-1}}_{t(m,h)}}\} \right] \nonumber
	\\
	&\leq& \frac{4 \beta_T}{n} 
	\sqrt{\left( \sum_{m\in \mathcal{M}_0(M)} \sum_{h=1}^{H_m} 1 \right) \cdot \left( \sum_{m\in \mathcal{M}_0(M)} \sum_{h=1}^{H_m} \left[\sum_{i\in S_m} \min \{ 1, || \phi_{V^i_{j_i(m)}} (\tb{\ti{s}}_{m,h}, \tb{\ti{a}}_{m,h})  ||^2_{{\mathbf{\Sigma}}^{i^{-1}}_{t(m,h)}}\} \right] \right)} \nonumber 
	\\
	&=& \frac{4 \beta_T}{n} 
	\sqrt{T \cdot \left( \sum_{m\in \mathcal{M}_0(M)} \sum_{h=1}^{H_m} \left[\sum_{i\in S_m} \min \{ 1, || \phi_{V^i_{j_i(m)}} (\tb{\ti{s}}_{m,h}, \tb{\ti{a}}_{m,h})  ||^2_{{\mathbf{\Sigma}}^{i^{-1}}_{t(m,h)}}\} \right] \right)},
	\label{eqn: A1_bound_1}
	\end{eqnarray}
	the above inequality follows from the Cauchy-Schwartz inequality (product of inner term with 1. Hence summation will become the inner product). To bound the other term in the above square root, we will use the Lemma \ref{lemma: min_lemma_abbasi} given in this SM as follows:
	\begin{align}
	&  \sum_{m\in \mathcal{M}_0(M)} \sum_{h=1}^{H_m} \left[\sum_{i\in S_m} \min \{ 1, || \phi_{V^i_{j_i(m)}} (\tb{\ti{s}}_{m,h}, \tb{\ti{a}}_{m,h})  ||^2_{{\mathbf{\Sigma}}^{i^{-1}}_{t(m,h)}}\} \right] \nonumber
	\\
	&  \overset{(i)}{\leq} \sum_{m\in \mathcal{M}_0(M)} \sum_{h=1}^{H_m} \left[ \min \left\lbrace \sum_{i\in S_m} 1, \sum_{i\in S_m} || \phi_{V^i_{j_i(m)}} (\tb{\ti{s}}_{m,h}, \tb{\ti{a}}_{m,h})  ||^2_{{\mathbf{\Sigma}}^{i^{-1}}_{t(m,h)}} \right\rbrace \right] \nonumber
	\\
	&  \overset{(ii)}{\leq} \sum_{m\in \mathcal{M}_0(M)} \sum_{h=1}^{H_m} \left[ \min \left\lbrace \sum_{i\in N} 1, \sum_{i\in N} || \phi_{V^i_{j_i(m)}} (\tb{\ti{s}}_{m,h}, \tb{\ti{a}}_{m,h})  ||^2_{{\mathbf{\Sigma}}^{i^{-1}}_{t(m,h)}} \right\rbrace \right]  \nonumber
	\\
	&  = n \cdot \sum_{m\in \mathcal{M}_0(M)} \sum_{h=1}^{H_m} \left[ \min \left\lbrace 1, \frac{1}{n} \sum_{i\in N} || \phi_{V^i_{j_i(m)}} (\tb{\ti{s}}_{m,h}, \tb{\ti{a}}_{m,h})  ||^2_{{\mathbf{\Sigma}}^{i^{-1}}_{t(m,h)}} \right\rbrace \right] \nonumber
	\\
	&  \overset{(iii)}{\leq} 2 n \left[ nd \log \left( \frac{trace(\lambda \mb{I}) + T \cdot \frac{1}{n} \sum_{i\in N} B_{\star}^2 nd }{nd}\right) - \log (det(\lambda \mb{I})) \right] \nonumber
	\\
	&  =  2 n  \left[nd \log \left( \frac{\lambda nd + T B_{\star}^2 nd }{nd}\right) - \log (\lambda nd) \right]  \nonumber
	\\
	&  \leq 2 n^2d \log \left( \frac{\lambda nd + T B_{\star}^2 nd }{\lambda nd} \right)  \nonumber
	\\
	&  = 2 n^2d \log \left( 1 +\frac{ T B_{\star}^2}{\lambda} \right), \nonumber
	\end{align}
	where $(i)$ follows by interchanging the summation and min operator. In $(ii)$ we replace the sum over $i\in S_m$ by $i\in N$; and $(iii)$ follows from Lemma \ref{lemma: min_lemma_abbasi} of \cite{abbasi2011improved}, and the fact that $\max_{m\in \mathcal{M}_0(M)} || \phi_{V^i_{j_i(m)}} (\cdot, \cdot)|| \leq B_{\star} \sqrt{nd}$.
	Combining this with the Equation \eqref{eqn: A1_bound_1}, we have
	\begin{align}
	A_1 
	& \leq  \frac{4 \beta_T}{n} \sqrt{T \cdot  2 n^2d \log \left( 1 +\frac{ T B_{\star}^2}{\lambda} \right) } \nonumber
	\\ 
	&= 4 \beta_T \sqrt{2 T d \log \left( 1 +\frac{ T B_{\star}^2}{\lambda} \right) }. 
	\label{eqn: A1_bound_final}
	\end{align}
	Next consider $A_2$, recall
	\begin{align}
	A_2 = \sum_{m\in \mathcal{M}_0(M)} \sum_{h=1}^{H_m} \left[ \frac{1}{n} \sum_{i\in S_m} \frac{B_{\star} + 1}{t_{j_i(m)}} \right] \nonumber
	&\overset{(i)}{\leq} \sum_{m\in \mathcal{M}_0(M)} \sum_{h=1}^{H_m} \left[ \frac{1}{n} \sum_{i\in N} \frac{B_{\star} + 1}{t_{j_i(m)}} \right] \nonumber 
	\\
	& \overset{(ii)}{=} (B_{\star} + 1) \cdot \frac{1}{n} \sum_{i\in N} \sum_{j_i = 1}^{J} \sum_{t = t_{j_i} + 1}^{t_{j_i + 1}} \frac{1}{t_{j_i}} \nonumber
	\\
	&\overset{(iii)}{\leq} (B_{\star} + 1) \cdot \frac{1}{n} \sum_{i\in N} \sum_{j_i = 1}^J \frac{2t_{j_i}}{t_{j_i}} \nonumber
	\\
	&= 2(B_{\star}+1)J \nonumber
	\\
	&\overset{(iv)}{\leq} 2(B_{\star}+1) \left[2 n^2 d \log \left( 1 + \frac{T B_{\star}^2 nd}{\lambda} \right) + 2n \log(T)\right] \nonumber
	\\
	&= 4 (B_{\star}+1) \left[n^2 d \log \left( 1 + \frac{T B_{\star}^2 nd}{\lambda} \right) + n \log(T)\right] \nonumber
	\\
	&\overset{(v)}{\leq} 4 n^2 d (B_{\star}+1) \left[\log \left( 1 + \frac{T B_{\star}^2 nd}{\lambda} \right) + \log(T)\right], \label{eqn: A2_bound_final}
	\end{align}
	where $(i)$ follows by replacing the summation over $i\in S_m$ by summation over $i \in N$. In $(ii)$ we use the fact that the total number of time steps can be represented either as $\sum_{m\in \mathcal{M}_0(M)} \sum_{h=1}^{H_m} 1$ or $\sum_{j_i = 1}^{J} \sum_{t = t_{j_i} + 1}^{t_{j_i + 1}} 1$. Since the time doubling condition $t \geq 2 t_{j_i}$ in the algorithm implies $t_{j_i + 1} \leq 2 t_{j_i}$ for all $j_i$, we have $(iii)$. In $(iv)$ we use the bound on $J$ given in Lemma \ref{lemma: calls_to_evi}. Finally, $(v)$ follows from the fact that $n \log(T) \leq n^2 d\log(T)$.
	
	Plugging Equations \eqref{eqn: A1_bound_final} and \eqref{eqn: A2_bound_final} in Equation \eqref{eqn: E1(S_m)_part_1} we have
	\begin{eqnarray}
	& & \sum_{m\in \mathcal{M}_0(M)} \sum_{h=1}^{H_m} \left[\frac{1}{n} \sum_{i\in S_m} \bar{c}(\tb{\ti{s}}_{m,h}, \tb{\ti{a}}_{m,h}; \tb{\ti{w}}^i) + \frac{1}{n} \sum_{i\in S_m} \mathbb{P}V^i_{j_i(m)}(\tb{\ti{s}}_{m,h}, \tb{\ti{a}}_{m,h}) - \frac{1}{n} \sum_{i\in S_m} Q^{i, (l)} (\tb{\ti{s}}_{m,h}, \tb{\ti{a}}_{m,h}) \right] \nonumber
	\\
	& & \leq 4 \beta_T \sqrt{2 T d \log \left( 1 +\frac{ T B_{\star}^2}{\lambda} \right) } + 4 n^2 d (B_{\star}+1) \left[\log \left( 1 + \frac{T B_{\star}^2 nd}{\lambda} \right) + \log(T)\right].
	\label{eqn: E1(S_m)_part1_M_0}
	\end{eqnarray}
	
	Finally, to bound $E_1(S_m)$, we need to bound 
	\begin{equation*}
	\sum_{m\in \mathcal{M}_0^c(M)} \sum_{h=1}^{H_m} \left[\frac{1}{n} \sum_{i\in S_m} \bar{c}(\tb{\ti{s}}_{m,h}, \tb{\ti{a}}_{m,h}; \tb{\ti{w}}^i) + \frac{1}{n} \sum_{i\in S_m} \mathbb{P}V^i_{j_i(m)}(\tb{\ti{s}}_{m,h}, \tb{\ti{a}}_{m,h}) - \frac{1}{n} \sum_{i\in S_m} Q^{i, (l)} (\tb{\ti{s}}_{m,h}, \tb{\ti{a}}_{m,h}) \right].
	\end{equation*}
	Recall that $\mathcal{M}_0^c(M)$ is the set of all intervals $m$ such that $j_i(m) = 0$ for all $i\in N$, i.e., the intervals before the first call of MAEVI sub-routine. Since $t_0 = 1$ by triggering condition $t\geq 2t_0$ for all agent, so the first MAEVI will be called at $t=2$ by all the agents. Therefore, we have
	\begin{align}
	&  \sum_{m\in \mathcal{M}_0^c(M)} \sum_{h=1}^{H_m} \left[\frac{1}{n} \sum_{i\in S_m} \bar{c}(\tb{\ti{s}}_{m,h}, \tb{\ti{a}}_{m,h}; \tb{\ti{w}}^i) + \frac{1}{n} \sum_{i\in S_m} \mathbb{P}V^i_{j_i(m)}(\tb{\ti{s}}_{m,h}, \tb{\ti{a}}_{m,h}) - \frac{1}{n} \sum_{i\in S_m} Q^{i, (l)} (\tb{\ti{s}}_{m,h}, \tb{\ti{a}}_{m,h}) \right]  \nonumber
	\\
	&  = \sum_{h=1}^{2} \left[\frac{1}{n} \sum_{i\in S_m} \bar{c}(\tb{\ti{s}}_{m,h}, \tb{\ti{a}}_{m,h}; \tb{\ti{w}}^i) + \frac{1}{n} \sum_{i\in S_m} \mathbb{P}V^i_{j_i(m)}(\tb{\ti{s}}_{m,h}, \tb{\ti{a}}_{m,h}) - \frac{1}{n} \sum_{i\in S_m} Q^{i, (l)} (\tb{\ti{s}}_{m,h}, \tb{\ti{a}}_{m,h}) \right]  \nonumber 
	\\
	& \overset{(i)}{\leq}  \sum_{h=1}^{2} \left[\frac{1}{n} \sum_{i\in N} \bar{c}(\tb{\ti{s}}_{1,h}, \tb{\ti{a}}_{1,h}, \tb{\ti{w}}^i) + \frac{1}{n} \sum_{i\in N} \mathbb{P}V^i_{j_i(1)}(\tb{\ti{s}}_{1,h}, \tb{\ti{a}}_{1,h}) \right]  \nonumber
	\\
	&  =  \sum_{h=1}^{2} \left[\frac{1}{n} \sum_{i\in N} \bar{c}(\tb{\ti{s}}_{1,h}, \tb{\ti{a}}_{1,h}, \tb{\ti{w}}^i) + \frac{1}{n} \sum_{i\in N} \mathbb{P}V^i_{0}(\tb{\ti{s}}_{1,h}, \tb{\ti{a}}_{1,h}) \right]  \nonumber
	\\
	& \overset{(ii)}{\leq} 2(n+1),
	\label{eqn: E1(S_m)_part1_M_0^c}
	\end{align}
	where $(i)$ follows by dropping a negative term and $(ii)$ is because $\bar{c}(\tb{\ti{s}}_{1,h}, \tb{\ti{a}}_{1,h}; \tb{\ti{w}}^i) \leq n$ as the maximum congestion observed by any agent is $n$ and the private component of the cost to each agent is $K^i \leq 1$ moreover, $|V^i_0(\tb{\ti{s}})| \leq 1$.
	
	Combining Equations \eqref{eqn: E1(S_m)_part1_M_0} and \eqref{eqn: E1(S_m)_part1_M_0^c}, we have
	\begin{equation}
	\sum_{m=1}^{M}  \sum_{h=1}^{H_m} E_1(S_m) 
	\leq  4\beta_T \sqrt{2 T d \log \left( 1 +\frac{ T B_{\star}^2}{\lambda} \right) } + 4nd (B_{\star}+1) \left[\log \left( 1 + \frac{T B_{\star}^2 nd}{\lambda} \right) + \log(T) \right] + 2(n+1). 
	\label{eqn: bound_E1(Sm)_final}
	\end{equation}
	Now, to bound $E_1$, we need to bound $\sum_{m=1}^{M}  \sum_{h=1}^{H_m} E_1(S_m^c)$.
	
	\textbf{Bounding $E_1(S_m^c)$}
	
	To bound $E_1(S^c_m)$ we first observe the following: 
	Since $i\in S^c_m$, this implies that the MAEVI is not called for agent $i$ in $m$-th interval; therefore, agent $i$ will not update its optimistic estimator. So, in the $m$-th interval, agent $i$ uses the same optimistic estimator used in the $(m-1)$-th interval. Thus, there is some $1< k_i < m$ for agent $i$ such that the last MAEVI call made by agent $i$ was at $(m-k_i)$-th interval, i.e., $Q^i_{j_i(m)} = Q^i_{j_i(m-1)} = Q^i_{j_i(m-2)} = \cdots = Q^i_{j_i(m-k_i)}$. And this $Q^i_{j_i(m-k_i)}$ is obtained from the MAEVI update, hence $i\in S_{m-k_i}$, whereas $i \in S^c_{m-k_i+1}, S^c_{m-k_i+2}, \cdots, S^c_{m}$. So, from the analysis done for those agents for whom the MAEVI is called at $(m-k_i)$-th interval, and Equation \eqref{eqn: inner_less_than_4beta} we have for all agents $i\in S_{m-k_i}$ 
	\begin{equation*}
	\left\langle \boldsymbol{\theta}^{\star} - \boldsymbol{\theta}_{m-k_i,h},   \phi_{V^i_{j_i(m-k_i)}} (\tb{\ti{s}}_{m,h}, \tb{\ti{a}}_{m,h})  \right\rangle \leq 4 \beta_T \min \left\{ 1,  || \phi_{V^i_{j_i(m-k_i)}} (\tb{\ti{s}}_{m,h}, \tb{\ti{a}}_{m,h})  ||_{{\mathbf{\Sigma}}^{i^{-1}}_{t(m-k_i,h)}}\right\}.
	\end{equation*}
	
	Moreover, from equation \eqref{eqn: e_1_s_m_interim}, we have 
	\begin{eqnarray*}
		&  & \bar{c}(\tb{\ti{s}}_{m,h}, \tb{\ti{a}}_{m,h}; \tb{\ti{w}}^i) + \mathbb{P}V^i_{j_i(m-k_i)}(\tb{\ti{s}}_{m,h}, \tb{\ti{a}}_{m,h}) - Q^{i, (l)} (\tb{\ti{s}}_{m,h}, \tb{\ti{a}}_{m,h}) 
		\\
		&\leq&
		\left\langle \boldsymbol{\theta}^{\star} - \boldsymbol{\theta}_{m-k_i,h}, \phi_{V^i_{j_i(m-k_i)}} (\tb{\ti{s}}_{m,h}, \tb{\ti{a}}_{m,h})  \right\rangle  + \frac{B_{\star} + 1}{t_{j_i(m-k_i)}}
		\\
		& \leq & 4 \beta_T \min \{ 1,  || \phi_{V^i_{j_i(m-k_i)}} (\tb{\ti{s}}_{m,h}, \tb{\ti{a}}_{m,h})  ||_{{\mathbf{\Sigma}}^{i^{-1}}_{t(m-k_i,h)}}\} + \frac{B_{\star} + 1}{t_{j_i(m-k_i)}}. 
	\end{eqnarray*}
	Also, note that $V^i_{j_i(m)} = V^i_{j_i(m-1)} = V^i_{j_i(m-2)} = \cdots = V^i_{j_i(m-k_i)}$, this is because MAEVI is not called in the intermediate intervals by agent $i$, as $V^i_{j_i(m)}(\tb{\ti{s}}_{m,h}) =  \min_{\tb{\ti{a}}} Q^i_{j_i(m)}(\tb{\ti{s}}_{m,h}, \tb{\ti{a}}) $. So for each agent, $i\in S^c_m$, the above equation can be written as
	\begin{equation*}
	\bar{c}(\tb{\ti{s}}_{m,h}, \tb{\ti{a}}_{m,h}; \tb{\ti{w}}^i) + \mathbb{P}V^i_{j_i(m)}(\tb{\ti{s}}_{m,h}, \tb{\ti{a}}_{m,h}) - Q^{i, (l)} (\tb{\ti{s}}_{m,h}, \tb{\ti{a}}_{m,h})  \leq 4 \beta_T \min \{ 1,  || \phi_{V^i_{j_i(m)}} (\tb{\ti{s}}_{m,h}, \tb{\ti{a}}_{m,h})  ||_{{\mathbf{\Sigma}}^{i^{-1}}_{t(m,h)}}\} + \frac{B_{\star} + 1}{t_{j_i(m)}}. 
	\end{equation*}
	Furthermore, this is true for all the agents $i\in S^c_m$. Taking summation over $i\in S^c_m$, we have 
	\begin{align}
	& \frac{1}{n} \sum_{i\in S^c_m} \left[ \bar{c}(\tb{\ti{s}}_{m,h}, \tb{\ti{a}}_{m,h}; \tb{\ti{w}}^i) + \mathbb{P}V^i_{j_i(m)}(\tb{\ti{s}}_{m,h}, \tb{\ti{a}}_{m,h}) - Q^{i, (l)} (\tb{\ti{s}}_{m,h}, \tb{\ti{a}}_{m,h}) \right] \nonumber
	\\
	& \leq
	\frac{4 \beta_T}{n} \sum_{i\in S^c_m} \min \{ 1,  || \phi_{V^i_{j_i(m)}} (\tb{\ti{s}}_{m,h}, \tb{\ti{a}}_{m,h})  ||_{{\mathbf{\Sigma}}^{i^{-1}}_{t(m,h)}}\} + \frac{1}{n} \sum_{i\in S^c_m} \frac{B_{\star} + 1}{t_{j_i(m)}}  \nonumber
	\\
	& \overset{(i)}{\leq}
	\frac{4 \beta_T}{n}  \min \left\lbrace \sum_{i\in S^c_m} 1,  \sum_{i\in S^c_m} || \phi_{V^i_{j_i(m)}} (\tb{\ti{s}}_{m,h}, \tb{\ti{a}}_{m,h})  ||_{{\mathbf{\Sigma}}^{i^{-1}}_{t(m,h)}} \right\rbrace + \frac{1}{n} \sum_{i\in S^c_m} \frac{B_{\star} + 1}{t_{j_i(m)}}  \nonumber
	\\
	&\overset{(ii)}{\leq}  \frac{4 \beta_T}{n}  \min \left\lbrace \sum_{i\in N} 1,  \sum_{i\in N} || \phi_{V^i_{j_i(m)}} (\tb{\ti{s}}_{m,h}, \tb{\ti{a}}_{m,h})  ||_{{\mathbf{\Sigma}}^{i^{-1}}_{t(m,h)}} \right\rbrace + \frac{1}{n} \sum_{i\in N} \frac{B_{\star} + 1}{t_{j_i(m)}}  \nonumber
	\\
	&\leq  4 \beta_T  \min \left\lbrace 1,  \frac{1}{n} \sum_{i\in N} || \phi_{V^i_{j_i(m)}} (\tb{\ti{s}}_{m,h}, \tb{\ti{a}}_{m,h})  ||_{{\mathbf{\Sigma}}^{i^{-1}}_{t(m,h)}} \right\rbrace + \frac{1}{n} \sum_{i\in N} \frac{B_{\star} + 1}{t_{j_i(m)}}, \nonumber
	\end{align}
	where $(i)$ follows by interchanging min and summation. In $(ii)$ we replace summation over $i\in S^c_m$ by summation over $i\in N$. 
	
	Since, $Q^i_{j_i(m)}$ is same as $Q^i_{j_i(m-k_i)}$, taking summation over $m$, and $h$ in the above we will have the same bound as we have for $E_1(S_m)$. This implies,
	\begin{align}
	E_1(S^c_m) & =  \sum_{m=1}^{M} \sum_{h=1}^{H_m} \left[ \frac{1}{n} \sum_{i\in S^c_m} \left[ \bar{c}(\tb{\ti{s}}_{m,h}, \tb{\ti{a}}_{m,h}; \tb{\ti{w}}^i) + \mathbb{P}V^i_{j_i(m)}(\tb{\ti{s}}_{m,h}, \tb{\ti{a}}_{m,h}) - Q^{i, (l)} (\tb{\ti{s}}_{m,h}, \tb{\ti{a}}_{m,h}) \right] \right] \nonumber
	\\
	&  \leq \sum_{m=1}^{M} \sum_{h=1}^{H_m} \left[ 4 \beta_T  \min \left\lbrace 1,  \frac{1}{n} \sum_{i\in N} || \phi_{V^i_{j_i(m)}} (\tb{\ti{s}}_{m,h}, \tb{\ti{a}}_{m,h})  ||_{{\mathbf{\Sigma}}^{i^{-1}}_{t(m,h)}} \right\rbrace + \frac{1}{n} \sum_{i\in N} \frac{B_{\star} + 1}{t_{j_i(m)}} \right] \nonumber
	\\
	&  \leq 4\beta_T \sqrt{2 T d \log \left( 1 +\frac{ T B_{\star}^2}{\lambda} \right) } + 4nd (B_{\star}+1) \left[\log \left( 1 + \frac{T B_{\star}^2 nd}{\lambda} \right) + \log(T) \right] + 2(n+1), 
	\label{eqn: bound_E1(Smc)_final}
	\end{align}
	the last inequality uses the same ideas as used for bounding $\sum_{m=1}^{M}  \sum_{h=1}^{H_m} E_1(S_m)$. Combining Equations \eqref{eqn: bound_E1(Sm)_final} and \eqref{eqn: bound_E1(Smc)_final}, we have the bound for $E_1$ as follows:
	\begin{align}
	E_1 & = \sum_{m=1}^{M}  \sum_{h=1}^{H_m} [E_1(S_m) + E_1(S^c_m)]  \nonumber
	\\
	&\leq 
	8 \beta_T \sqrt{2 T d \log \left( 1 +\frac{ T B_{\star}^2}{\lambda} \right) } + 8 nd (B_{\star}+1) \left[\log \left( 1 + \frac{T B_{\star}^2 nd}{\lambda} \right) + 2 \log(T) \right] + 4 (n+1). \nonumber
	\end{align}
	This ends the proof.
\end{proof}

\subsection{Proof of Proposition \ref{prop: E2_bound} (Bounding $E_2$)}
\label{app: bound_E2}

\begin{proof}
	Recall $E_2$ is given by
	\begin{equation*}
	E_2  = \sum_{m=1}^M \sum_{h=1}^{H_m} \left[ \frac{1}{n} \sum_{i=1}^n V^i_{j_i(m)}(\tb{\ti{s}}_{m,h+1}) - \frac{1}{n} \sum_{i=1}^n \mathbb{P}V^i_{j_i(m)}(\tb{\ti{s}}_{m,h}, \tb{\ti{a}}_{m,h}) \right].
	\end{equation*}
	The first thing to note is that $E_2$ is the sum of the martingale differences. However, the function $V^i_{j_i(m)}$ is random, not necessarily bounded. So we can apply Azuma-Hoeffding inequality. Let us define an filtration $\{\mathcal{F}_{m,h}\}_{m,h}$ such that $\mathcal{F}_{m,h}$ is the $\sigma$-field of all the history up until $(\tb{\ti{s}}_{m,h}, \tb{\ti{a}}_{m,h})$ but doesn't contain $\tb{\ti{s}}_{m,h+1}$. Thus, $(\tb{\ti{s}}_{m,h}, \tb{\ti{a}}_{m,h})$ is $\mathcal{F}_{m,h}$-measurable. Moreover, $V^i_{j_i(m)}$ is $\mathcal{F}_{m,h}$ measurable. By definition of operator $\mathbb{P}$, we have $\mathbb{E}[V^i_{j_i(m)}| \mathcal{F}_{m,h}] = \mathbb{P} V^i_{j_i(m)}$ for all $i\in N$, which shows that $E_2$ is martingale difference sequence. To deal with the problem that $V^i_{j_i(m)}$ might not be bounded, define an auxiliary sequence
	\begin{equation*}
	\widetilde{V}^i_{j_i(m)} \coloneqq \min \{ B_{\star}, V^i_{j_i(m)} \},
	\end{equation*}
	It follows that $\widetilde{V}^i_{j_i(m)}$ is $\mathcal{F}_{m,h}$-measurable. 
	\begin{eqnarray}
	E_2 &=& \sum_{m=1}^M \sum_{h=1}^{H_m} \left[ \frac{1}{n} \sum_{i=1}^n \widetilde{V}^i_{j_i(m)}(\tb{\ti{s}}_{m,h+1}) - \frac{1}{n} \sum_{i=1}^n \mathbb{P} \widetilde{V}^i_{j_i(m)}(\tb{\ti{s}}_{m,h}, \tb{\ti{a}}_{m,h}) \right] \nonumber
	\\
	& & + \sum_{m=1}^M \sum_{h=1}^{H_m} \left[ \frac{1}{n} \sum_{i=1}^n [V^i_{j_i(m)} - \widetilde{V}^i_{j_i(m)}] (\tb{\ti{s}}_{m,h+1}) - \frac{1}{n} \sum_{i=1}^n \mathbb{P} [V^i_{j_i(m)} - \widetilde{V}^i_{j_i(m)}] (\tb{\ti{s}}_{m,h}, \tb{\ti{a}}_{m,h}) \right]. \nonumber
	\end{eqnarray}
	Since $\widetilde{V}^i_{j_i(m)}$ is bounded, we can apply the Azuma-Hoeffding inequality Lemma \ref{lemma: azuma-hoeff} and get with probability at least $1-\delta/2n$ that 
	\begin{align*}
	E_2 &= \frac{1}{n} \sum_{i=1}^n 2B_{\star} \sqrt{2T \log \left(\frac{T}{\delta/2n} \right)} + 
	\\
	& ~~~~ \sum_{m=1}^M \sum_{h=1}^{H_m} \left[ \frac{1}{n} \sum_{i=1}^n [V^i_{j_i(m)} - \widetilde{V}^i_{j_i(m)}] (\tb{\ti{s}}_{m,h+1}) - \frac{1}{n} \sum_{i=1}^n \mathbb{P} [V^i_{j_i(m)} - \widetilde{V}^i_{j_i(m)}] (\tb{\ti{s}}_{m,h}, \tb{\ti{a}}_{m,h}) \right].
	\end{align*}
	Note that under the MAEVI analysis and optimism, we have $\widetilde{V}^i_{j_i(m)} = V^i_{j_i(m)}$ for all $j_i(m) \geq 1$ for all $i\in N$. Moreover, initialization $\widetilde{V}^i_0 = V^i_0$ for all $i\in N$ implies the second term in RHS is zero. Thus, with probability at least $1-\delta$ we have 
	\begin{equation*}
	E_2 = \frac{1}{n} \sum_{i=1}^n 2B_{\star} \sqrt{2T \log \left(\frac{2nT}{\delta} \right)} = 2B_{\star} \sqrt{2T \log \left(\frac{2nT}{\delta} \right)},
	\end{equation*}
	This ends the proof.
\end{proof}

\section{SOME USEFUL RESULTS}
\label{app: additional-useful-results}
\begin{theorem}[Theorem 1, \cite{abbasi2011improved}]
	\label{thm: abbasi_yadkori_thm_1}
	Let $\{\mathcal{F}_t\}_{t=0}^\infty$ be a filtration. Suppose $\{\eta_t\}_{t=1}^\infty$ is a $\mathbb{R}$-valued stochastic process such that $\eta_t$ is $\mathcal{F}_{t}$-measurable and $\eta_t | \mathcal{F}_{t-1}$ is $B$-sub-Gaussian. Let $\{\phi_t\}_{t=1}^\infty$ be an $\mathbb{R}^d$-valued stochastic process such that $\phi_t$ is $\mathcal{F}_{t-1}$-measurable. Assume that $\mathbf{\Sigma}$ is an $d \times d$ positive definite matrix. For any $t\geq 1$, define 
	\begin{equation*}
	\mathbf{\Sigma}_t = \mathbf{\Sigma} + \sum_{k=1}^t \phi_k \phi_k^\top ~and~ \mathbf{a}_t = \sum_{k=1}^t \eta_k \phi_k.
	\end{equation*}
	Then, for any $\delta>0$, with probability at least $\delta$, for all $t$, we have 
	\begin{equation*}
	\| \mathbf{\Sigma}_t^{-1/2}\mathbf{a}_t\|_2 \leq B \sqrt{2 \log \left( \frac{\det(\mathbf{\Sigma}_t)^{1/2}}{\delta \cdot \det(\mathbf{\Sigma})^{1/2}} \right)} .
	\end{equation*}
\end{theorem}

\begin{theorem}[Determinant-trace inequality; Lemma 11 \cite{abbasi2011improved}]
	\label{thm: det_trace_inequality}
	Assume $\phi_1, \phi_2, \dots, \phi_t \in \mathbb{R}^{d}$ and for any $s \leq t~, ||\phi_s ||_2 \leq L$. Let $\lambda > 0$ and $\mathbf{\Sigma}_t = \lambda \mathbf{I} + \sum_{s=1}^t \phi_s \phi_s^{\top}$. Then
	\begin{equation*}
	det(\mathbf{\Sigma}_t) \leq (\lambda + t L^2/d)^d.
	\end{equation*}
\end{theorem}

\begin{lemma}[Lemma 11 in \cite{abbasi2011improved}]
	\label{lemma: min_lemma_abbasi}
	Let $\{\phi_t\}_{t=1}^{\infty}$ be in $\mathbb{R}^d$ such that $|| \phi_t|| \leq L$ for all $t$. Assume $\mb{\Sigma}_0$ is psd matrix in $\mathbb{R}^{d\times d}$, and let $\mb{\Sigma}_t = \mb{\Sigma}_0 + \sum_{s=1}^t \phi_s \phi_s^{\top}$. Then we have 
	\begin{equation*}
	\sum_{s=1}^t \min \{ 1, || \phi_s||_{\mb{\Sigma}^{-1}_{s-1}}\} \leq 2 \left[ d \log \left( \frac{trace(\mb{\Sigma}_0) + tL^2}{d}\right) - \log det(\mb{\Sigma}_0) \right].
	\end{equation*}
\end{lemma}

\begin{lemma}[Azuma-Hoeffding inequality, anytime version]
	\label{lemma: azuma-hoeff}
	Let $\{X_t\}_{t=0}^{\infty}$ be a real-valued martingale such that for every $t \geq 1$, it holds that $|X_t - X_t-1| \leq B$  for some $B \geq 0$. Then for any $0 < \delta \leq 1/2$, with probability at least $1 - \delta$, the following holds for all $t \geq 0$ 
	\begin{equation*}
	|X_t - X_0| \leq 2B \sqrt{2t \log \left(\frac{t}{\delta} \right) }.
	\end{equation*}
\end{lemma}

\begin{lemma}[Kushner-Clark Lemma \cite{kushner2003stochastic,metivier1984applications}]
	\label{app: K-C_lemma}
	Let $\mathcal{X}\subseteq \mathbb{R}^p$ be a compact set and let $h: \mathcal{X} \rightarrow \mathbb{R}^p$ be a continuous function. Consider the following recursion in $p$-dimensions 
	\begin{equation}
	\label{eqn: x_recursion}
	x_{t+1} = \Gamma\{x_t + \gamma_t[h(x_t) + \zeta_t + \beta_t]\}.
	\end{equation}
	Let $\hat{\Gamma}(\cdot)$ be transformed projection operator defined for any $x\in \mathcal{X}\subseteq \mathbb{R}^{p}$ as
	\begin{equation*}
	\hat{\Gamma}(h(x)) = \lim_{0< \eta \rightarrow 0} \left\lbrace \frac{\Gamma(x+\eta h(x)) - x}{\eta} \right\rbrace,
	\end{equation*}
	then the ODE associated with Equation (\ref{eqn: x_recursion}) is $\dot x = \hat{\Gamma} (h(x))$.
	\begin{assumption}
		\label{ass: K-C_lemma}
		Kushner-Clark lemma requires the following assumptions
		\begin{enumerate}
			\item Stepsize $\{\gamma_t\}_{t\geq 0}$ satisfy $\sum_{t} \gamma_t = \infty$, and $\gamma_t \rightarrow 0$ as $t\rightarrow \infty$.
			\item The sequence $\{\beta_t\}_{t\geq 0}$ is a bounded random sequence with $\beta_t \rightarrow 0$ almost surely as $t\rightarrow \infty$.
			\item For any $\epsilon > 0$, the sequence $\{\zeta_t\}_{t\geq 0}$ satisfy
			\begin{equation*}
			\lim_t~\mathbb{P}\left( sup_{p\geq t} \left\Vert \sum_{\tau = t}^p \gamma_{\tau}\zeta_{\tau} \right\Vert
			\geq \epsilon \right) = 0.
			\end{equation*}
		\end{enumerate}
	\end{assumption}
	
	Kushner-Clark lemma is as follows: suppose that ODE $\dot x = \hat{\Gamma} (h(x))$ has a compact set $\mathcal{K}^{\star}$ as its asymptotically stable equilibria, then under Assumption  \ref{ass: K-C_lemma}, $x_t$ in Equation (\ref{eqn: x_recursion}) converges almost surely to $\mathcal{K}^{\star}$ as $t\rightarrow \infty$. 
\end{lemma}

\end{document}